\documentclass[letterpaper]{article} 
\usepackage{aaai25}  
\usepackage{times}  
\usepackage{helvet}  
\usepackage{courier}  
\usepackage[hyphens]{url}  
\usepackage{graphicx} 
\usepackage{natbib}  
\usepackage{caption} 
\usepackage{algorithmic}

\usepackage{newfloat}
\usepackage{listings}
\DeclareCaptionStyle{ruled}{labelfont=normalfont,labelsep=colon,strut=off} 
\usepackage{amsmath}
\usepackage{amsfonts}
\usepackage{subcaption}
\usepackage{enumerate}
\usepackage{amsthm}
\usepackage{booktabs}
\usepackage{multirow}
\usepackage{dsfont} 
\usepackage{amssymb}
\usepackage{colortbl}
\usepackage{pifont}
\DeclareMathOperator*{\argmax}{arg\,max}
\usepackage{afterpage}
\usepackage[ruled,vlined]{algorithm2e}
\usepackage[table,dvipsnames ]{xcolor}
\usepackage[toc,page,header]{appendix}
\usepackage{etoc}
\usepackage{multicol}

\theoremstyle{plain}
\newtheorem{theorem}{Theorem}[section]
\newtheorem{proposition}[theorem]{Proposition}

\theoremstyle{definition}
\newtheorem{definition}[theorem]{Definition}

\theoremstyle{remark}

\newtheorem{example}{Example}

\definecolor{commentcolor}{RGB}{110,154,155}   
\newcommand{\PyComment}[1]{\ttfamily\textcolor{commentcolor}{\# #1}}  
\newcommand{\PyCode}[1]{\ttfamily\textcolor{black}{#1}} 
\newlength\myindent
\definecolor{Gray}{gray}{0.8}

\newcolumntype{a}{>{\columncolor{Gray}}c}
\newcolumntype{a}{>{\columncolor{Gray}}c}

\title{Faithful and Accurate Self-Attention Attribution for Message Passing Neural Networks via the Computation Tree Viewpoint}
\author{
    Yong-Min Shin\textsuperscript{\rm 1},
    Siqing Li\textsuperscript{\rm 2},
    Xin Cao\textsuperscript{\rm 2},
    Won-Yong Shin\textsuperscript{\rm 1}\thanks{Corresponding Author.}
}
\affiliations{
    \textsuperscript{\rm 1}Yonsei University
    \textsuperscript{\rm 2}University of New South Wales

    jordan3414@yonsei.ac.kr, \{siqing.li, xin.cao\}@unsw.edu.au, wy.shin@yonsei.ac.kr
}

\usepackage{bibentry}

\begin{document}

\maketitle

\begin{abstract}
The self-attention mechanism has been adopted in various popular message passing neural networks (MPNNs), enabling the model to adaptively control the amount of information that flows along the edges of the underlying graph. Such attention-based MPNNs (Att-GNNs) have also been used as a baseline for multiple studies on explainable AI (XAI) since attention has steadily been seen as natural model interpretations, while being a viewpoint that has already been popularized in other domains (\textit{e.g.,} natural language processing and computer vision). However, existing studies often use na\"ive calculations to derive attribution scores from attention, undermining the potential of attention as interpretations for Att-GNNs. In our study, we aim to fill the gap between the widespread usage of Att-GNNs and their potential \textit{explainability} via attention. To this end, we propose \textsc{GAtt}, edge attribution calculation method for self-attention MPNNs based on the {\it computation tree}, a rooted tree that reflects the computation process of the underlying model. Despite its simplicity, we empirically demonstrate the effectiveness of \textsc{GAtt} in three aspects of model explanation: faithfulness, explanation accuracy, and case studies by using both synthetic and real-world benchmark datasets. In all cases, the results demonstrate that \textsc{GAtt} greatly improves edge attribution scores, especially compared to the previous na\"ive approach. Our code is available at \url{https://github.com/jordan7186/GAtt}.
\end{abstract}



\section{Introduction}
\label{section:introduction}
\paragraph{Background \& motivation.}
In graph learning, graph neural networks (GNNs)~\cite{Wu2021GNNSurvey} have been used as the \textit{de facto} architecture, since they can effectively encode the graph structure along with node (or edge) features. Among various GNNs, several models have successfully incorporated the self-attention mechanism~\cite{Vaswani2017attention} into message passing neural networks (MPNNs)~\cite{Gilmer2017mpnn, Bronstein2021mpnn}. Such (self-)attention-based MPNNs (dubbed \textbf{Att-GNNs}) have been one of the staple GNN architectures, and the self-attention mechanism of Att-GNNs themselves has been extensively analyzed in the literature~\cite{Knyazev2019graphattention, Lee2019gatsurvey, Sun2023gatsurvey}.\footnote{As such, we refer to the term `GNN' as MPNN-style architectures unless explicitly stated otherwise.} Furthermore, several studies have focused solely on analyzing GAT~\cite{Velickovic2018gat}, the most representative Att-GNN model~\cite{Mustafa2023gat, Fountoulakiso2023gat}.

Similarly as in other neural network models, GNNs are regarded as black-box models that lack interpretability, which has led to numerous studies developing explanation methods for GNNs~\cite{Li2022GNNXAISurvey, Yuan2023GNNXAISurvey}. While such explanation methods have been widely developed, attention has also been frequently considered as a fundamental tool for GNN explanations~\cite{Ying2019GNNExplainer, Luo2020PGExplainer, SanchezLengeling2020XAIGNNeval}. The choice of attention as a baseline is natural, as self-attention itself can be viewed as a direct way to provide model interpretations without any separate explanation method~\cite{Lee2017attentionasexpl, Ghaeini2018attentionasexpl, Hao2021attattr, Aflalo2022attention, Deiseroth2023attention}. This viewpoint has already been extensively investigated in transformers, the most representative architecture with attention~\cite{Bahdanau2014attention, Xu2015attentionvisualization, Vig2019attentionvisualization, Dosovitskiy2021ViT16X16, Caron2021attentionvisualization}. There is even a significant body of research debating the validity of self-attention as explanations in natural language processing (NLP)~\cite{JainW2019attentiondebate, Wiegreffe2019attentiondebate, Bibal2022attentiondebate}. However, there has been no such in-depth discussion from the domain of GNN explanations, mostly employing the layer-wise average of attention weights retrieved from a GAT model as explanations at best.

We argue that \textbf{such na\"ive usage of attention for interpretations largely undermines the potential of Att-GNNs as an explainable model}. In the case of transformers, a number of advanced attribution methods using attention have been proposed to calculate token attributions, and have been empirically proven that attention can be effectively used to decipher the underlying model~\cite{Abnar2020attentionflow, Chefer2021attentionmapiccv, Chefer2021attentionmapcvpr, Hao2021attattr}. Analogous to transformers, our study aims to formulate a post-processing method for the attention weights in Att-GNNs that is able to extract high-quality \textit{edge attributions} ({\it i.e.}, to assign contributions of edges to the model) and capture the behavior of Att-GNNs more precisely. To the best of our knowledge, we are the first to address this issue within the scope of general Att-GNNs, thus filling in the literature of explanations via attention (see the red part of Table~\ref{table:introduction}).

\begin{table}[t]
\centering
\caption{Overview of prior works on using attention as explanations. Despite various methods being developed for calculating token attributions for transformers, no corresponding method has yet been developed for calculating edge attributions for Att-GNNs.}\label{table:introduction}
    \resizebox{\linewidth}{!}{
    \begin{tabular}{lcc}
    \toprule
    \textbf{Model} & \textbf{Na\"ive methods} & \textbf{Advanced methods} \\
    \midrule
    \multirow{4}{*}{Transformers} &  & \cite{Abnar2020attentionflow} \\
     & \textit{Raw attention} & \cite{Chefer2021attentionmapiccv} \\
     & \cite{Vig2019attentionvisualization} & \cite{Chefer2021attentionmapcvpr} \\
     &  & \cite{Hao2021attattr} etc.\\\midrule
     & \textit{Layer-wise averaging} & \cellcolor{red!25} \\ 
    Att-GNNs & \cite{Ying2019GNNExplainer} & \cellcolor{red!25} \textbf{This work (\textsc{GAtt})} \\
     & \cite{Luo2020PGExplainer} & \cellcolor{red!25} \\
    \bottomrule
   \end{tabular}
   }
   \vskip -0.1in
\end{table}

\paragraph{Main contributions.}
In this study, we address the problem of developing an effective edge attribution method using attention weights in Att-GNNs. Our key insight is that \textbf{edge attributions with attention can be advanced by aligning with the feed-forward process of MPNNs}, \textit{i.e.}, thinking in terms of the \textit{computation tree} viewpoint, a rooted subtree that shows the local computation structure around a target node (see the middle part of Figure~\ref{fig:examplemain}). Based on observing the computation tree, we assert that the edge attribution function should encompass two crucial principles: {\bf P1)} proximity to the target node and {\bf P2)} its position in the computation tree, thus aligning with the feed-forward process.

To this end, we introduce \textsc{GAtt}, a simple yet effective solution to the edge attribution problem by integrating the {\it computation tree} of a given target node. Specifically, \textsc{GAtt} adds attention weights in the underlying Att-GNN across the computation tree while adjusting their influence by employing targeted multiplication factors for attention weights guiding towards the target node. As an example, Figure~\ref{fig:examplemain} visualizes edge attribution scores from different edge attribution calculation methods using the same model. The attribution scores from \textsc{GAtt} (see the right red box in Figure~\ref{fig:examplemain}) show that the model places high emphasis on the correct infection path (highlighted as blue nodes). Such conclusion could not have been reached if we were to use simple layer-wise averaging (see the left box in Figure~\ref{fig:examplemain}) as the tool for interpretations. To prove the effectiveness of \textsc{GAtt}, we run extensive experiments by answering pivotal facets of interpretations—\textit{faithfulness} and \textit{explanation accuracy}—of Att-GNNs across diverse real-world and synthetic datasets. Despite the simplicity of \textsc{GAtt}, empirical results demonstrate that the application of \textsc{GAtt} to process attention weights within the underlying model produces substantively improved explanation capabilities, excelling in both faithfulness and explanation accuracy. We also perform an ablation study in which we introduce two variants of \textsc{GAtt}, namely \textsc{GAtt}$_{\textsc{sim}}$ and \textsc{GAtt}$_{\textsc{avg}}$, each of which corresponds to a removal of one critical design element (\textit{i.e.}, \textbf{P1} or \textbf{P2}) of our method. Our analysis reveals clear deterioration of the quality of the edge attributions in all measures for both variants, which justifies the necessity of our two design elements. Finally, we remark that \textsc{GAtt} is a straightforward calculation module (\textit{i.e.}, does not involve any optimization/learning process), therefore brings the benefit of being hyperparameter-free and deterministic. In summary, we conclude that \textbf{Att-GNNs are indeed highly explainable} when adopting the proper interpretation, \textit{i.e.,} \textbf{adjustment of attention weights} by \textbf{taking the viewpoint of the computation tree}. Note that graph transformers~\cite{Ying2021graphtransformers, Kreuzer2021SAN, Chen2023graphtransformer} are beyond the scope of this study since transformers have already been analyzed and advanced by numerous studies (see Table~\ref{table:introduction}). Our contributions are summarized as follows:
\begin{figure}[t]
  \centering
    \includegraphics[width=\columnwidth]{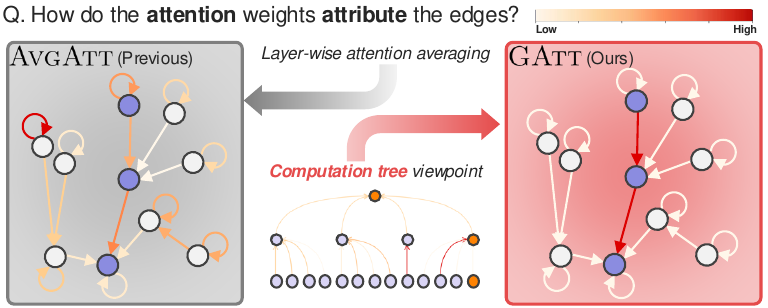}
  \caption{An visualization of our method (\textsc{GAtt}, right) against the previous approach (\textsc{AvgAtt}, left) on the Infection dataset, where the correct infection path is highlighted as the blue nodes.}
  \vskip -0.1in
  \label{fig:examplemain}
\end{figure}
\begin{itemize}
    \item {\bf Key observations:} We make key observations and design principles that are crucial in edge attribution calculation by integrating the computation tree of the target node during its feed-forward process.
    \item {\bf Novel methodology:} We propose \textsc{GAtt}, a new method to calculate edge attributions from attention weights in Att-GNNs by integrating the computation tree of the given GNN model.
    \item {\bf Extensive evaluations:} We extensively demonstrate that Att-GNNs are shown to be more faithful and accurate when using our proposed method compared to the simple alternative.
\end{itemize}

It should be noted that as long as Att-GNN architectures are employed, \textsc{GAtt} is \textit{model-agnostic} and standalone without any learning modules. We refer to Appendix~\ref{section:relatedworks} for a comprehensive review of related studies.

\section{Edge Attribution Calculation in Att-GNNs}
\label{section:attributioncalculation}

In this section, we first describe the notation used in the paper. Then, we formalize the problem of calculating edge attributions in Att-GNNs, and propose \textsc{GAtt}, an approach to incorporate the computation tree into edge attributions.

\subsection{Notations} 
Let us denote a given undirected graph as a tuple $G = (\mathcal{V}, \mathcal{E})$, where $\mathcal{V}$ is the set of nodes and $\mathcal{E}$ is the set of edges. We denote the edge connecting two nodes $v_i, v_j \in \mathcal{V}$ as $e_{ij} \in \mathcal{E}$. We consider undirected graphs, \textit{i.e.}, $e_{j,i} \in \mathcal{E}$ if $e_{i,j} \in \mathcal{E}$. The set of neighbors of node $v_i$ is denoted as $\mathcal{N}_i$. 

\subsection{Problem Statement} 
\label{subsection:problemstatement}
We are given a graph $G = (\mathcal{V}, \mathcal{E})$, the Att-GNN model $f$ with $L$ layers, and a target node $v_i \in \mathcal{V}$ of interest. The attention weights calculated from $f$ are denoted as $\mathcal{A} = \{{\bf A}(l)\}_{l=1}^{L}$, where ${\bf A}(l) \in \mathbb{R}^{|\mathcal{V}| \times |\mathcal{V}|}$ and $[{\bf A}(l)]_{j,i} = \alpha_{i,j}^l$ is the attention weight of edge $e_{i,j}$ in the $l$-th layer (with $l=1$ being the input layer). The problem of edge attribution calculation is characterized by an edge attribution function $\Phi(v, \mathcal{A}, e_{i,j}) \triangleq \phi_{i,j}^v$ such that the \textit{edge attribution score} $\phi_{i,j}^v$ accounts for the contribution of edge $e_{i,j}$ to the underlying model's calculation for node $v$ (\textit{i.e.,} faithfulness to $f$). 

In our study, our objective is to design $\Phi$ using the {\it computation tree} in Att-GNNs alongside several observations and key design principles, which will be specified later. To design such a function $\Phi$, we argue that the \textbf{computation tree of Att-GNNs} should be considered for the precise calculation of $\phi_{i,j}^v$, incorporating its several key properties. Note that, although most post hoc instance-level explanation methods for GNNs~\cite{Ying2019GNNExplainer, Luo2020PGExplainer} also have a similar objective in terms of calculating $\phi_{i,j}^v$, they do not take advantage of attention weights $\mathcal{A}$. Additionally, although we mainly consider \textit{node-level} tasks throughout the paper as a representative task, we also demonstrate that the idea of \textsc{GAtt} can also be extended for \textit{graph-level} tasks (see Appendix~\ref{appendix:graphleveltask}).

\subsection{From Attention to Attribution}
\label{subsection:proposedattribution}
We first visualize the computation tree in a Att-GNN, which will lead to several important observations to guide \textsc{GAtt}, an edge attribution calculation method given the attention weights in the Att-GNN model.

\begin{figure}[t]
  \centering
  \begin{subfigure}[b]{0.31\linewidth}
    \includegraphics[width=\linewidth]{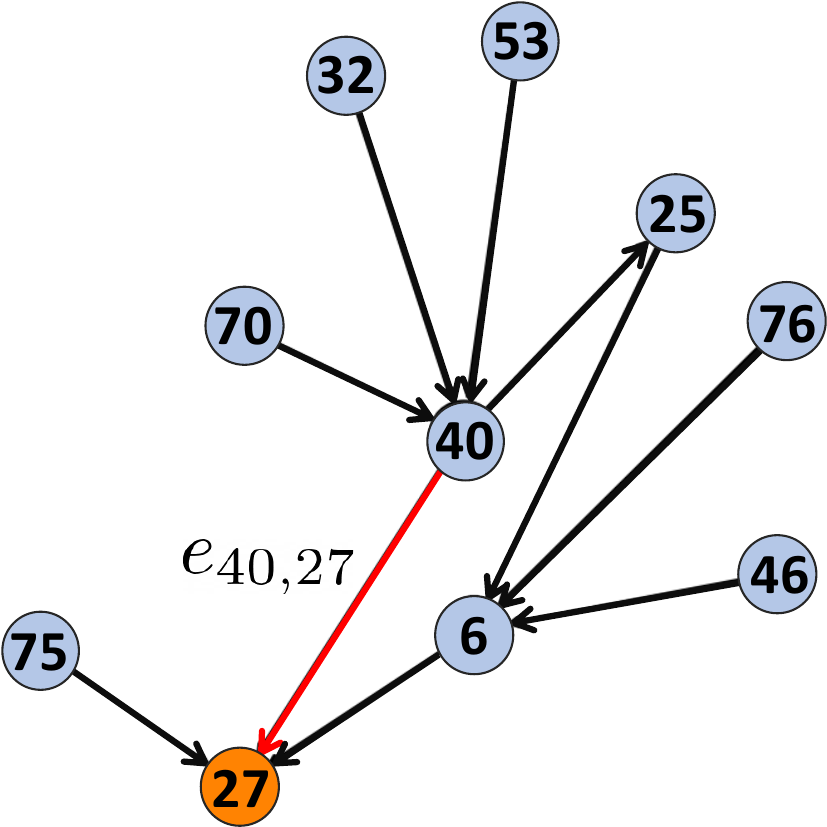}
    \caption{2-hop subgraph.}
    \label{subfig:infectiondatasetexample}
  \end{subfigure} 
  \begin{subfigure}[b]{0.66\linewidth}
    \includegraphics[width=\linewidth]{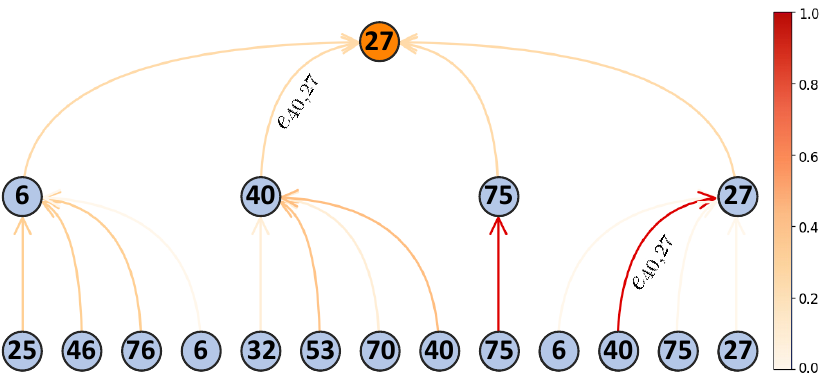}
    \caption{Computation tree.}
    \label{subfig:infectiondatasetcomputationgraph}
  \end{subfigure}
  \caption{A visualization for a 2-layer Att-GNN on target node 27 on the infection dataset. Figure~\ref{subfig:infectiondatasetexample} shows the local 2-hop subgraph with the edge $e_{40, 27}$ marked as red. Figure~\ref{subfig:infectiondatasetcomputationgraph} shows the computation tree in the Att-GNN, where the information flows from leaf nodes to node 27 at the root. The edges are colored by the attention weights from the model, while highlighting the two occurrences of edge $e_{40, 27}$.}
  \vskip -0.1in
  \label{fig:infectiondataset}
\end{figure}

\paragraph{Visualizing the Computation Tree.} To provide an illustrative example, we train a 2-layer GAT model~\cite{Velickovic2018gat} with a single attention head on the synthetic infection benchmark dataset~\cite{Faber2021gnnxaidataset}. Figure~\ref{subfig:infectiondatasetexample} shows the 2-hop subgraph from target node 27, which contains all nodes and edges that the model involves from node 27's point of view. The computation tree in the GAT is commonly expressed as a \textit{rooted subtree}~\cite{Sato2021randomfeature}, as shown in Figure~\ref{subfig:infectiondatasetcomputationgraph} for node 27. In the figure, the information flows from leaf nodes at depth 2 to the root node 27 at depth 0, which exhibits an apparently different structure from that of the subgraph in Figure~\ref{subfig:infectiondatasetexample}. Note that the attention weights are calculated in each graph attention layer for each edge in $\mathcal{E}$.

\paragraph{Design Principles.}\label{subsubsection:Observations} 
We begin by making several observations from the computation tree: 
\begin{enumerate}[{\bf (O1)}]
    \item Identical edges can appear multiple times in the computation tree. For example, edge $e_{40, 27}$ in Figure~\ref{subfig:infectiondatasetexample} appears twice in Figure~\ref{subfig:infectiondatasetcomputationgraph}. \label{obs:nonunique}
    \item Nodes do not appear uniformly in the computation tree. Specifically, nodes that are $k$-hops away from the target node do not exist in depth $k'$ for $0 < k' < k$ (\textit{e.g.,} node 70 appears only at depth 2 while node 40 appears three times).
    \item The graph attention layer always includes self-loops during its feed-forward process.
\end{enumerate}

Based on the above observations, we would like to state two design principles that are desirable when designing the edge attribution function $\Phi$.

\begin{enumerate}[{\bf (P1)}]
    \item Proximity effect: Edges within closer proximity to the target node tend to highly impact the model's prediction compared with distant edges, since they \textbf{are likely to appear more frequently} in the computation tree.
    \item Contribution adjustment: The contribution of an edge in the computation tree should be \textbf{adjusted} by its position ({\it i.e.}, other edges in the path towards the root).
\end{enumerate}

By these standards, we revisit Figure~\ref{subfig:infectiondatasetcomputationgraph}. We first see that edges close to the target node such as $e_{40, 27}$ appear twice, whereas distant edges such as $e_{70, 40}$ appear only once (\textbf{P1}). Moreover, we empirically show that this principle ({\bf P1}) holds in most real-world datasets (refer to Appendix~\ref{appendix:proxitmityeffectexperiment}). Additionally, for the attention weights from the last graph attention layer (\textit{i.e.,} edges connecting nodes at depth 1 to the root node), each edge tends to have roughly the value of $0.25$ for the attention weights. In consequence, the information flowing from the first graph attention layer (\textit{i.e.,} edges connecting leaf nodes to nodes at depth 1) will be diminished by roughly $0.25$ as it reaches the root node (\textbf{P2}).

\paragraph{Proposed Method.}\label{subsection:mainmethod}
To design the edge attribution function $\Phi$, we start by formally defining the computation tree alongside the {\it flow} and the {\it attention flow}.
\begin{definition}[Computation tree]\label{definition:computationgraph}
    The computation tree for an $L$-layer Att-GNN in our study is defined as a rooted subtree of height $L$ with the target node as the root node. For each node in the tree at depth $d$, the neighboring nodes and itself are at depth $d+1$ with edges directed towards node $v$. 
\end{definition}
According to Definition~\ref{definition:computationgraph}, we define the concept of flows in the computation tree.
\begin{definition}[Flow in a computation tree]\label{definition:flow}
    Given a computation tree as a rooted subtree of height $L$ with the root (target) node $v$, we define a flow $\lambda_{i,j,v}^l$ as the list of edges that sequentially appear in a path of length $l$ starting from a given edge $e_{i,j}$ within the computation tree and ending with some edge $e_{*,v}$.\footnote{As stated in~\textbf{(O\ref{obs:nonunique})}, nodes/edges are not unique in the computation tree. Nonetheless, we will use the node indices assigned from the original graph and avoid differentiating them in the computation tree as long as it does not cause any confusion.} We indicate the $k$-th position within the flow (\textit{i.e.}, starting from the bottom of the computation tree) as $\lambda_{i,j,v}^l (k)$ for $k\in [1,L]$. We denote the set of all flows in the computation tree with node $v$ at its root that starts from edge $e_{i,j}$ with length $m \in [1,L]$ as $\Lambda_v^m (e_{i,j})$. 
\end{definition}
From Definition~\ref{definition:flow}, it follows that $\lambda_{i,j,v}^l(1)=e_{i,j}$ and $\lambda_{i,j,v}^l(l)=e_{*,v}$ for all flows in $\Lambda_v(e_{i,j})$.
\begin{definition}[Attention flow in a computation tree]\label{def:attentionflow}
    Given a flow $\lambda_{v_0, v_1, w}^m = [e_{v_0, v_1}, \cdots, e_{*, w}]$ of length $m \leq L$ for an $L$-layer Att-GNN model, we define an attention flow $\alpha[\lambda_{v_0, v_1, w}^m]$ as the corresponding attention weights assigned to each edge by the associated graph attention layers: 
    \begin{equation}
    \label{eq:attentionflowdefinition}
    \alpha[\lambda_{v_0, v_1, w}^m] = [\alpha_{v_0, v_1}^{L-m+1}, \cdots, \alpha_{*, w}^{L}].
    \end{equation}
\end{definition}
Then, it follows that $\alpha[\lambda_{v_0, v_1, w}^m](i)=\alpha_{v_{i-1},v_i}^{L-m+i}$.

\begin{example}
\label{ex:flowattentionflow}
    In Figure~\ref{subfig:infectiondatasetcomputationgraph}, $\Lambda_{27}(e_{40,27})$ includes two flows, \textit{i.e.,} $\lambda_{40,27,27}^1 = [e_{40,27}]$ and $\lambda_{40,27,27}^2 = [e_{40,27}, e_{27,27}]$, along with the corresponding attention flows $\alpha[\lambda_{40,27,27}^{1}] = [0.25]$ and $\alpha[\lambda_{40,27,27}^{2}] = [0.9, 0.25]$, respectively.
\end{example}
Finally, we are ready to present \textsc{GAtt}.
\begin{definition}[\textsc{GAtt}]
\label{def:edgeattributiondefinition}
    Given a target node $v$, an edge $e_{i,j}$ of interest, the set of flows, $\Lambda_v(e_{i,j})$, and the attention flows for all flows in $\Lambda_v(e_{i,j})$, we define the edge attribution of $e_{i,j}$ in $L$-layer Att-GNN as
\begin{equation}
    \phi^v_{i,j} = \sum_{m'=1}^L \sum_{\lambda_{i,j,v}^{m'} \in \Lambda_v^{m'}(e_{i,j})} C(\alpha[\lambda_{i,j,v}^{m'}]) \alpha[\lambda_{i,j,v}^{m'}](1),\label{eq:edgeattributiondefinition}
\end{equation}
\end{definition}

\noindent where $C(\alpha[\lambda_{i,j,v}^m]) = \prod_{2\leq k \leq m}\alpha[\lambda_{i,j,v}^m](k)$ (or $1$ if $m=1$). Eq.~(\ref{eq:edgeattributiondefinition}) can be interpreted as follows. We first find all occurrences of the target edge $e_{i,j}$ in the computation tree, and then re-weight its corresponding attention score (\textit{i.e.,} $\alpha[\lambda_{i,j,v}^m](1)$) by the product of all attention weights that appear \textit{after} $e_{i,j}$ (\textit{i.e.,} $\alpha[\lambda_{i,j,v}^m](k)$ for $k \geq 2$) in the flow before the summation over all relevant flows. Next, let us turn to addressing how our design principles ({\bf P1)} and ({\bf P2)} are met. ({\bf P1}) holds as we \textit{add} the contributions from each flow rather than taking the average, therefore the total number of occurrences of $e_{i,j}$ is directly expressed in the edge attribution. \textbf{(P2)} is fulfilled by the adjustment factor $C(\alpha[\lambda_{i,j,v}^m])$, since its value is dependent on the position of $\lambda_{i,j,v}^m(1)$. Essentially, $C(\alpha[\lambda_{i,j,v}^m])$ takes the chain of calculation from an edge to the target node into account. We provide an insightful example below.

\begin{example}
    Let us recall $\lambda^{2}_{40, 27, 27} = [e_{40,27}, e_{27,27}]$ and its attention flow $\alpha[\lambda^{2}_{40, 27, 27}] = [0.9, 0.25]$ on node 27 from Example~\ref{ex:flowattentionflow}. At face value, the contribution of edge $e_{40, 27}$ within the flow $\lambda^{2}_{40, 27, 27}$ should be $0.9$. However, this is inappropriate since the information will eventually get muted significantly by $\alpha_{27, 27}^2 = 0.25$; thus, we need to consider the adjustment factor $C(\alpha[\lambda^{2}_{40, 27, 27}])$ before calculating the final edge attribution. From Definition~\ref{def:edgeattributiondefinition}, the edge attribution $\phi^{27}_{40, 27}$ from the attention weights is calculated as $\phi^{27}_{40, 27} = 1 \times 0.25 + 0.25 \times 0.9 = 0.475$.
\end{example}

\paragraph{Efficient Calculation of \textsc{GAtt}.}
Although \textsc{GAtt} is defined as Eq.~(\ref{eq:edgeattributiondefinition}), directly using this to compute the edge attribution $\phi_{i,j}^v$ is not desirable since it involves constructing the computation tree in the form of a rooted subtree for each node $v$, as well as computing over all relevant attention flows, resulting in high redundancy during computation and not being proper for batch computation. To overcome these computational challenges, as another contribution, we introduce a \textit{matrix-based computation} method that is much preferred in practice. To this end, we first define
\[
    \mathbf{C}_L(k) = 
\begin{cases}
    {\bf I},              & \text{if } k=0,\\
    {\bf A}(L){\bf A}(L-1)\cdots {\bf A}(L-k+1),& \text{otherwise.}
\end{cases}
\]
Then, we would like to establish the following proposition.
\begin{proposition}
\label{proposition:gattcomputation}    
For a given set of attention weights $\mathcal{A} = \{{\bf A}(l)\}_{l=1}^{L}$ for an $L$-layer Att-GNN with $L\geq1$, \textsc{GAtt} in Definition~\ref{def:edgeattributiondefinition} is equivalent to
\begin{equation}
\label{eq:gattmatrixcomputation}
    \phi^v_{i,j} = \sum_{m=1}^{L} [\mathbf{C}_L(L-m)]_{v, j} [{\bf A}(m)]_{j,i}.
\end{equation}
\end{proposition}
We refer to Appendix~\ref{appendix:theoremproof} for the proof. Proposition~\ref{proposition:gattcomputation} signifies that \textsc{GAtt} sums the attention scores, weighted by the sum of the products of attention weights $[{\bf A}(m)]_{j,i}$ along the paths from node $j$ to node $v$, over all graph attention layers. We also provide another \textsc{GAtt} calculation method optimized for \textit{batch calculations} in Appendix~\ref{appendix:batchcalculation}.
  
\paragraph{Complexity Analysis.}We first analyze the \textit{computational complexity} of \textsc{GAtt} with matrix-based calculation. According to Eq.~(\ref{eq:gattmatrixcomputation}), the bottleneck for calculating $\phi_{i,j}^v$ is to acquire $\prod_{k = m+1}^{L} {\bf A}(k)$. However, this matrix can be pre-computed and does not require re-calculation after its initial acquirement. Since we only count the number of multiplications in the summation, the computational complexity is finally given by $O(L)$, which is extremely efficient. Next, according to Eq.~(\ref{eq:gattmatrixcomputation}), the \textit{memory complexity} requires delving into $\mathbf{C}_{L}(L-m)$ and $\mathbf{A}(m)$. For an $L$-layer Att-GNN, while storing all attention weights in $\mathbf{A}(m)$ requires $O(L|\mathcal{E}|)$, $\mathbf{C}_{L}(L-m)$ requires at most $O(L|| T^{L-1}||_0)$, where $T$ denotes the adjacency matrix and $||\cdot||_0$ is the 0-norm. In conclusion, the total memory complexity is bounded by $O(L||T^{L-1}||_0 + L|\mathcal{E}|)$. In addition to the above theoretical findings, we empirically provide runtime evaluations, which demonstrate that \textsc{GAtt} is reasonably fast and scalable, achieving up to \textbf{58.05 times faster} runtime against PGExplainer~\cite{Luo2020PGExplainer} when calculating edge attributions for 10,000 nodes (see Appendix~\ref{subsection:SuppRuntimeResults}).

\begin{table*}[t]
\small
     \caption{Experimental results on the faithfulness measure for \textsc{GAtt}, \textsc{AvgAtt}, and random attribution for GAT/GATv2 on 7 real-world datasets. Results for 2/3-layer GAT/GATv2s are shown for each case (the best performer is highlighted as \textbf{bold}).}
    \centering
    \resizebox{\linewidth}{!}{
    \begin{tabular}{lccccccc}
    \toprule
    \multirow{2}{*}{Dataset} &  & & 2-layer GAT/GATv2 & & & 3-layer GAT/GATv2 & \\ \cmidrule{3-8}
    & &  \textsc{GAtt} & \textsc{AvgAtt}  & Random & \textsc{GAtt} & \textsc{AvgAtt}  & Random \\
    \midrule
    \multirow{3}{*}{Cora} & $\Delta_{\text{PC}}$ & \textbf{0.8468}/\textbf{0.1040} & 0.1764/0.0121 & -0.0056/-0.0036  & \textbf{0.8642}/\textbf{0.1696} & 0.0967/0.0168 & 0.0045/0.0045 \\
    & $\Delta_{\text{NE}}$ & \textbf{0.7112}/\textbf{0.0930} & 0.1526/0.0100 & -0.0076/0.0019  & \textbf{0.7690}/\textbf{0.1664} & 0.0859/0.0186 & 0.0040/0.0037 \\
    & $\Delta_{\text{P}}$ & \textbf{0.9755}/\textbf{0.9623} & 0.7251/0.6226 & 0.4389/0.4891  & \textbf{0.9875}/\textbf{0.9966} & 0.7075/0.8897 & 0.5235/0.6107 \\ \midrule
    \multirow{3}{*}{Citeseer} & $\Delta_{\text{PC}}$ & \textbf{0.8516}/\textbf{0.0658} & 0.3096/0.0180 & 0.0012/-0.0043  & \textbf{0.8711}/\textbf{0.0432} & 0.2110/0.0107 & -0.0073/-0.0034 \\
    & $\Delta_{\text{NE}}$& \textbf{0.7653}/\textbf{0.0700} & 0.2780/0.0186 & 0.0021/0.0019  & \textbf{0.8291}/\textbf{0.0551} & 0.2006/0.0140 & 0.0015/0.0025 \\
    & $\Delta_{\text{P}}$ & \textbf{0.9846}/\textbf{0.9771} & 0.9213/0.9510 & 0.3695/0.4258  & \textbf{0.9920}/\textbf{0.9961} & 0.8979/0.9692 & 0.4039/0.7569 \\ \midrule
    \multirow{3}{*}{Pubmed} & $\Delta_{\text{PC}}$ & \textbf{0.8812}/\textbf{0.0631} & 0.1648/0.0126 & -0.0064/0.0021  & \textbf{0.8489}/\textbf{0.0367} & 0.0592/0.0023 & 0.0015/-0.0016 \\
    & $\Delta_{\text{NE}}$ & \textbf{0.8201}/\textbf{0.0915} & 0.1477/0.0169 & -0.0068/0.0078  & \textbf{0.8612}/\textbf{0.0484} & 0.0600/0.0028 & 0.0009/-0.0015 \\
    & $\Delta_{\text{P}}$ & \textbf{0.9915}/\textbf{0.9972} & 0.8834/0.9361 & 0.3974/0.1327  & \textbf{0.9993}/\textbf{0.9996} & 0.8932/0.9153 & 0.5172/0.5242 \\ \midrule
    \multirow{3}{*}{Arxiv} & $\Delta_{\text{PC}}$ & \textbf{0.7790}/\textbf{0.0546} & 0.0794/-0.0593 & 0.0007/0.0028  & \textbf{0.7721}/\textbf{0.0508} & 0.0465/-0.0252 & -0.0004/-0.0003 \\
    & $\Delta_{\text{NE}}$ & \textbf{0.8287}/\textbf{0.0164} & 0.0804/-0.0390 & 0.0016/-0.0067  & \textbf{0.8282}/-0.0012 & 0.0478/-0.0216 & -0.0017/\textbf{0.0000} \\
    & $\Delta_{\text{P}}$ & \textbf{0.9908}/\textbf{0.8995} & 0.8470/0.2560 & 0.4962/0.5107  & \textbf{0.9985}/\textbf{0.9366} & 0.8331/0.3934 & 0.5004/0.5034 \\ \midrule
    \multirow{3}{*}{Cornell} & $\Delta_{\text{PC}}$ & \textbf{0.8089}/\textbf{0.2660} & 0.3391/0.0209 & -0.0284/0.0421  & \textbf{0.7173}/\textbf{0.0899} & 0.3065/-0.0512 & -0.0273/-0.0129 \\
    & $\Delta_{\text{NE}}$ & \textbf{0.7820}/\textbf{0.1526} & 0.3199/-0.0488 & -0.0231/0.0235  & \textbf{0.7160}/\textbf{0.0520} & 0.3491/-0.0294 & -0.0060/-0.0017 \\
    & $\Delta_{\text{P}}$ & \textbf{0.9532}/\textbf{0.8372} & 0.7416/0.5130 & 0.5074/0.5660  & \textbf{0.9270}/\textbf{0.6406} & 0.6907/0.3969 & 0.4787/0.4953 \\ \midrule
    \multirow{3}{*}{Texas} & $\Delta_{\text{PC}}$ & \textbf{0.7818}/\textbf{0.0801} & 0.3676/-0.0406 & -0.0762/0.0025  & \textbf{0.6866}/\textbf{0.1504} & 0.2443/0.0486 & 0.0414/0.0040 \\
    & $\Delta_{\text{NE}}$ & \textbf{0.7977}/0.1443 & 0.3809/\textbf{0.1478} & -0.0659/0.0145  & \textbf{0.6132}/\textbf{0.0896} & 0.1645/0.0579 & 0.0202/0.0149 \\
    & $\Delta_{\text{P}}$ & \textbf{0.8726}/\textbf{0.7299} & 0.6803/0.3669 & 0.4733/0.5198  & \textbf{0.9197}/\textbf{0.8195} & 0.7072/0.5565 & 0.5562/0.5426 \\ \midrule
    \multirow{3}{*}{Wisconsin} & $\Delta_{\text{PC}}$ & \textbf{0.6898}/\textbf{0.1751} & 0.2649/0.0556 & 0.0596/0.0120  & \textbf{0.7616}/0.0323 & 0.3034/0.0337 & -0.0059/\textbf{0.0407} \\
    & $\Delta_{\text{NE}}$ & \textbf{0.6421}/\textbf{0.1554} & 0.2340/0.0636 & 0.0414/0.0157  & \textbf{0.7409}/0.0243 & 0.2762/\textbf{0.0574} & -0.0010/0.0400 \\
    & $\Delta_{\text{P}}$ & \textbf{0.8985}/\textbf{0.8501} & 0.7067/0.6060 & 0.5427/0.5006  & \textbf{0.8982}/\textbf{0.7582} & 0.6906/0.3980 & 0.5119/0.5333 \\
    \bottomrule
    \end{tabular}
    }
\vskip -0.05in
\label{table:FaithfulCitationmain}
\end{table*}

\section{Can \textit{Attention} Interpret Att-GNNs?}
\label{section:GATinterpretation}
In this section, we carry out empirical studies to validate the effectiveness of \textsc{GAtt} on interpreting two representative Att-GNN models: \textbf{GAT}~\cite{Velickovic2018gat} and \textbf{GATv2}~\cite{Brody2022GATv2}, with a single-attention head. Despite only a subset of all experimental results being presented due to page limitations, we have also demonstrated that \textsc{GAtt} can be generally applied to other Att-GNNs by showing the results for another model, \textbf{SuperGAT}~\cite{Kim2021othergat} (see Appendix~\ref{subsection:SuperGAT}). Additionally, we have found that the trend in performance for \textbf{multi-head} attention is consistent with the case for single-head attention (see Appendix~\ref{subsection:SuppMultihead}). Finally, we have shown that the \textbf{regularization} during training has negligible effects on \textsc{GAtt} (see Appendix~\ref{suppl:regularization}).

\begin{table*}[t]
\small
\centering
    \caption{Experimental results on the explanation accuracy for the synthetic datasets using 3-layer GAT/GATv2s, measured in terms of the AUROC. The results for directly using attention weights as explanation are colored as red. The best and runner-up performers are marked as \textbf{bold} and \underline{underline}, respectively, for each dataset and model.}
    \resizebox{\linewidth}{!}{
    \begin{tabular}{llcccccccccc}
    \toprule
    Model & Dataset & \cellcolor{red!25} {\textsc{GAtt}} & \cellcolor{red!25} \textsc{AvgAtt} & SA & GB & IG & GNNEx & PGEx & GM & FDnX & Random \\ \midrule
    \multirow{2}{*}{GAT}  & BA-Shapes & \cellcolor{red!25} \underline{0.9591} & \cellcolor{red!25} 0.7977 & 0.9563 & 0.6231 & 0.6231 & 0.8916 & 0.8289 & 0.5316 & \textbf{0.9917} & 0.4975 \\
    & Infection & \cellcolor{red!25} \textbf{0.9976} & \cellcolor{red!25} 0.8786 & 0.8237 & 0.8949 & \underline{0.9472} & 0.9272 & 0.7173 & 0.6859 & 0.6574 & 0.4811 \\ \midrule
    \multirow{2}{*}{GATv2}  & BA-Shapes & \cellcolor{red!25} 0.9617 & \cellcolor{red!25} 0.7876 & \underline{0.9626} & 0.5260 & 0.5232 & 0.9318 & 0.5000 & 0.5123 & \textbf{0.9923} & 0.4976 \\
    & Infection & \cellcolor{red!25} \textbf{0.8628} & \cellcolor{red!25} 0.4719 & 0.7711 & 0.7250 & 0.7849 & 0.7611 & \underline{0.8178} & 0.5355 & 0.5059 & 0.5002 \\
    \bottomrule
   \end{tabular}
   }
    \label{table:Accuracymainexperiment}
\end{table*}

\subsection{Is \textit{Attention} Faithful to the GNN?}
\label{subsection:faithfulnessexperiment}

We focus primarily on one of the most important properties in evaluating the performance of an explanation method: \textit{faithfulness}, which measures how closely the attribution reflects the inner workings of the underlying model~\cite{jacovi2020faithfulnessnlp, Chrysostomou2020faithfulnessattention, Liu2022faithfulnessviolationtest, Li2022GNNXAISurvey}. Measuring the faithfulness involves 1) manipulating the input according to the attribution scores of interest and 2) observing the change in the model's response. We specify our experiment settings below.

\paragraph{Datasets.} In our experiments, we use seven citation datasets. Specifically, we use four \textit{homophilic} datasets, including Cora, Citeseer, Pubmed~\cite{Yang2016planetoid}, and one \textit{large-scale} dataset, Arxiv~\cite{Hu2020OGB}, and three \textit{heterophilic} datasets, including Cornell, Texas, and Wisconsin~\cite{Pei2020geomgcn}. We refer to Appendix~\ref{appendix:dataset} for the details including the dataset statistics.

\paragraph{Baseline Methods.} Since the analysis of edge attribution from attention in Att-GNNs has not been studied previously, we present our own baseline approaches. We first compare the proposed \textsc{GAtt} against another attention-based explanation method~\cite{Ying2019GNNExplainer, Luo2020PGExplainer, SanchezLengeling2020XAIGNNeval}, named as \textsc{AvgAtt}, which attributes each edge as the average of the attention weights over different layers and attention heads. We additionally include random attribution as another baseline (`Random'), by randomly assigning scores in $[0, 1]$ to each edge. 

\paragraph{Attention Reduction.} It is generally known that removing $e_{i,j}$ from the graph to measure its effect may cause the out-of-distribution problem~\cite{Hooker2019fidelity,Has2021OODXAI}, a common pitfall for perturbation-based approaches. To mitigate this, we opt mask the attention coefficients ({\it i.e.}, attention weights before softmax) corresponding to edge $e_{i,j}$ with zeros in the computation tree, which reduces the effect of $e_{i,j}$ without removal. Moreover, we do not mask the attention weights after softmax, which cannot occur in a normal feed-forward process of Att-GNNs since the attention distribution is not properly normalized. In other words, we only mask attention coefficients from one edge at a time, which is compared with the original response of the Att-GNN model~\cite{Tomsett2020faithfulness}.

\paragraph{Measurement.} Denoting the output probability vector of Att-GNN for node $v$ as ${\bf p}_v$ and the output probability vector after the attention reduction for $e_{i,j}$ as ${\bf p}_{v\backslash e_{i,j}}$, we measure the model's behavior from three points of view: 1) {\it decline in prediction confidence} $\Delta_\text{PC}$~\cite{Guo2017modelcalibration} defined as the decrease of the probability for the predicted label (\textit{i.e.,} $\Delta_{\text{PC}} = {\bf p}_v[k] - {\bf p}_{v\backslash e_{i,j}}[k]$, where $k = \argmax_k {\bf p}_v[k]$), 2) {\it change in negative entropy} $\Delta_\text{NE}$~\cite{Moon2020negativeentropy} defined as the increase of `smoothness' of the probability vector (\textit{i.e.,} $\Delta_{\text{NE}} = - \sum {\bf p}_{v\backslash e_{i,j}}
\log {\bf p}_{v\backslash e_{i,j}} + \sum {\bf p}_v
\log {\bf p}_v$), which also reflects the model's confidence, and 3) {\it change in prediction} $\Delta_\text{P}$~\cite{Tomsett2020faithfulness}, which observes whether $k \neq k'$, where $\argmax_k {\bf p}_v[k]$ and $\argmax_{k'} {\bf p}_{v\backslash e_{i,j}}[k']$, where ${\bf p}_v[k]$ is the $k$-th entry of ${\bf p}_v$.

\paragraph{Quantitative Analysis of Faithfulness.} We investigate the relationship between the model's output difference from attention reduction following edge attribution scores and the edge attribution scores themselves. In each dataset, we randomly select 100 nodes as target nodes $v$ and calculate the values of \textsc{GAtt} for all edges $(i,j)$ that affect the target node (\textit{i.e.}, $\phi^v_{i,j}$). We also perform attention reduction for the same edges $(i,j)$ and measure $\Delta_\text{PC}$, $\Delta_\text{NE}$, and $\Delta_\text{P}$ to observe the correlation between \textsc{GAtt} values. Specifically, we adopt the Pearson correlation for $\Delta_\text{PC}$ and $\Delta_\text{NE}$. For $\Delta_\text{P}$, we use the area under receiver operating characteristic (AUROC), basically measuring the quality of attribution scores as a predictor of whether the prediction of the target node will change after attention reduction.

Table~\ref{table:FaithfulCitationmain} summarizes the experimental results with respect to the faithfulness on the seven real-world datasets, using pre-trained 2-layer and 3-layer GAT/GATv2s with a single attention head for each dataset. The results strongly indicate that \textsc{GAtt} substantially increases the faithfulness of edge attributions of the GAT/GATv2s models, producing a more reliable attribution score compared to \textsc{AvgAtt} and random attribution. Although \textsc{AvgAtt} shows modest performance in $\Delta_\text{P}$, it performs poorly in terms of changes in confidence (\textit{i.e.,} $\Delta_{\text{PC}}$ and $\Delta_{\text{NE}}$), sometimes performing worse than random attribution. This is because \textsc{AvgAtt} does not account for the proximity effect and contribution adjustment and rather na\"ively averages the attention weights over different layers and attention heads with no context of the computation tree. As previously mentioned, we find that the performance trend for \textit{multi-head} attention is consistent with the case for single-head attention (see Appendix~\ref{subsection:SuppMultihead}). We also provide experimental results via {\it visualizations} in Appendix~\ref{subsection:SupplFullResults} by plotting histograms for $\Delta_\text{P}$, which indicates that \textsc{GAtt} successfully assesses whether the model prediction changes after attention reduction.

\subsection{Does \textit{Attention} Reveal Accurate Graph Explanations?}
\label{subsection:accuracyexperiment}

We evaluate the edge attributions of GATs in comparison with ground truth explanations. Since only the synthetic datasets are equipped with proper ground truth explanations, we only use these datasets during evaluations. 

\paragraph{Datasets.} We use the BA-shapes and Infection synthetic benchmark datasets. \textit{BA-shapes}~\cite{Ying2019GNNExplainer} attaches 80 house-shaped motifs to a base graph made from the Barab{\'a}si-Albert model with 300 nodes, where the edges included in the motif are set as the ground truth explanations. \textit{Infection benchmark}~\cite{Faber2021gnnxaidataset} generates a backbone graph from the Erd{\"o}s-R{\'e}nyi model; then, a small portion of the nodes are assigned as `infected', and the ground truth explanation is the path from an infected node to the target node. We expect that edge attributions should highlight such ground truth explanations for GATs with sufficient performance.\footnote{We refer to Appendix~\ref{appendix:dataset} for further descriptions and statistics of datasets.}

\paragraph{Baseline Methods.} In our experiments, we mainly compare among attention-based edge attribution calculation methods (\textit{i.e.,} \textsc{GAtt} and \textsc{AvgAtt}) including Random attribution. Additionally, we consider seven popular \textit{post-hoc} explanation methods: Saliency (SA)~\cite{Simonyan2013saliency}, Guided Backpropagation (GB)~\cite{Springenberg2014guidedbackprop}, Integrated Gradient (IG)~\cite{Sundararajan2017integratedgradient}, GNNExplainer (GNNEx)~\cite{Ying2019GNNExplainer}, PGExplainer (PGEx)~\cite{Luo2020PGExplainer}, GraphMask (GM)~\cite{Schlichtkrull2021graphmask}, and FastDnX (FDnX)~\cite{Pereira2023FastDnX}. We emphasize that post-hoc explanation methods are treated as a complementary tool of inherent explanations, thus belonging to a \textbf{different category}~\cite{Du2020PosthocvsInherent}. However, we include them for a more comprehensive comparison. 

\paragraph{Experimental Results.} Table~\ref{table:Accuracymainexperiment} summarizes the results on the explanation accuracy for two synthetic datasets with ground truth explanations. As in prior studies~\cite{Ying2019GNNExplainer, Luo2020PGExplainer}, we treat evaluation as a binary classification of edges, aiming to predict whether each edge belongs to ground truth explanations by using the attribution scores as probability values. In this context, we adopt the AUROC as our metric. For both datasets, we observe that \textsc{GAtt} is much superior to \textsc{AvgAtt}. Even compared to the representative post-hoc explanation methods, \textsc{GAtt} shows a surprisingly competitive performance. For Infection, \textsc{GAtt} shows the best performance, and while \textsc{GAtt} places second and third for BA-Shapes, it still achieves over $0.95$ AUROC scores. This indicates that the attention weights can inherently capture the GAT/GATv2s' behavior as long as the attribution calculation is provided by \textsc{GAtt}.

\subsection{Ablation \& Case Study}
\label{subsection:GAttsim}

\paragraph{Ablation Study.} \textsc{GAtt} in Definition~\ref{def:edgeattributiondefinition} is developed in the sense of satisfying the two design principles (\textit{i.e.}, proximity effect (\textbf{P1}) and contribution adjustment (\textbf{P2})). We now perform an ablation study to validate the effectiveness of each design element using the GAT model. To this end, we devise two variants \textsc{GAtt}$_\textsc{sim}$ and \textsc{GAtt}$_\textsc{avg}$ by simply adding all attention weights uniformly corresponding to the target edge in the computation tree and replacing the weighted summation in Eq.~(\ref{eq:edgeattributiondefinition}) with averaging to remove the effects of the proximity effect, respectively.
More specifically, \textsc{GAtt}$_\textsc{sim}$ and \textsc{GAtt}$_\textsc{avg}$ are defined as
\begin{align}
    & \sum_{m'=1}^{L} \sum_{\lambda_{i,j,v}^{m'} \in \Lambda^{m'}_v(e_{i,j})} \alpha[\lambda_{i,j,v}^m](1), \text{ and} \\
    & \dfrac{1}{|\Lambda_v(e_{i,j})|} \sum_{m'=1}^{L} \sum_{\lambda_{i,j,v}^{m'} \in \Lambda^{m'}_v(e_{i,j})} C(\alpha[\lambda_{i,j,v}^m]) \alpha[\lambda_{i,j,v}^m](1),
\end{align}
respectively. The properties of different edge attribution calculation methods are summarized in Table~\ref{table:Methodcomparison}.

\begin{table}[t]
\caption{Properties of different edge attribution methods.}\label{table:Methodcomparison}
    \resizebox{\linewidth}{!}{
    \begin{tabular}{lcccc}
    \toprule
    Method & \textsc{GAtt} & \textsc{GAtt}$_\textsc{sim}$ & \textsc{GAtt}$_\textsc{avg}$ & \textsc{AvgAtt} \\
    \midrule
    (\textbf{P1}) Proximity effect & \textcolor{Green}{\ding{52}} & \textcolor{Green}{\ding{52}} & \textcolor{red}{\ding{55}} & \textcolor{red}{\ding{55}} \\
    (\textbf{P2}) Contribution adjustment & \textcolor{Green}{\ding{52}} & \textcolor{red}{\ding{55}} & \textcolor{Green}{\ding{52}} & \textcolor{red}{\ding{55}} \\
    \bottomrule
   \end{tabular}
   }
\end{table}
\begin{table}
\caption{Performance comparison among different edge attribution calculation methods for GATs.}\label{table:ablationstudy}
    \resizebox{\linewidth}{!}{
    \begin{tabular}{llcccc}
    \toprule
    Dataset & Model & \textsc{GAtt} & \textsc{GAtt}$_\textsc{sim}$ & \textsc{GAtt}$_\textsc{avg}$ & \textsc{AvgAtt} \\
    \midrule
    \multirow{2}{*}{Cora} & 2-layer & \textbf{0.8477} & 0.7708 & 0.8109 & 0.1768 \\
     & 3-layer & \textbf{0.8624} & 0.6392 & 0.6900 & 0.0966 \\ \midrule
    \multirow{2}{*}{Citeseer} & 2-layer & \textbf{0.8516} & 0.8058 & 0.4761 & 0.3096 \\
     & 3-layer & \textbf{0.8711} & 0.6671 & 0.8202 & 0.2110 \\\midrule
    \multirow{2}{*}{Pubmed} & 2-layer & \textbf{0.8812} & 0.7683 & 0.5915 & 0.1648 \\
     & 3-layer & \textbf{0.8489} & 0.4197 & 0.8302 & 0.0592 \\
    \bottomrule
   \end{tabular}
   }
   \vskip -0.1in
\end{table}

We compare the performance among \textsc{GAtt}, \textsc{GAtt}$_\textsc{sim}$, \textsc{GAtt}$_\textsc{avg}$, and \textsc{AvgAtt} by running experiments with respect to the faithfulness on the Cora, Citeseer, and Pubmed datasets using GATs. Table~\ref{table:ablationstudy} summarizes the results of ablation by reporting the Pearson's coefficient values for $\Delta_{\text{PC}}$. We observe that \textsc{GAtt} consistently outperforms both \textsc{GAtt}$_\textsc{sim}$ and \textsc{GAtt}$_\textsc{avg}$ for all cases. In particular, we observe the performance degradation of \textsc{GAtt}$_\textsc{sim}$ is generally more severe for 3-layer GATs. This is because the effects of the contribution adjustment (\textbf{P2}) and the cardinality of $\Lambda_v(e_{i,j})$ are more significant in a 3-layer GAT since the length of each attention flow is longer and the number of flows to consider is much higher compared to the case of 2-layer GATs.

\begin{figure}[t]
\small
  \centering
  \begin{subfigure}[b]{\linewidth}
    \includegraphics[width=\linewidth]{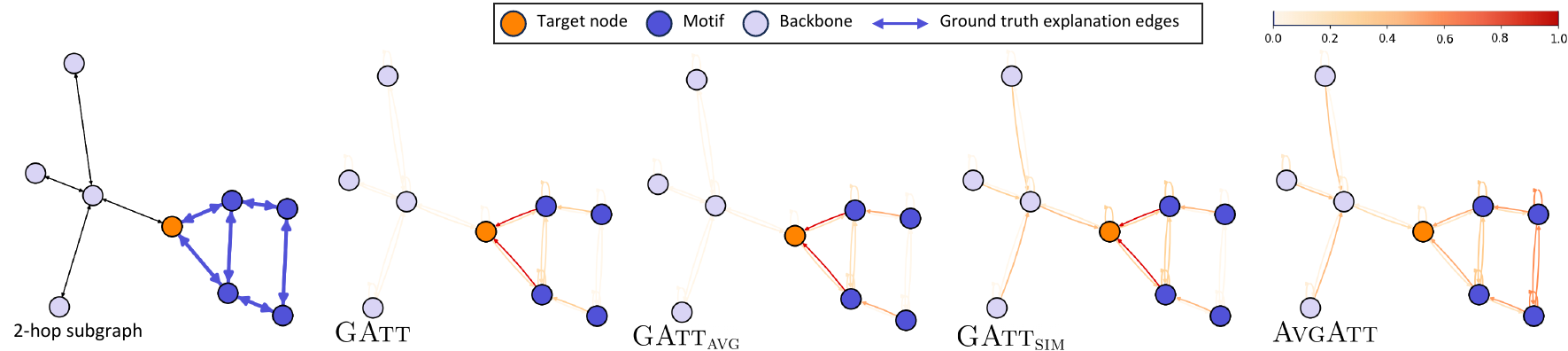}
    \caption{Node 679 on BA-Shapes.}
    \label{subfig:BAShapesCaseStudy}
  \end{subfigure}
  \begin{subfigure}[b]{\linewidth}
    \includegraphics[width=\linewidth]{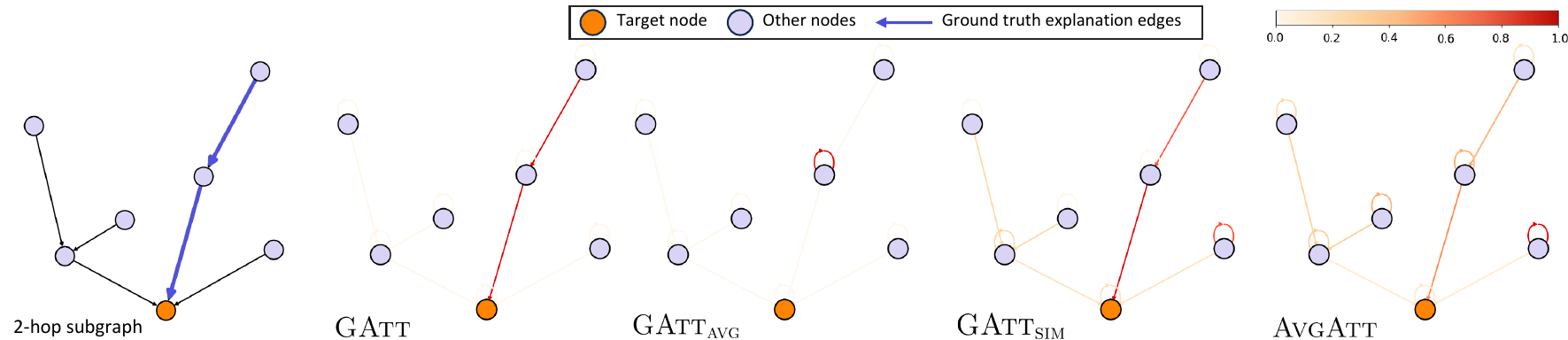}
    \caption{Node 2 on Infection.}
    \label{subfig:InfectionCaseStudy}
  \end{subfigure}
  \caption{Case study on the BA-Shapes and Infection datasets for a 2-layer GAT.}
  \label{figure:CaseStudy}
  \vskip -0.2in
\end{figure}

\paragraph{Case Study.} We conduct case studies on the BA-Shapes and Infection datasets for a 2-layer GAT, while visualizing how different methods behave. In Figure~\ref{figure:CaseStudy}, each of two cases shows a randomly selected target node (marked as orange) and the edges in ground truth explanations (blue edges). We aim to observe how much the attribution scores from \textsc{GAtt}, \textsc{GAtt}$_{\textsc{avg}}$ and \textsc{GAtt}$_{\textsc{sim}}$ focus on the ground truth explanation edges. Indeed, for both datasets, \textsc{GAtt} focuses primarily on the edges in ground truth explanations, while the attribution scores from \textsc{GAtt}$_{\textsc{sim}}$ and \textsc{AvgAtt} tend to be more spread throughout the entire 2-hop local graph. This indicates that the attention weights in the GAT indeed recognize the ground truth explanations under \textsc{GAtt} calculations. In the case of \textsc{GAtt}$_{\textsc{avg}}$, the attribution patterns are not much different from \textsc{GAtt} in BA-Shapes. However, \textsc{GAtt}$_{\textsc{avg}}$ in the Infection dataset attributes its attribution scores to a single self-loop edge that does not belong to the ground truth explanations, failing to provide adequate explanations. Interestingly, on BA-Shapes, \textsc{GAtt} tends to strongly emphasize edges that are closer in proximity even within the house-shaped motifs, which coincides with the pitfall addressed in~\cite{Faber2021gnnxaidataset}. Further extensive case studies including \textit{more target nodes} and \textit{3-layer} models exhibit a similar tendency to Figure~\ref{figure:CaseStudy} (see Appendices~\ref{subsection:SuperGAT} and~\ref{appendix:casestudies} for SuperGAT and GAT/GATv2, respectively).

\section{Conclusion and Future work}
\label{section:DiscusssionandConclusion}
In this study, we have investigated the largely underexplored problem of interpreting Att-GNNs. Although Att-GNNs were not considered as a candidate for inherently explainable models, our empirical evaluations have demonstrated affirmative results when our proposed method, \textsc{GAtt}, built upon the computation tree, can be used to effectively calculate edge attribution scores. Although \textsc{GAtt} is generally applicable, this work does not include a systematic analysis on how different designs of attention weights will interact with \textsc{GAtt}, which we leave for future work. 

\section*{Acknowledgments}
This work was supported by the National Research Foundation of Korea (NRF) grant funded by the Korea government (MSIT) (No. 2021R1A2C3004345, No. RS-2023-00220762).

\bibliography{aaai25}


\appendix
\onecolumn

\begin{center}
    {\LARGE \bf \textit{Appendix} for `Faithful and Accurate Self-Attention Attribution for Message Passing Neural Networks via the Computation Tree Viewpoint'}
\end{center}
\hfill\hfill

{\large \bf List of Additional Experiments}
\begin{itemize}
    \item \textbf{Runtime} comparison among \textbf{different calculations} of \textsc{GAtt}: Appendix~\ref{appendix:batchcalculation}
    \item Further applications of \textsc{GAtt} to other Att-GNN models (using \textbf{SuperGAT}): Appendix~\ref{subsection:SuperGAT}
    \item {\bf Comprehensive expansion} results of results from the main manuscript (GAT, GATv2, and ablation study): Appendix~\ref{subsection:fullexperimentalresults}
    \item Further experiments for \textbf{multi}-head attention: Appendix~\ref{subsection:SuppMultihead}
    \item Effect of \textbf{regularization} during training: Appendix~\ref{suppl:regularization}
    \item Experiments on \textbf{graph}-level tasks: Appendix~\ref{appendix:graphleveltask}
    \item Empirical observations on the \textbf{proximity effect}: Appendix~\ref{appendix:proxitmityeffectexperiment}
    \item \textbf{Runtime} comparison among \textbf{different explanation methods}: Appendix~\ref{subsection:SuppRuntimeResults}
    \item Further \textbf{visualizations} of faithfulness for \textsc{GAtt}, $\textsc{GAtt}_{\textsc{sim}}$, and \textsc{AvgAtt}: Appendix~\ref{subsection:SupplFullResults}
    \item Further \textbf{case studies} via visualizations: Appendices~\ref{appendix:casestudies} (for GAT and GATv2) and~\ref{subsection:SuperGAT} (for SuperGAT)
\end{itemize}

\hfill\hfill

\begin{center}
    {\large \bf Table of Notations}
\end{center}
\begin{table}[h!]
\centering
\vskip -0.1in
    \resizebox{\linewidth}{!}{
    \begin{tabular}{ll}
    \toprule
    Notation & Description \\
    \midrule
    $G$ & A given graph \\
    $\mathcal{V}$ & The set of nodes \\
    $\mathcal{E}$ & The set of edges \\
    $\mathcal{N}_i$ & The set of neighbors of node $v_i$ \\
    $L$ & The number of layers in the Att-GNN \\
    $\mathcal{A} = \{{\bf A}(l)\}_{l=1}^{L}$ & The set of attention weights in a $L$-layer Att-GNN \\
    ${\bf A}(l)$ & The attention weight matrix for the $l$-th layer in Att-GNN \\
    $\alpha_{i,j} = [{\bf A}(l)]_{j,i}$ & The attention weight of $e_{i,j}$ in the $l$-th layer in the Att-GNN \\
    $\Phi$ & The edge attribution function \\
    $\phi_{i,j}^v$ & The edge attribution score of $e_{i,j}$ for a target node $v$\\
    $\lambda_{i,j,v}^l$ & The list of edges of length $l$ starting from $e_{i,j}$ and ending with the root node $v$ within the computation tree \\
    $\Lambda_v^m (e_{i,j})$ & The set of all flows that starts from $e_{i,j}$ with length $m \in [1,L]$ for node $v$ \\
    $\alpha[\lambda_{v_0, v_1, w}^m]$ & The list of attention weights corresponding to $\lambda_{v_0, v_1, w}^m$ in the Att-GNN \\
    ${\bf p}_v$ & The output probability vector of Att-GNN for node $v$ \\
    \bottomrule
   \end{tabular}
   }
   \vskip -0.1in
\end{table}

\section{Related Work}
\label{section:relatedworks}
In this section, we discuss previous relevant studies on two major themes. Specifically, we first address previous attempts to use attention weights as interpretations in domains other than graphs, and then we provide an overview of explainable AI (XAI) for GNNs.

\paragraph{Interpreting models with attention.}
There have been a handful of studies that utilize attention as a tool for model interpretations. The term `attention' itself refers to the biological inspiration of the design where human vision also relies on a dynamic focusing on relevant parts of real-world scenery~\cite{rensink2000attention, Bibal2022attentiondebate}. This intuition is directly reflected in how the attention mechanism plays a role in machine learning models in general, namely to adjust the flow of information and to allow some parts of the input / latent representations, while the adjustment itself is a learned function of the input/latent representations~\cite{Bahdanau2014attention, Vaswani2017attention}. Aside from the performance benefits of attention, this role of attention within the model of interest makes attention a good conceptual fit to explain decisions. After all, it is much more natural for a sufficiently trained model to transfer the \textit{critical} parts of the input (to the next layer) by assigning higher attention scores rather than vice versa (which can also be applied analogously to the human case). This in turn suggests that attention can be used to discover critical parts of the input that affected the model's decisions, and therefore being an effective proxy of the model's inner mechanism.

In machine learning, the attention mechanism has been used as model interpretations since its early development, typically as an intuitive visualization tool that highlights the inner workings of the underlying model~\cite{Bahdanau2014attention, Xu2015attentionvisualization, Vig2019attentionvisualization, Dosovitskiy2021ViT16X16, Caron2021attentionvisualization}. As attention usage as interpretations become more widespread, there have been multiple studies that extensively scrutinize attention explainability in the domain of natural language processing (NLP)~\cite{Bibal2022attentiondebate}. In~\cite{JainW2019attentiondebate}, attention was found to have low correlations with other attribution methods. However,~\cite{Wiegreffe2019attentiondebate} pointed out an unfair setting in the previous work and argued that attention can still be an effective explanation. Furthermore, there have been studies that focused on post-processing attention in transformers for token attributions. \cite{Abnar2020attentionflow} proposed attention rollout and attention flow, and follow-up studies~\cite{Chefer2021attentionmapcvpr, Chefer2021attentionmapiccv, Hao2021attattr} incorporate other attribution calculation methods or gradients with attention weights. Although our study lies in a similar objective, our focus is on graph data with the Att-GNN architecture, where attention has been mostly used as a na\"ive baseline and has been largely under-explored in the literature~\cite{Ying2019GNNExplainer, Luo2020PGExplainer, SanchezLengeling2020XAIGNNeval}.

\paragraph{Explainability in GNNs.}
The primary goal of XAI is to provide a comprehensive understanding of the decision of neural networks. Early works of explaining GNNs involved applying previously developed XAI methods (\textit{e.g.}, Saliency~\cite{Simonyan2013saliency}, Integrated gradients~\cite{Sundararajan2017integratedgradient}, Guided backpropagation~\cite{Springenberg2014guidedbackprop}) directly to the underlying GNN model~\cite{Pope2019fidelity, Baldassarre2019earlygnnxai} In recent years, various studies have developed explanation methods tailored to GNN models. As one of the pioneering work, GNNExplainer~\cite{Ying2019GNNExplainer} identified a subset of edges and node features around the target node that affect the underlying model's decision. PGExplainer~\cite{Luo2020PGExplainer} trained a separate parameterized mask predictor to generate edge masks that identify important edges. However, many approaches typically necessitate an optimization/learning framework, making the explanation performance dependent on various hyperparameters ({\it e.g.}, the number of iterations and random seeds). GraphMask~\cite{Schlichtkrull2021graphmask} learns a single-layer MLP classifier to predict whether an edge in each layer can be replaced by a learned baseline vector. Finally, FastDnx~\cite{Pereira2023FastDnX} first performs knowledge distillation in a surrogate SGC model, where the explanation is eventually retrieved by solving a simple convex program. However, recent analysis demonstrated that such explanations are also known to be suboptimal, as they essentially perform a single step projection to an information-controlled space, and their high dependency on optimization and learning makes them prone to being sensitive to random seeds~\cite{Miao2022GSAT}. Although explanations of GNN models are still an active research area~\cite{Li2022GNNXAISurvey, Yuan2023GNNXAISurvey}, most studies overlooked attention as a powerful method for explanations. Several studies~\cite{Ying2019GNNExplainer, Luo2020PGExplainer, SanchezLengeling2020XAIGNNeval} have introduced GATs, a representative model in Att-GNNs, as a baseline by averaging attention over layers. In light of this, our study aims to scrutinize attention as a paramount candidate to explain Att-GNNs, and develop a simple attribution method based on attention that avoids the pitfalls of previous GNN explainability methods by being deterministic (do not require random seeds) and avoiding the usage of hyperparameters.

\section{Experimental Settings}
\subsection{Dataset Description and Statistics} 
\label{appendix:dataset}
We provide a more detailed description of the synthetic datasets used in Section~\ref{subsection:accuracyexperiment}. 

\begin{table*}[h]
\small
\centering
    \caption{Statstics for the synthetic datasets used in the experiments.}
    \resizebox{\linewidth}{!}{
    \begin{tabular}{lcccccc}
    \toprule
    Dataset & Num. nodes & Num. edges & Num. features & Num. classes & Num. motifs & Num. unique explanations\\
    \midrule
    BA-Shapes & 700 & 1,426 & 50 & 4 & 80 & - \\
    Infection (2-layer GAT) & 5,000 & 10,086  & 2 & 4 & - & 1,057 \\
    Infection (3-layer GAT) & 5,000 & 10,086  & 2 & 5 & - & 974 \\
    \bottomrule
   \end{tabular}
   }
   \label{table:SupplSyntheticDatasetStat}
\end{table*}

\begin{table*}[h]
\small
\centering
\caption{Statstics for the real-world datasets used in the experiments.}
    \begin{tabular}{lcccc}
    \toprule
    Dataset & Num. nodes & Num. edges & Num. features & Num. classes \\
    \midrule
    Cora & 2,708 & 10,556 & 1,433 & 7 \\
    Citeseer & 3,327 & 9,104 & 3.703 & 6\\
    Pubmed & 19,717 & 88,648 & 500 & 3 \\
    Arxiv & 169,343 &  1,166,243 & 128 & 40 \\
    Cornell & 183 & 295 & 1703 & 5 \\
    Texas & 183 & 309 & 1703 & 5 \\
    Wisconsin & 251 & 499 & 1703 & 5 \\
    \bottomrule
   \end{tabular}
   \vskip -0.1in
    \label{table:SupplRealWorldDatasetStat}
\end{table*}

\begin{itemize}
    \item \textit{BA-shapes}~\cite{Ying2019GNNExplainer} is a synthetic graph that attaches 80 house-shaped motifs to a base graph made from the Barab{\'a}si-Albert model with 300 nodes. The task is to classify whether a node belongs to one of four cases, including the top, middle, and bottom of a house, or the base graph. The edges belonging to the house-shaped motifs is regarded as the ground truth explanation. We use the one-hot encodings of each node's degree as the node features to ease the training.
    \item \textit{Infection benchmark}~\cite{Faber2021gnnxaidataset} is also a synthetic dataset where the backbone graph is generated from the Erd{\"o}s-R{\'e}nyi model. Then, a small portion of the nodes are assigned to `infected', where the information is encoded as a one-hot vector as a node feature (indicating `infected' versus `normal'). The node label is set as either 1) `infected' or 2) the length of the shortest path to the nearest infected node, if not infected. For an $L$-layer Att-GNN, all nodes that are more than $L$-hops away from the nearest infected node are considered as the same class. The ground truth explanation for the infection benchmark graph is the edges along the path from the nearest infected node to the target node. During evaluations, we only observe the performance of edge attribution calculation methods on nodes with a unique ground truth path (as suggested by the original authors). The number of unique explanations indicates the number of nodes having the unique shortest path to the nearest infected node.
\end{itemize}

In addition, we summarize the statistics of the synthetic and real-world datasets used in our experiments in Tables~\ref{table:SupplSyntheticDatasetStat} and~\ref{table:SupplRealWorldDatasetStat}, respectively. Cora, Citeseer, Pubmed~\cite{Yang2016planetoid} and Arxiv~\cite{Hu2020OGB} are citation network where each node represents a paper, and an edge represents a citation between two papers. These four datasets are homophilic, \textit{i.e.}, nodes that belong to the same class have a higher probability to be connected. On the other hand, Cornell, Texas, and Wisconsin~\cite{Pei2020geomgcn} is part of the WebKD dataset, which is a web page network collected from computer science departments of various
universities. For these datasets, nodes represent web pages and edges represent hyperlinks between the web pages. Also, these three datasets are heterophilic, \textit{i.e.}, nodes that belong to different classes are likely to be connected.

\subsection{Training Details of GAT/GATv2s in Experiments}

\begin{table}[h!]
\centering
\caption{Training details for different datasets used in the experiment.}\label{table:trainingdetails}
    \resizebox{\linewidth}{!}{
    \begin{tabular}{llccccccccc}
    \toprule
    & & BA-Shapes & Infection & Cora & Citeseer & Pubmed & Arxiv & Cornell & Texas & Wisconsin \\
    \midrule
   & Num. epochs & 1000 & 500 & 60 & 100 & 100 & 3000 & 65 & 32 & 50 \\
   GAT  & Hidden dim. & 16 & 8 & 64 & 64 & 64 & 64 & 64 & 64 & 64 \\ 
   (2 layer) & Learning rate & 0.01 & 0.005 & 0.001 & 0.001 & 0.001 & 0.005 & 0.005 & 0.001 & 0.0001 \\ 
   & Test acc. & 0.9500 & 0.8680 & 0.8202 & 0.7312 & 0.7294 & 0.5375 & 0.5135 & 0.5676 & 0.5098 \\ \midrule
   & Num. epochs & 1000 & 500 & 60 & 100 & 100 & 3000 & 65 & 32 & 50 \\
   GAT  & Hidden dim. & 16 & 8 & 64 & 64 & 64 & 64 & 64 & 64 & 64 \\ 
   (3 layer) & Learning rate & 0.01 & 0.005 & 0.001 & 0.001 & 0.001 & 0.005 & 0.005 & 0.001 & 0.0001 \\ 
   & Test acc. & 0.9571 & 0.9272 & 0.8362 & 0.7272 & 0.7885 & 0.5188 & 0.5135 & 0.6216 & 0.5098 \\ \midrule
   & Num. epochs & 1000 & 1000 & 60 & 60 & 60 & 3000 & 100 & 32 & 50 \\
   GATv2  & Hidden dim. & 16 & 16 & 64 & 64 & 64 & 64 & 64 & 64 & 64 \\ 
   (2 layer) & Learning rate & 0.01 & 0.01 & 0.001 & 0.001 & 0.001 & 0.005 & 0.001 & 0.001 & 0.0001 \\ 
   & Test acc. & 0.9357 & 0.9476 & 0.8292 & 0.7275 & 0.7157 & 0.5290 & 0.4865 & 0.4054 & 0.4314 \\ \midrule
   & Num. epochs & 1000 & 1000 & 60 & 60 & 60 & 3000 & 100 & 32 & 50 \\
   GATv2  & Hidden dim. & 16 & 16 & 64 & 64 & 64 & 64 & 64 & 64 & 64 \\ 
   (3 layer) & Learning rate & 0.01 & 0.01 & 0.001 & 0.001 & 0.001 & 0.005 & 0.001 & 0.001 & 0.0001 \\ 
   & Test acc. & 0.9714 & 0.9992 & 0.7644 & 0.7286 & 0.7197 & 0.5243 & 0.4054 & 0.4865 & 0.4902 \\ \midrule
   \end{tabular}
   }
\end{table}

Table~\ref{table:trainingdetails} summarizes the number of epochs, hidden dimension, learning rate, and the test accuracy for each model  ({\it i.e.}, GAT/GATv2) and dataset used in the experiments. Furthermore, we describe the detailed settings regarding data splits. For the synthetic datasets, we split the nodes into training and test sets with the ratio of 50:50. For the real-world citation datasets, we randomly select 100 nodes per class to form a training set, and consider the rest of the nodes as test sets. Finally, we use the Adam optimizer for all cases.


\subsection{Implementation Specifications} \label{subsection:implementation}
We use Python 3.10.12, with Pytorch 2.0.1, and Pytorch Geometric 2.3.1 (We use Pytorch Geometric 2.5.2 only when GraphMask~\cite{Schlichtkrull2021graphmask} is required). The experiments were run on a machine with an Intel(R) Core(TM) i7-10700K CPU @ 3.80GHz, 64GB of RAM, and a single Nvidia GeForce RTX 3080 graphics card.

\section{Batch Calculation of \textsc{GAtt}}
\label{appendix:batchcalculation}
\begin{algorithm}[h!]
    \PyComment{A(i): Attention weight matrix for the i-th layer, i = 1, ..., L} \\
    \PyComment{v: target node} \\
    \PyCode{select\_matrix = torch.zeros\_like(A(1))}\\
    \PyCode{select\_matrix[v, :] = 1} \\
    \PyCode{GAtt = select\_matrix * A(L)}\\
    \PyCode{for i in range(1, L):} \\
    \PyCode{\hskip2.0em GAtt += C(L-i)[v, :].expand\_as(A(1)).t() * A(i)} \
\caption{PyTorch-style pseudocode for batch calculation of \textsc{GAtt}}
\label{algo:batchcomputation}
\end{algorithm}
Recall that Proposition~\ref{proposition:gattcomputation} basically represents a matrix-based computation of GATT for an $L$-layer GAT. As mentioned in Section~\ref{subsection:mainmethod}, we can pre-compute ${\bf C}(1), \cdots, {\bf C}(L-1)$, which can be reused when \textsc{GAtt} is calculated for any edge $e_{i,j}$. This approach is efficient for calculating \textsc{GAtt} for a single edge, but is still inefficient in computing \textsc{GAtt} for a batch of edges. Here, we provide a PyTorch-style pseudocode for \textsc{GAtt} in Algorithm~\ref{algo:batchcomputation} that is more suited for batch computations.

Since \texttt{GAtt} in Algorithm~\ref{algo:batchcomputation} is a \texttt{torch.Tensor} object, we can acquire the attribution values according to \textsc{GAtt} simply by retrieving the values of the proper index ({\it i.e.}, $\phi^v_{i,j}=$ \texttt{GAtt[j,i]}).

\begin{figure}[h!]
     \centering
     \includegraphics[width=0.5\textwidth]{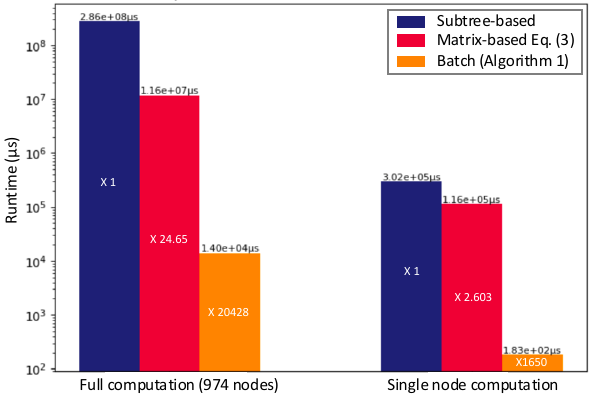}
     \caption{Runtime comparison for the Infection dataset for calculating \textsc{GAtt} using three different calculation strategies: 1) Straightforward computation via constructing a rooted subtree (denoted Subtree), 2) matrix-based computation using Eq.~(\ref{eq:gattmatrixcomputation}) in the main manuscript for each node, and 3) batch computation using Algorithm~\ref{algo:batchcomputation}. The text indicates the relative speedups with respect to the original Subtree strategy.} 
    \label{fig:appendixgattruntime}
\end{figure}

To empirically show the effectiveness of Algorithm~\ref{algo:batchcomputation} ({\it i.e.}, batch-based \textsc{GAtt} calculation), we carry out an experiment in which the runtime of calculating \textsc{GAtt} is measured for the target nodes in the Infection dataset (974 nodes in total). As shown in Figure~\ref{fig:appendixgattruntime}, the results clearly reveal that, in the batch computation strategy, using Algorithm~\ref{algo:batchcomputation} is much preferred compared to the other two calculation strategies, including calculations based on a rooted subtree and a matrix-based calculation using Eq.~(\ref{eq:gattmatrixcomputation}). For a runtime comparison of the batch computation against other edge attribution methods, we refer to Appendix~\ref{subsection:SuppRuntimeResults}.

\section{Overview of Attention}
\subsection{Self-Attention}
In the transformer model, self-attention~\cite{Vaswani2017attention} has been regarded as the key module that not only brings performance benefits but also provides interpretations of the model. In this subsection, we describe the core component of the self-attention module. 

We assume that we are given $n$ input tokens with $d'$-dimensions ${\bf T} \in \mathbb{R}^{n \times d'}$. In self-attention, we calculate the attention weights by first linearly transforming ${\bf T}$ into three different representations (of the same resulting dimensions), including the matrices of query ${\bf Q}$, key ${\bf K}$, and the value ${\bf V}$, which are defined as ${\bf Q} = {\bf TW}_{Q}$, ${\bf K} = {\bf TW}_{K}$, and ${\bf V} = {\bf TW}_{V}$,
where ${\bf W}_{Q}$, ${\bf W}_{K}$, and ${\bf W}_{V}$ are learnable parameters. Then, each query representation is compared to the key representation of all other input tokens in the same set (hence `self') to calculate how much it should attend to other parts of the input during computation. In other words, we generate a query-dependent distribution of attention scores. This process can be neatly expressed as follows:
\begin{equation}
\label{eq:originalselfattention}
    \text{SelfAttention}({\bf Q, K, V}) = \text{softmax}\left(\dfrac{{\bf QK}^\top}{\sqrt{d_a}}\right){\bf V},
\end{equation}
where $d_a$ is the dimension of the tokens after initial linear transformation, and the softmax function is applied row-wise. 

\subsection{Attention-based GNNs} \label{subsection:SuppAttGNNdescription}
\paragraph{Message passing neural networks (MPNN).} GNNs adopt the message passing mechanism as the building block of each layer, which enables GNNs to naturally encode the graph structure to the node representations. The representation ${\bf h}_v^{(l)}$ of node $v$ in the $l$-th layer of GNNs is calculated as
\begin{equation}
\label{eq:mpnn}
    {\bf h}_v^{(l+1)} = \textsc{Readout} \left({\bf h}_v^{(l)}, \bigoplus_{u \in \mathcal{N}_v} \textsc{Message}({\bf h}_v^{(l)}, {\bf h}_u^{(l)})\right),
\end{equation}
where $\textsc{Message}$ is the message function computed for each edge $e_{v,u}$, $\bigoplus$ is the permutation invariant aggregation function (\textit{e.g.,} summation), and $\textsc{Readout}$ is the readout function that takes the representation of node $v$ and the aggregated message to return a new representation. Note that the recursive design of Eq.~(\ref{eq:mpnn}) has a unique consequence on the actual computation in GNNs; for an $L$-layer GNN, the calculation for node $v$ is essentially encoding a rooted subtree of height $L$, with $v$ at its root~\cite{Morris2019WL}. A popular variant of MPNN incorporates the self-attention mechanism into the message passing mechanism. Due to the effectiveness and unique interpretation potential of attention, such attention-based MPNNs (Att-GNNs) have been separately investigated by the literature (refer to the introduction for detailed discussion). Among those models, we focus on the following three Att-GNNs as the representative architecture.

\paragraph{GAT.} GATs incorporate the concept of self-attention in message passing. The key idea is to treat the set of input tokens in the original self-attention as the hidden representations of nodes in $\mathcal{N}_i \cup \{i\}$, where $\mathcal{N}_i$ is the set of neighbors of node $i$. Specifically, the $l$-th graph attention layer of GATs is given by~\cite{Velickovic2018gat}:
\begin{equation}
\label{eq:gatconv}
    {\bf h}_i^{(l+1)} = \sum_{j \in \mathcal{N}_i \cup \{i\}} \alpha_{j,i} {\bf h}_j^{(l)},
\end{equation}
where the attention weight $\alpha_{j,i}$ is defined as $\text{softmax}(\text{LeakyReLU}({\bf a}^\top [{\bf W} {\bf h}_j || {\bf W} {\bf h}_i]))$ for learnable parameters ${\bf a}$ and ${\bf W}$. Conceptually, nodes $i$ and $j$ are considered as queries and keys, respectively, which determine the attention weights $\alpha_{j,i}$ in Eq.~(\ref{eq:gatconv}), and ${\bf h}_j^{(l)}$ is considered as values.

\paragraph{GATv2.} We also describe the attention layer in GATv2~\cite{Brody2022GATv2}. GATv2 is a modified architecture of GAT by improving upon the self-attention module to increase the model expressiveness. The difference between the two models can be summarized by the following two equations expressing the attention coefficient $e({\bf h}_i, {\bf h}_j)$:
\begin{align}
    \text{GAT: }e({\bf h}_i, {\bf h}_j) &= \text{LeakyReLU}({\bf a}^\top [{\bf W}{\bf h}_i || {\bf W}{\bf h}_j]), \\
    \text{GATv2: }e({\bf h}_i, {\bf h}_j) &= {\bf a}^\top \text{LeakyReLU}({\bf W} [{\bf h}_i || {\bf h}_j]),
\end{align}
where ${\bf a}$ and ${\bf W}$ are learnable parameters and ${\bf h}_i$ is the intermediate representation of node $i$. Despite the different self-attention mechanisms, the structure of the computation tree is essentially identical to that of GAT. Therefore, the application of \textsc{GAtt} to GATv2 is fairly straightforward.

\paragraph{SuperGAT.} There exists two advanced variants of SuperGAT~\cite{Kim2021othergat}, namely $\text{SuperGAT}_{\text{SD}}$ and $\text{SuperGAT}_{\text{MX}}$. For $\text{SuperGAT}_{\text{SD}}$, the unnormalized attention coefficient is calculated by scaling down the dot-product version of attention:
\begin{equation}
    \text{SuperGAT}_{\text{SD}}: e({\bf h}_i, {\bf h}_j) = \dfrac{1}{\sqrt{F}}({\bf W}{\bf h}_i)^\top \cdot ({\bf W}{\bf h}_j),
\end{equation}
where $F$ denotes the dimension of ${\bf h}_i$ and ${\bf h}_j$, ${\bf a}$ and ${\bf W}$ are learnable parameters, and ${\bf h}_i$ is the intermediate representation of node $i$. On the other hand, the unnormalized attention for $\text{SuperGAT}_{\text{MX}}$ is the product between the original GAT attention coefficient and the dot-product version:
\begin{equation}
    \text{SuperGAT}_{\text{MX}}: e({\bf h}_i, {\bf h}_j) = \text{LeakyReLU}({\bf a}^\top [{\bf W} {\bf h}_j || {\bf W} {\bf h}_i])) \cdot \sigma(({\bf W}{\bf h}_i)^\top \cdot ({\bf W}{\bf h}_j)),
\end{equation}
where $\sigma$ denotes the sigmoid function. Similar as to the case with GATv2, the application of \textsc{GAtt} to SuperGAT is straightforward since the structure of the computation tree is identical.

\begin{figure}[h]
  \centering
  \begin{subfigure}[b]{0.19\linewidth}
    \includegraphics[width=\linewidth]{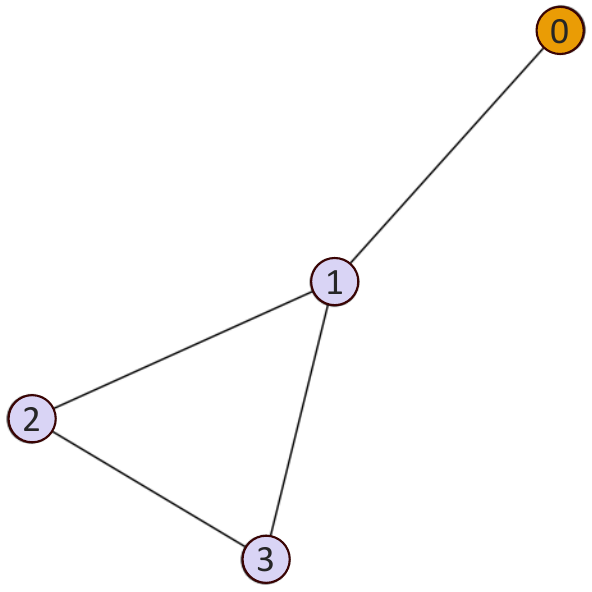}
    \caption{Example graph.}
    \label{subfig:examplegraph}
  \end{subfigure} 
  \begin{subfigure}[b]{0.54\linewidth}
    \includegraphics[width=\linewidth]{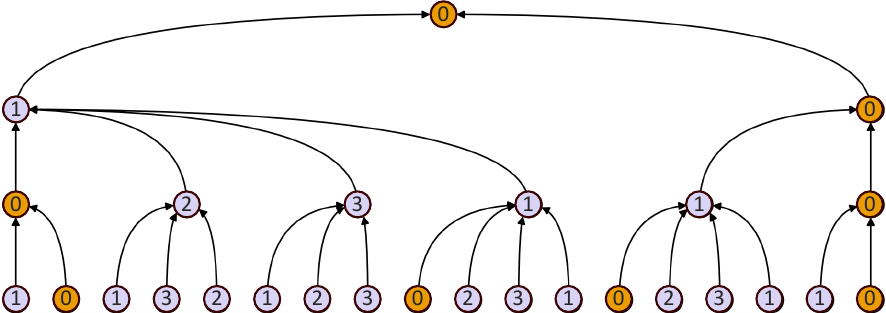}
    \caption{Computation tree of GAT.}
    \label{subfig:computationgraphGAT}
  \end{subfigure}
  \begin{subfigure}[b]{0.23\linewidth}
    \includegraphics[width=\linewidth]{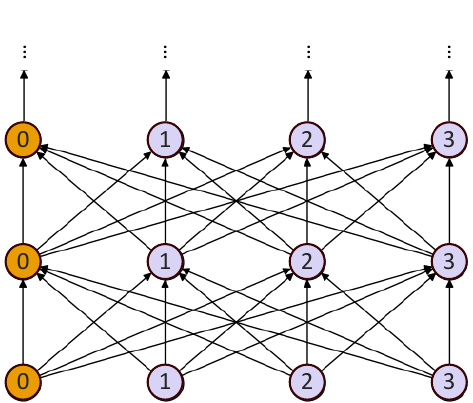}
    \caption{Computation description of (graph) transformer.}
    \label{subfig:computationgraphTransformer}
  \end{subfigure}
  \caption{A visualization of the computation trees (without color-coding the attention values) for GAT (Figure~\ref{subfig:computationgraphGAT}) and graph transformer (Figure~\ref{subfig:computationgraphTransformer}) for a given graph (Figure~\ref{subfig:examplegraph}). The target node 0 is colored in yellow.}
  \label{fig:AppComGraphComparison}
\end{figure}

\subsection{Comparison of Computation Trees Between Att-GNNs and Transformers} \label{appendix:computationcomparison}
Although both Att-GNNs and transformers employ self-attention as the core module of their architectures, the way we consider how the computation trees of Att-GNNs are taken into consideration is different from the case of how transformers are analyzed~\cite{Abnar2020attentionflow, Chefer2021attentionmapcvpr, Chefer2021attentionmapiccv}. Despite the existence of multiple transformer architectures that have been proposed for graph learning (which are also known as graph transformers)~\cite{Ying2021graphtransformers, Kreuzer2021SAN, Chen2023graphtransformer}, the self-attention module still generally follows the original one in~\cite{Vaswani2017attention}. In this subsection, we discuss the differences between how the computation tree is described for Att-GNNs and how the computation process of graph transformers is discussed during model analysis. To this end, we use a vanilla version of graph transformers where each input token represents a node in the underlying graph.

Figure~\ref{fig:AppComGraphComparison} illustrates the typical computation trees for a 3-layer GAT and the first 2 layers of a graph transformer given an example graph when the target node 0 is given. Before the analysis, we observe the following critical difference between Att-GNNs and graph transformers:
\begin{itemize}
    \item Att-GNNs handle the edges existing only in the underlying graph, while graph transformers use a fully-connected graph for the self-attention in general.
\end{itemize}
Often, due to this characteristic, a typical challenge for graph transformers is developing an effective way to inject structural information into the self-attention module as additional features~\cite{Chen2023graphtransformer}. Additionally, this is the root cause for most of the points that arise regarding the comparison between Att-GNNs and graph transformers. Specifically, we will analyze the computation tree of the (vanilla) graph transformer in Figure~\ref{subfig:computationgraphTransformer} by reflecting on our own observations in Section~\ref{subsubsection:Observations}, which we make as follows:
\begin{enumerate}[{\bf (O1)}]
    \item ``\textit{(For Att-GNNs) Identical edges can appear multiple times in the computation tree}": This is not the case when we look at each layer of the computation tree of the graph transformer, since each edge is represented once per layer. For example, the self-loop of node 0 appears once per layer in Figure~\ref{subfig:computationgraphTransformer}, while it appears twice in Figure~\ref{subfig:computationgraphGAT} at the bottom level.
    \item ``\textit{(For Att-GNNs) Nodes do not appear uniformly in the computation tree. Specifically, nodes that are $k$-hops away from the target node do not exist in depth} $k'$ for $0 < k' < k$": This is also not the case for graph transformers, since all edges are represented over all layers in the computation tree. For example, edge $e_{2,3}$ appears for all layers in Figure~\ref{subfig:computationgraphTransformer}, while it appears in Figure~\ref{subfig:computationgraphGAT} only at the bottom level for the case of GAT.
    \item ``\textit{(For Att-GNNs) The graph attention layer always includes self-loops during its feed-forward process}": This is the only point that holds for both Att-GNNs and graph transformers.
\end{enumerate}

From the analysis above, we require a different approach for the computation tree viewpoint, apart from previous studies that attempted to interpret transformer models using attention weights~\cite{JainW2019attentiondebate, Wiegreffe2019attentiondebate, Bibal2022attentiondebate, Abnar2020attentionflow, Chefer2021attentionmapcvpr, Chefer2021attentionmapiccv}.

\section{Further Experimental Results}
\label{appendix:furtherexperiments}
We include the experimental results that were not shown in the main manuscript.

\begin{figure}[h!]
  \centering
  \begin{subfigure}[b]{0.45\linewidth}
    \includegraphics[width=\linewidth]{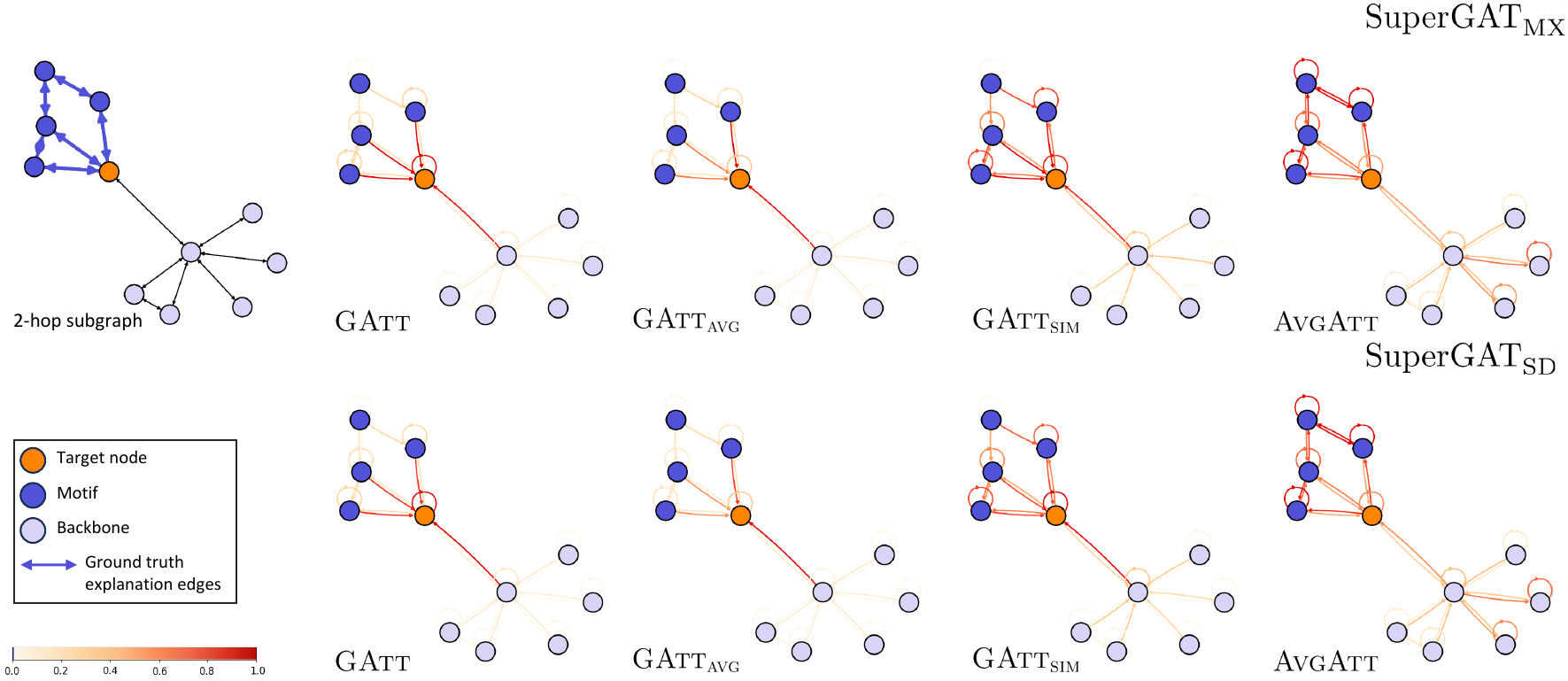}
    \caption{BA-Shapes (2-layer).}
    \label{subfig:SupplCaseStudySuperGATBAShapes2layer}
  \end{subfigure} 
  \begin{subfigure}[b]{0.45\linewidth}
    \includegraphics[width=\linewidth]{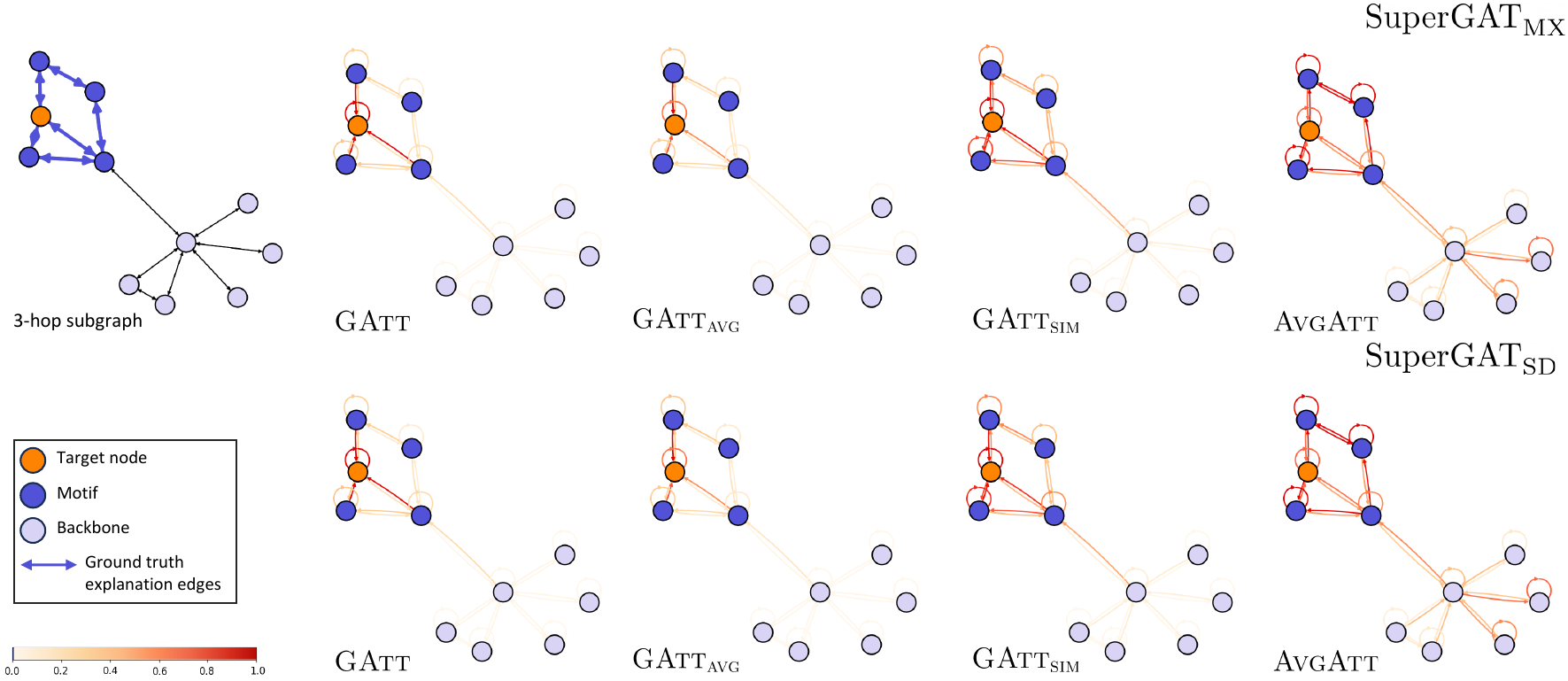}
    \caption{BA-Shapes (3-layer).}
    \label{subfig:SupplCaseStudySuperGATBAShapes3layer}
  \end{subfigure}
  \begin{subfigure}[b]{0.45\linewidth}
    \includegraphics[width=\linewidth]{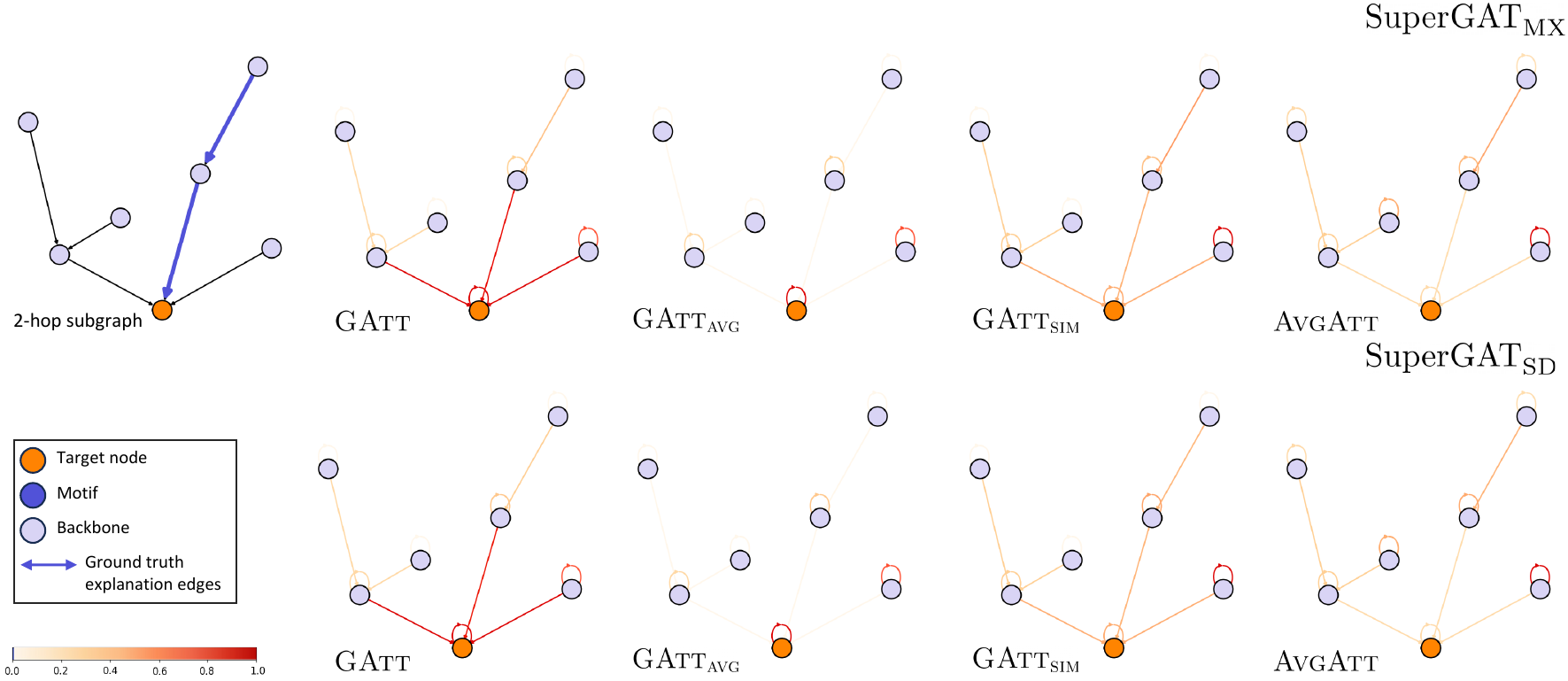}
    \caption{Infection (2-layer).}
    \label{subfig:SupplCaseStudySuperGATInfection2layer}
  \end{subfigure} 
  \begin{subfigure}[b]{0.45\linewidth}
    \includegraphics[width=\linewidth]{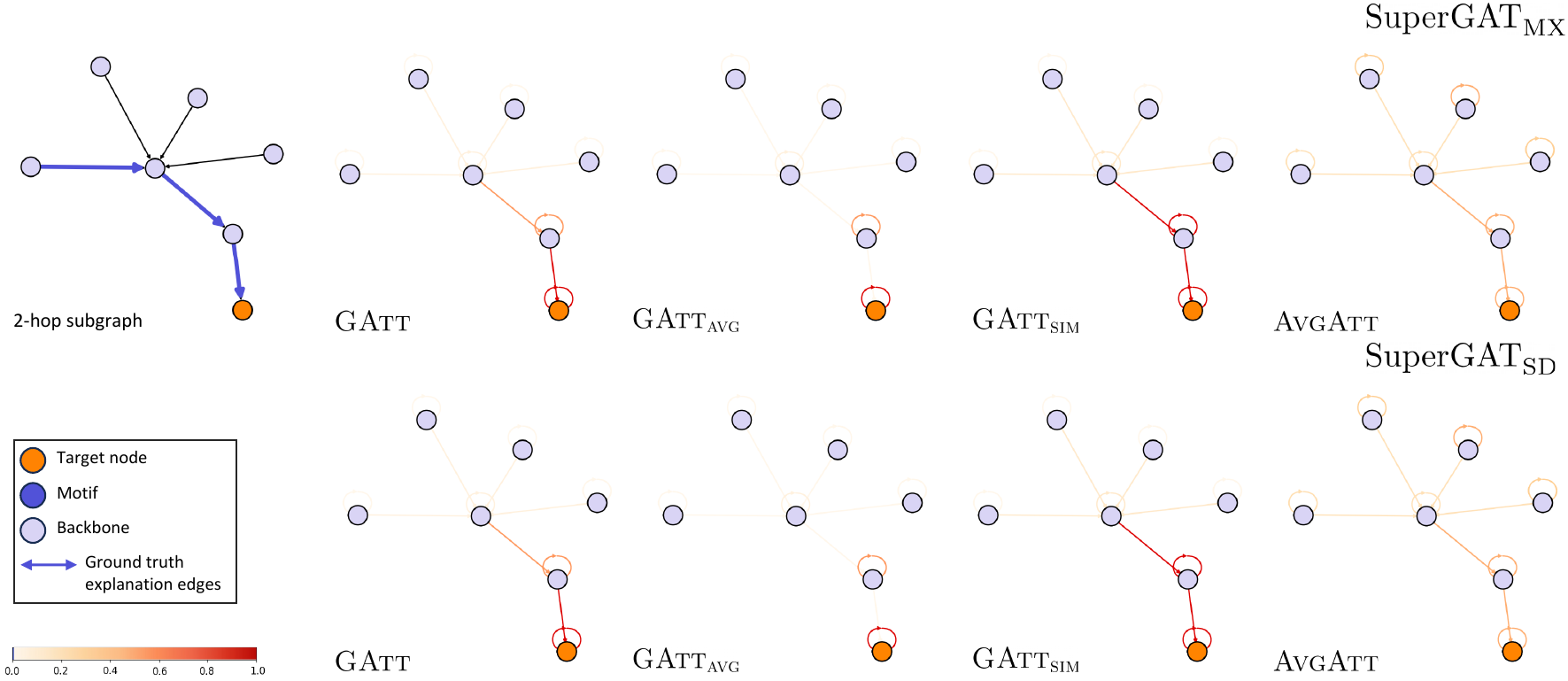}
    \caption{Infection (3-layer).}
    \label{subfig:SupplCaseStudySuperGATInfection3layer}
  \end{subfigure}
  \caption{Case studies for SuperGAT on BA-Shapes and Infection.}
  \label{fig:SupplCaseStudySuperGAT}
\end{figure}

\subsection{Further Experiments Using SuperGAT (Table~\ref{table:FaithfulSuperGAT}, Figure~\ref{fig:SupplCaseStudySuperGAT})}\label{subsection:SuperGAT}

\begin{table*}[h!]
\small
\centering
     \caption{Experimental results on both variants of SuperGAT with respect to the faithfulness for \textsc{GAtt}, \textsc{AvgAtt}, and random attribution on the Cora dataset. Results for 2-layer and 3-layer SuperGATs are shown for each case. The best performer is highlighted as \textbf{bold}.}
    \resizebox{0.8\linewidth}{!}{
    \begin{tabular}{lllcccccc}
    \toprule
    \multirow{2}{*}{Model} & \multirow{2}{*}{Measure} & \multirow{2}{*}{Metric} & & 2-layer & & & 3-layer & \\ \cmidrule{4-9}
     &  &  & \textsc{GAtt} & \textsc{AvgAtt}  & Random & \textsc{GAtt} & \textsc{AvgAtt}  & Random \\
    \midrule
    \multirow{ 7}{*}{$\text{SuperGAT}_{\text{MX}}$} & \multirow{ 3}{*}{$\Delta_{\text{PC}}$} & $\rho_{\text{Pearson}}$ ($\uparrow$) & \textbf{0.1345}  &  0.0207  & -0.0030  & \textbf{0.1209} & 0.0070 & -0.0003 \\
    & & $\tau_{\text{Kendall}}$ ($\uparrow$) & -0.0783  &  \textbf{0.0413}  & -0.0095  & -0.0330 & -0.0006 & \textbf{-0.0004} \\
    & & $\rho_{\text{Spearman}}$ ($\uparrow$) & -0.0790  &  \textbf{0.0606}  & -0.0131  & -0.0273 & -0.0008 & \textbf{-0.0006} \\
    \cmidrule{2-9}
    & \multirow{ 3}{*}{$\Delta_{\text{NE}}$} & $\rho_{\text{Pearson}}$ ($\uparrow$) & \textbf{0.1124}  &  0.0173  & -0.0009  & \textbf{0.1101} & 0.0062 & 0.0022 \\
    & & $\tau_{\text{Kendall}}$ ($\uparrow$) & -0.0823  &  \textbf{0.0510}  & 0.0011  & -0.0153 & -0.0004 & \textbf{-0.0001} \\
    & & $\rho_{\text{Spearman}}$ ($\uparrow$) & -0.0831  &  \textbf{0.0713}  & 0.0015  & -0.0108 & \textbf{0.0000} & -0.0001 \\
    \cmidrule{2-9}
    & $\Delta_{\text{P}}$ & AUROC ($\uparrow$) & \textbf{0.9808}  &  0.5794  & 0.5131  & \textbf{0.9893} & 0.6269 & 0.5299 \\
    \midrule
    \multirow{ 7}{*}{$\text{SuperGAT}_{\text{SD}}$} & \multirow{ 3}{*}{$\Delta_{\text{PC}}$} & $\rho_{\text{Pearson}}$ ($\uparrow$) & \textbf{0.1659}  &  0.0329  & 0.0017  & \textbf{0.1199} & 0.0117 & -0.0008 \\
    & & $\tau_{\text{Kendall}}$ ($\uparrow$) & -0.0294  &  \textbf{0.0404}  & 0.0078  & \textbf{0.0012} & -0.0013 & -0.0013 \\
    & & $\rho_{\text{Spearman}}$ ($\uparrow$) & -0.0275  &  \textbf{0.0562}  & 0.0110  & \textbf{0.0085} & -0.0017 & -0.0019 \\
    \cmidrule{2-9}
    & \multirow{ 3}{*}{$\Delta_{\text{NE}}$} & $\rho_{\text{Pearson}}$ ($\uparrow$) & \textbf{0.1478}  &  0.0305  & 0.0050  & \textbf{0.1094} & 0.0114 & 0.0003 \\
    & & $\tau_{\text{Kendall}}$ ($\uparrow$) & -0.0250  &  \textbf{0.0322}  & 0.0043  & \textbf{0.0018} & -0.0077 & 0.0004 \\
    & & $\rho_{\text{Spearman}}$ ($\uparrow$) & -0.0226  &  \textbf{0.0458}  & 0.0060  & \textbf{0.0118} & -0.0102 & 0.0006 \\
    \cmidrule{2-9}
    & $\Delta_{\text{P}}$ & AUROC ($\uparrow$) & \textbf{0.9820}  &  0.5692  & 0.3942  & \textbf{0.9886} & 0.6877 & 0.2744 \\
    \bottomrule
    \end{tabular}
    }
    \label{table:FaithfulSuperGAT}
\end{table*}

We perform an additional experiment by applying \textsc{GAtt} to SuperGAT~\cite{Kim2021othergat}, another widely used Att-GNN model to demonstrate the general applicability. SuperGAT offers two types of attention variants, namely $\text{SuperGAT}_{\text{SD}}$ and $\text{SuperGAT}_{\text{MX}}$, according to the ways of calculating the attention weights. Specifically, $\text{SuperGAT}_{\text{SD}}$ employs a scaled dot-product attention design, while $\text{SuperGAT}_{\text{MX}}$ utilizes both the original attention from GAT and dot-product attention (We refer to Appendix~\ref{subsection:SuppAttGNNdescription} for a more formal description). To empirically observe the performance of \textsc{GAtt}, we train 2-layer and 3-layer SuperGATs with a single attention head for both attention variants on the Cora dataset, each of which achieves the test accuracy of $>0.76$.

Table~\ref{table:FaithfulSuperGAT} summarizes the results of comparing \textsc{GAtt} with \textsc{AvgAtt} and random attribution with respect to the faithfulness on the Cora dataset. Although the correlation measures (\textit{i.e.}, $\Delta_{\text{PC}}$ and $\Delta_{\text{NE}}$) are generally weaker compared to the case of GATs, \textsc{GAtt} shows better faithfulness in terms of $\rho_{\text{Pearson}}$ and AUROC.

Furthermore, we performed additional case studies on the BA-Shapes and Infection datasets to visualize the effect of edge attribution methods on SuperGATs. Figure~\ref{fig:SupplCaseStudySuperGAT} shows visualizations of the edge attribution calculation scores using \textsc{GAtt}, \textsc{GAtt}$_\textsc{avg}$, \textsc{GAtt}$_\textsc{sim}$, and \textsc{AvgAtt} on node 600 in BA-Shapes and on node 2 in Infection. We observe similar tendencies to those of GAT and GATv2 on BA-Shapes (see Figures~\ref{fig:SupplCaseStudiesBAShapes} and \ref{fig:SupplCaseStudiesBAShapesGATv2}). However, we observe that the edge attributions for both variants of SuperGAT do not reveal the ground truth explanations as stongly as GAT or GATv2, especially on the Infection dataset. This implies that the behavior of \textsc{GAtt} is likely to be affected by the underlying model architecture itself. As mentioned in Section~\ref{section:DiscusssionandConclusion}, we leave the analysis of the interactions between \textsc{GAtt} and the design of different types of attention for future work.

\subsection{Expanded Results from the Main Manuscript (Tables~\ref{table:SuppablationstudyFull}--\ref{table:FaithfulCitationDetailAppendixGATv2})}\label{subsection:fullexperimentalresults}

We comprehensively show the experimental results of the faithfulness for GAT and GATv2 in the main manuscript by expanding Table~\ref{table:FaithfulCitationmain} by including Kendall's tau ($\tau_{\text{Kendall}}$) and Spearman's rho ($\rho_{\text{Spearman}}$) alongside Pearson's rho ($\rho_{\text{Pearson}}$ to measure $\Delta_{\text{PC}}$ and $\Delta_{\text{NE}}$. Tables~\ref{table:FaithfulCitationDetailAppendix} and~\ref{table:FaithfulCitationDetailAppendixGATv2} summarize the entire set of experimental results for GAT and GATv2, respectively. From Table~\ref{table:FaithfulCitationDetailAppendix}, it is shown that, when GAT is used, \textsc{GAtt} achieves superior performance compared to \textsc{AvgAtt} and random attribution in most cases. However, in Table~\ref{table:FaithfulCitationDetailAppendixGATv2}, we observe that the results for GATv2 are weaker for $\Delta_{\text{PC}}$ and $\Delta_{\text{NE}}$ compared to the case of GAT and are less consistent in terms of $\tau_{\text{Kendall}}$ and $\rho_{\text{Spearman}}$ for Pubmed and Arxiv datasets. This is due to the difference in the model architecture, as the attention weights in GAT are static in the sense that the ranking of the attention weights is independent of the query node. This characteristic of GAT serves as an advantage in terms of the correlation between the edge attributions and the model output, as the attention weights behave more predictively compared to GATv2. Finally, we would like to emphasize that the performance of \textsc{GAtt} is still vastly superior in terms of $\Delta_{\text{P}}$, since this is the most important measurement in practice as it predicts whether the model will actually change its prediction after attention reduction.

\begin{table}[h!]
\centering
\caption{Performance comparison among different edge attribution calculation methods for GATs.}\label{table:SuppablationstudyFull}
    \resizebox{0.5\linewidth}{!}{
    \begin{tabular}{llcccc}
    \toprule
    Dataset & Model & \textsc{GAtt} & \textsc{GAtt}$_\textsc{sim}$ & \textsc{GAtt}$_\textsc{avg}$ & \textsc{AvgAtt} \\
    \midrule
    \multirow{2}{*}{Cora} & 2-layer & \textbf{0.7793} & 0.7565 & 0.7650 & 0.0461 \\
     & 3-layer & \textbf{0.7598} & 0.6603 & 0.6940 & 0.0954 \\ \midrule
    \multirow{2}{*}{Citeseer} & 2-layer & \textbf{0.8514} & 0.8295 & 0.7317 & 0.1995 \\
     & 3-layer & \textbf{0.8019} & 0.7210 & 0.7890 & 0.1734 \\\midrule
    \multirow{2}{*}{Pubmed} & 2-layer & \textbf{0.7792} & 0.7413 & 0.6796 & 0.1158 \\
     & 3-layer & \textbf{0.7136} & 0.5727 & 0.7051 & 0.0721 \\
    \bottomrule
   \end{tabular}
   }
\end{table}

Furthermore, we also provide a more comprehensive result from the ablation study in Table~\ref{table:ablationstudy}. Table~\ref{table:SuppablationstudyFull} summarizes the results by averaging over all 7 measures and metrics (i.e., Kendall's tau ($\tau_{\text{Kendall}}$) and Spearman's rho ($\rho_{\text{Spearman}}$) for $\Delta_{\text{PC}}$ and $\Delta_{\text{NE}}$, and AUROC for $\Delta_{\text{P}}$). The results show the same performance trend, where the variants of \textsc{GAtt} and \textsc{AvgAtt} performs worse for all 2 and 3-layer cases.

\begin{table*}[h!]
\small
\centering
     \caption{Experimental results with respect to the faithfulness for \textsc{GAtt}, \textsc{AvgAtt}, and random attribution on four real-world datasets. Results for 2-layer and 3-layer GATs are shown for each case. The best performer is highlighted as \textbf{bold}.}
     \resizebox{0.8\linewidth}{!}{
    \begin{tabular}{lllcccccc}
    \toprule
    \multirow{2}{*}{Dataset} & \multirow{2}{*}{Measure} & \multirow{2}{*}{Metric} & & 2-layer GAT & & & 3-layer GAT & \\ \cmidrule{4-9}
     &  &  & \textsc{GAtt} & \textsc{AvgAtt}  & Random & \textsc{GAtt} & \textsc{AvgAtt}  & Random \\
    \midrule
    \multirow{ 7}{*}{Cora} & \multirow{ 3}{*}{$\Delta_{\text{PC}}$} & $\rho_{\text{Pearson}}$ & \textbf{0.8468}  &  0.1764  & -0.0056  & \textbf{0.8642} & 0.0967 & 0.0045 \\
    & & $\tau_{\text{Kendall}}$ & \textbf{0.7051}  & -0.1826 & 0.0082  & \textbf{0.6512} & -0.0537 & -0.0025 \\
    & & $\rho_{\text{Spearman}}$ & \textbf{0.6516} & -0.1240 & 0.0061  & \textbf{0.5679} & -0.0379 & -0.0018 \\
    \cmidrule{2-9}
\textbf{}    & \multirow{ 3}{*}{$\Delta_{\text{NE}}$} & $\rho_{\text{Pearson}}$ & \textbf{0.7112}  & 0.1526 & -0.0076 & \textbf{0.7690} & 0.0859 & 0.0040 \\
    & & $\tau_{\text{Kendall}}$ & \textbf{0.7948} &  -0.2463  & 0.0060  & \textbf{0.7616} & -0.0820 & 0.0007 \\
    & & $\rho_{\text{Spearman}}$ & \textbf{0.7371} &  -0.1736  & 0.0044 & \textbf{0.6737} & -0.0580 & 0.0005 \\
    \cmidrule{2-9}
    & $\Delta_{\text{P}}$ & AUROC & \textbf{0.9755} & 0.7251 & 0.4389 & \textbf{0.9875} & 0.7075 & 0.5235 \\
    \midrule
    \multirow{ 7}{*}{Citeseer} & \multirow{ 3}{*}{$\Delta_{\text{PC}}$} & $\rho_{\text{Pearson}}$ & \textbf{0.8516} & 0.3096 & 0.0012  & \textbf{0.8711} & 0.2110 & -0.0073 \\ 
    & & $\tau_{\text{Kendall}}$ & \textbf{0.7584} & -0.0106 & 0.0041 & \textbf{0.6456} & -0.0130 & 0.0021 \\ 
    & & $\rho_{\text{Spearman}}$ & \textbf{0.8321} & -0.0187 & 0.0057  & \textbf{0.7318} & -0.0191 & 0.0031 \\
    \cmidrule{2-9}
    & \multirow{ 3}{*}{$\Delta_{\text{NE}}$} & $\rho_{\text{Pearson}}$ & \textbf{0.7653} &  0.2780 & 0.0021 & \textbf{0.8291} & 0.2006 & -0.0058 \\
    & & $\tau_{\text{Kendall}}$ & \textbf{0.8469} &  -0.0312 & -0.0003 & \textbf{0.7235} & -0.0263 & 0.0023 \\
    & & $\rho_{\text{Spearman}}$ & \textbf{0.9206} &  -0.0517 & -0.0004 & \textbf{0.8204} & -0.0376 & 0.0032 \\
    \cmidrule{2-9}
    & $\Delta_{\text{P}}$ & AUROC & \textbf{0.9846} & 0.9213 & 0.3695 & \textbf{0.9920} & 0.8979 & 0.4039 \\
    \midrule
    \multirow{ 7}{*}{Pubmed} & \multirow{ 3}{*}{$\Delta_{\text{PC}}$} & $\rho_{\text{Pearson}}$ & \textbf{0.8812} &  0.1648 & -0.0064 & \textbf{0.8489} & 0.0592 & 0.0009 \\  
    & & $\tau_{\text{Kendall}}$ & \textbf{0.6268} &  -0.0797 & 0.0002  & \textbf{0.5349} & -0.0964 & -0.0003 \\ 
    & & $\rho_{\text{Spearman}}$ & \textbf{0.6746} &  -0.1097 & -0.0003  & \textbf{0.5946} & -0.1348 & -0.0004 \\  
    \cmidrule{2-9}
    & \multirow{ 3}{*}{$\Delta_{\text{NE}}$} & $\rho_{\text{Pearson}}$ & \textbf{0.8201}  & 0.1477 & -0.0068 & \textbf{0.8612} & 0.0600 & 0.0015 \\ 
    & & $\tau_{\text{Kendall}}$ & \textbf{0.7031} &  -0.0823 & 0.0025 & \textbf{0.5378} & -0.1138 & -0.0004 \\
    & & $\rho_{\text{Spearman}}$ & \textbf{0.7568} & -0.1133 & 0.0033 & \textbf{0.6187} & -0.1628 & -0.0006 \\ 
    \cmidrule{2-9}
    & $\Delta_{\text{P}}$ & AUROC & \textbf{0.9915} &  0.8834 & 0.3974 & \textbf{0.9993} & 0.8932 & 0.5172 \\
    \midrule
    \multirow{ 7}{*}{Arxiv} & \multirow{ 3}{*}{$\Delta_{\text{PC}}$} & $\rho_{\text{Pearson}}$ & \textbf{0.7790} & 0.0794 & 0.0007 & \textbf{0.7721} & 0.0465 & -0.0004 \\
    & & $\tau_{\text{Kendall}}$ & \textbf{0.2047} &  0.0128 & 0.0009  & \textbf{0.1327} & -0.0158 & -0.0041 \\
    & & $\rho_{\text{Spearman}}$ & \textbf{0.2590} & 0.0187 & 0.0013  & \textbf{0.1778} & -0.0246 & -0.0061 \\
    \cmidrule{2-9}
    & \multirow{ 3}{*}{$\Delta_{\text{NE}}$} & $\rho_{\text{Pearson}}$ & \textbf{0.8287}  & 0.0804 & 0.0016 & \textbf{0.8282} & 0.0478 & -0.0017 \\
    & & $\tau_{\text{Kendall}}$ & \textbf{0.2619} & 0.0053 & -0.0010 & \textbf{0.1557} & -0.0086 & -0.0038 \\
    & & $\rho_{\text{Spearman}}$ & \textbf{0.3275} & -0.0066 & -0.0015 & \textbf{0.2106} & -0.0142 & -0.0056 \\
    \cmidrule{2-9}
    & $\Delta_{\text{P}}$ & AUROC & \textbf{0.9908} &  0.8470 & 0.4962 & \textbf{0.9985} & 0.8331 & 0.5004 \\
    \midrule
    \multirow{ 7}{*}{Cornell} & \multirow{ 3}{*}{$\Delta_{\text{PC}}$} & $\rho_{\text{Pearson}}$ ($\uparrow$) & \textbf{0.8089}  &  0.3391  & -0.0284  & \textbf{0.7173} & 0.3065 & -0.0273 \\
    & & $\tau_{\text{Kendall}}$ ($\uparrow$) & \textbf{0.4750}  & 0.1753 & -0.0545  & \textbf{0.4685} & 0.2088 & -0.0451 \\
    & & $\rho_{\text{Spearman}}$ ($\uparrow$) & \textbf{0.6129}  & 0.2336 & -0.0772  & \textbf{0.6392} & 0.2943 & -0.0648 \\
    \cmidrule{2-9}
    & \multirow{ 3}{*}{$\Delta_{\text{NE}}$} & $\rho_{\text{Pearson}}$ ($\uparrow$) & \textbf{0.7820}  & 0.3199 & -0.0231  & \textbf{0.7160} & 0.3491 & -0.0060 \\
    & & $\tau_{\text{Kendall}}$ ($\uparrow$) & \textbf{0.4260}  & 0.1351 & -0.0343  & \textbf{0.4423} & 0.2168 & -0.0246 \\
    & & $\rho_{\text{Spearman}}$ ($\uparrow$) & \textbf{0.5746}  & 0.1828 & -0.0509  & \textbf{0.6034} & 0.3084 & -0.0355 \\
    \cmidrule{2-9}
    & $\Delta_{\text{P}}$ & AUROC ($\uparrow$) & \textbf{0.9532}  & 0.7416 & 0.5074  & \textbf{0.9270} & 0.6907 & 0.4787 \\
    \midrule
    \multirow{ 7}{*}{Texas} & \multirow{ 3}{*}{$\Delta_{\text{PC}}$} & $\rho_{\text{Pearson}}$ ($\uparrow$) & \textbf{0.7818}  & 0.3676 & -0.0762  & \textbf{0.6866} & 0.2443 & 0.0414 \\
    & & $\tau_{\text{Kendall}}$ ($\uparrow$) & \textbf{0.5447}  & 0.2157 & -0.0356  & \textbf{0.1700} & 0.1304 & 0.0034 \\
    & & $\rho_{\text{Spearman}}$ ($\uparrow$) & \textbf{0.6669}  & 0.3070 & -0.0516  & \textbf{0.3104} & 0.1912 & 0.0050 \\
    \cmidrule{2-9}
    & \multirow{ 3}{*}{$\Delta_{\text{NE}}$} & $\rho_{\text{Pearson}}$ ($\uparrow$) & \textbf{0.7977}  & 0.3809 & -0.0659  & \textbf{0.6132} & 0.1645 & 0.0202 \\
    & & $\tau_{\text{Kendall}}$ ($\uparrow$) & \textbf{0.6659}  & 0.2273 & -0.0488  & \textbf{0.1106} & 0.0505 & -0.0104 \\
    & & $\rho_{\text{Spearman}}$ ($\uparrow$) & \textbf{0.7940}  & 0.3226 & -0.0709  & \textbf{0.2288} & 0.0782 & -0.0153 \\
    \cmidrule{2-9}
    & $\Delta_{\text{P}}$ & AUROC ($\uparrow$) & \textbf{0.8726}  & 0.6803 & 0.4733  & \textbf{0.9197} & 0.7072 & 0.5562 \\
    \midrule
    \multirow{ 7}{*}{Wisconsin} & \multirow{ 3}{*}{$\Delta_{\text{PC}}$} & $\rho_{\text{Pearson}}$ ($\uparrow$) & \textbf{0.6898}  & 0.2649 & 0.0596  & \textbf{0.7616} & 0.3034 & -0.0059 \\
    & & $\tau_{\text{Kendall}}$ ($\uparrow$) & \textbf{0.5758}  & 0.1151 & 0.0413  & \textbf{0.5199} & 0.1211 & -0.0194 \\
    & & $\rho_{\text{Spearman}}$ ($\uparrow$) & \textbf{0.7024}  & 0.1532 & 0.0613  & \textbf{0.6452} & 0.1703 & -0.0291 \\
    \cmidrule{2-9}
    & \multirow{ 3}{*}{$\Delta_{\text{NE}}$} & $\rho_{\text{Pearson}}$ ($\uparrow$) & \textbf{0.6421 } & 0.2340 & 0.0414  & \textbf{0.7409} & 0.2762 & -0.0010 \\
    & & $\tau_{\text{Kendall}}$ ($\uparrow$) & \textbf{0.6869}  & 0.0943 & 0.0207  & \textbf{0.7551} & 0.1236 & -0.0069 \\
    & & $\rho_{\text{Spearman}}$ ($\uparrow$) & \textbf{0.8105}  & 0.1212 & 0.0314  & \textbf{0.8653} & 0.1765 & -0.0103 \\
    \cmidrule{2-9}
    & $\Delta_{\text{P}}$ & AUROC ($\uparrow$) & \textbf{0.8985}  & 0.7067 & 0.5427  & \textbf{0.8982} & 0.6906 & 0.5119 \\
    \bottomrule
    \end{tabular}
    }
    \vskip -0.1in
    \label{table:FaithfulCitationDetailAppendix}
\end{table*}

\begin{table*}[h!]
\small
\vskip -0.1in
\centering
     \caption{Experimental results with respect to the faithfulness for \textsc{GAtt}, \textsc{AvgAtt}, and random attribution on four real-world datasets. Results for 2-layer and 3-layer GATv2s are shown for each case. The best performer is highlighted as \textbf{bold}.}
     \resizebox{0.8\linewidth}{!}{
    \begin{tabular}{lllcccccc}
    \toprule
    \multirow{2}{*}{Dataset} & \multirow{2}{*}{Measure} & \multirow{2}{*}{Metric} & & 2-layer GATv2 & & & 3-layer GATv2 & \\ \cmidrule{4-9}
     &  &  & \textsc{GAtt} & \textsc{AvgAtt}  & Random & \textsc{GAtt} & \textsc{AvgAtt}  & Random \\
    \midrule
    \multirow{8}{*}{Cora} & \multirow{ 3}{*}{$\Delta_{\text{PC}}$} & $\rho_{\text{Pearson}}$ & \textbf{0.1040}  & 0.0121 & -0.0036  & \textbf{0.1696} & 0.0168 & 0.0045 \\
    & & $\tau_{\text{Kendall}}$ & \textbf{0.1128}  & -0.0632 & -0.0035  & \textbf{0.2135} & -0.0025 & -0.0049 \\
    & & $\rho_{\text{Spearman}}$ & \textbf{0.1176}  & -0.0855 & -0.0046  & \textbf{0.2343} & -0.0032 & -0.0067 \\
    \cmidrule{2-9}
    & \multirow{ 3}{*}{$\Delta_{\text{NE}}$} & $\rho_{\text{Pearson}}$ & \textbf{0.0930}  & 0.0100 & 0.0019  & \textbf{0.1664} & 0.0186 & 0.0037 \\
    & & $\tau_{\text{Kendall}}$ & \textbf{0.1247}  & -0.0624 & 0.0049  & \textbf{0.2377} & -0.0004 & -0.0037 \\
    & & $\rho_{\text{Spearman}}$ & -0.1329  & -0.0839 & \textbf{0.0064}  & \textbf{0.2568} & -0.0006 & -0.0050 \\
    \cmidrule{2-9}
    & $\Delta_{\text{P}}$ & AUROC & \textbf{0.9623}  & 0.6226 & 0.4891  & \textbf{0.9966} & 0.8897 & 0.6107 \\
    \midrule
    \multirow{8}{*}{Citeseer} & \multirow{ 3}{*}{$\Delta_{\text{PC}}$} & $\rho_{\text{Pearson}}$ & \textbf{0.0658}  & 0.0180 & -0.0043  & \textbf{0.0432} & 0.0107 & -0.0034 \\
    & & $\tau_{\text{Kendall}}$ & \textbf{0.0918}  & -0.0224 & -0.0079  & \textbf{0.1232} & -0.0267 & -0.0007 \\
    & & $\rho_{\text{Spearman}}$ & \textbf{0.0955}  & -0.0313 & -0.0108  & \textbf{0.1284} & -0.0367 & -0.0010 \\
    \cmidrule{2-9}
    & \multirow{ 3}{*}{$\Delta_{\text{NE}}$} & $\rho_{\text{Pearson}}$ & \textbf{0.0700}  & 0.0186 & 0.0019  & \textbf{0.0551} & 0.0140 & 0.0025 \\
    & & $\tau_{\text{Kendall}}$ & \textbf{0.1254}  & -0.0537 & -0.0041  & \textbf{0.1434} & -0.0268 & 0.0008 \\
    & & $\rho_{\text{Spearman}}$ & \textbf{0.1247}  & -0.0740 & -0.0056  & \textbf{0.1473} & -0.0367 & 0.0011 \\
    \cmidrule{2-9}
    & $\Delta_{\text{P}}$ & AUROC & \textbf{0.9771}  & 0.9510 & 0.4258  & \textbf{0.9961} & 0.9692 & 0.7569 \\
    \midrule
    \multirow{8}{*}{Pubmed} & \multirow{ 3}{*}{$\Delta_{\text{PC}}$} & $\rho_{\text{Pearson}}$ & \textbf{0.0631}  & 0.0126 & 0.0021  & \textbf{0.0367} & 0.0023 & -0.0016 \\
    & & $\tau_{\text{Kendall}}$ & -0.0327  & \textbf{0.0111} & -0.0043  & -0.0525 & \textbf{0.0213} & 0.0022 \\
    & & $\rho_{\text{Spearman}}$ & -0.0326  & \textbf{0.0143} & -0.0055  & -0.0540 & \textbf{0.0282} & 0.0029 \\
    \cmidrule{2-9}
    & \multirow{ 3}{*}{$\Delta_{\text{NE}}$} & $\rho_{\text{Pearson}}$ & \textbf{0.0915}  & 0.0169 & 0.0078  & \textbf{0.0484} & 0.0028 & -0.0015 \\
    & & $\tau_{\text{Kendall}}$ & -0.0238  & \textbf{0.0051} & -0.0009  & \textbf{0.0484} & 0.0028 & -0.0015 \\
    & & $\rho_{\text{Spearman}}$ & -0.0245  & \textbf{0.0065} & -0.0011  & -0.0424 & \textbf{0.0235} & 0.0036 \\
    \cmidrule{2-9}
    & $\Delta_{\text{P}}$ & AUROC & \textbf{0.9972}  & 0.9361 & 0.1327  & \textbf{0.9996} & 0.9153 & 0.5242 \\
    \midrule
    \multirow{8}{*}{Arxiv} & \multirow{ 3}{*}{$\Delta_{\text{PC}}$} & $\rho_{\text{Pearson}}$ & \textbf{0.0546}  & -0.0593 & 0.0028  & \textbf{0.0508} & -0.0252 & -0.0003 \\
    & & $\tau_{\text{Kendall}}$ & -0.0032  & \textbf{0.0341} & 0.0018  & -0.0292 & -0.0232 & \textbf{0.0000} \\
    & & $\rho_{\text{Spearman}}$ & 0.0043  & \textbf{0.0524} & 0.0026  & -0.0348 & -0.0338 & \textbf{0.0001} \\
    \cmidrule{2-9}
    & \multirow{ 3}{*}{$\Delta_{\text{NE}}$} & $\rho_{\text{Pearson}}$ & \textbf{0.0164}  & -0.0390 & -0.0067  & -0.0012 & -0.0216 & \textbf{0.0000} \\
    & & $\tau_{\text{Kendall}}$ & -0.0241  & \textbf{0.0487} & -0.0023  & -0.0204 & -0.0220 & \textbf{0.0007} \\
    & & $\rho_{\text{Spearman}}$ & -0.0217  & \textbf{0.0747} & -0.0034  & -0.0257 & -0.0348 & \textbf{0.0011} \\
    \cmidrule{2-9}
    & $\Delta_{\text{P}}$ & AUROC & \textbf{0.8995}  & 0.2560 & 0.5107  & \textbf{0.9366} & 0.3934 & 0.5034 \\
    \midrule
    \multirow{ 7}{*}{Cornell} & \multirow{ 3}{*}{$\Delta_{\text{PC}}$} & $\rho_{\text{Pearson}}$ ($\uparrow$) & \textbf{0.2660}  &  0.0209  & 0.0421  & \textbf{0.0899} & -0.0512 & -0.0129 \\
    & & $\tau_{\text{Kendall}}$ ($\uparrow$) & \textbf{0.1813}  &  -0.0366  & -0.0366  & \textbf{0.0786} & -0.0475 & 0.0071 \\
    & & $\rho_{\text{Spearman}}$ ($\uparrow$) & \textbf{0.2859}  &  -0.0565  & -0.0227  & \textbf{0.1102} & -0.0693 & 0.0098 \\
    \cmidrule{2-9}
    & \multirow{ 3}{*}{$\Delta_{\text{NE}}$} & $\rho_{\text{Pearson}}$ ($\uparrow$) & \textbf{0.1526}  &  -0.0488  & 0.0235  & \textbf{0.0520} & -0.0254 & -0.0017 \\
    & & $\tau_{\text{Kendall}}$ ($\uparrow$) & \textbf{0.1584}  &  -0.0400  & -0.0130  & \textbf{0.0283} & -0.0338 & 0.0158 \\
    & & $\rho_{\text{Spearman}}$ ($\uparrow$) & \textbf{0.2536}  &  -0.0613  & -0.0197  & \textbf{0.0389} & -0.0462 & 0.0235 \\
    \cmidrule{2-9}
    & $\Delta_{\text{P}}$ & AUROC ($\uparrow$) & \textbf{0.8372}  &  0.5130  & 0.5660  & \textbf{0.6406} & 0.3969 & 0.4953 \\
    \midrule
    \multirow{ 7}{*}{Texas} & \multirow{ 3}{*}{$\Delta_{\text{PC}}$} & $\rho_{\text{Pearson}}$ ($\uparrow$) & \textbf{0.0801}  & -0.0406  & 0.0025  & \textbf{0.1504} & 0.0486 & 0.0040 \\
    & & $\tau_{\text{Kendall}}$ ($\uparrow$) & \textbf{0.0962}  &  0.0345  & 0.0050  & 0.0203 & \textbf{0.0322} & -0.0396 \\
    & & $\rho_{\text{Spearman}}$ ($\uparrow$) & \textbf{0.1300}  &  0.0503  & 0.0077  & 0.0354 & \textbf{0.0522} & -0.0579 \\
    \cmidrule{2-9}
    & \multirow{ 3}{*}{$\Delta_{\text{NE}}$} & $\rho_{\text{Pearson}}$ ($\uparrow$) & 0.1443  &  \textbf{0.1478}  & 0.0145  & \textbf{0.0896} & 0.0579 & 0.0149 \\
    & & $\tau_{\text{Kendall}}$ ($\uparrow$) & \textbf{0.0914}  &  0.0845  & 0.0132  & -0.0208 & \textbf{0.0372} & -0.0366 \\
    & & $\rho_{\text{Spearman}}$ ($\uparrow$) & \textbf{0.1357}  &  0.1240  & 0.0192  & -0.0146 & \textbf{0.0607} & -0.0540 \\
    \cmidrule{2-9}
    & $\Delta_{\text{P}}$ & AUROC ($\uparrow$) & \textbf{0.7299}  &  0.3669  & 0.5198  & \textbf{0.8195} & 0.5565 & 0.5426 \\
    \midrule
    \multirow{ 7}{*}{Wisconsin} & \multirow{ 3}{*}{$\Delta_{\text{PC}}$} & $\rho_{\text{Pearson}}$ ($\uparrow$) & \textbf{0.1751}  &  0.0556  & 0.0120  & 0.0323 & 0.0337 & \textbf{0.0407} \\
    & & $\tau_{\text{Kendall}}$ ($\uparrow$) & 0.0022  &  \textbf{0.0585}  & 0.0072  & \textbf{0.0286} & -0.0103 & -0.0078 \\
    & & $\rho_{\text{Spearman}}$ ($\uparrow$) & 0.0204  &  \textbf{0.0918}  & 0.0105  & \textbf{0.0319} & -0.0180 & -0.0112 \\
    \cmidrule{2-9}
    & \multirow{ 3}{*}{$\Delta_{\text{NE}}$} & $\rho_{\text{Pearson}}$ ($\uparrow$) & \textbf{0.1554}  &  0.0636  & 0.0157  & 0.0243 & \textbf{0.0574} & 0.0400 \\
    & & $\tau_{\text{Kendall}}$ ($\uparrow$) & 0.0028  &  \textbf{0.0483}  & 0.0189  & 0.0137 & \textbf{0.0160} & -0.0026 \\
    & & $\rho_{\text{Spearman}}$ ($\uparrow$) & 0.0140  &  \textbf{0.0736}  & 0.0272  & 0.0117 & \textbf{0.0218} & -0.0037 \\
    \cmidrule{2-9}
    & $\Delta_{\text{P}}$ & AUROC ($\uparrow$) & \textbf{0.8501}  &  0.6060  & 0.5006  & \textbf{0.7582} & 0.3980 & 0.5333 \\
    \bottomrule
    \end{tabular}
    }
    \vskip -0.1in
    \label{table:FaithfulCitationDetailAppendixGATv2}
\end{table*}

\clearpage

\subsection{Further Experiments for Multi-head Attention (Tables~\ref{table:FaithfulCitationmultihead}--\ref{table:AttentionHeadAanlysis})}\label{subsection:SuppMultihead}
We perform additional analysis on the effect of multi-head attention, which allows each attention head to potentially capture different patterns. Using GAT as the representative model, as a direct extension of \textsc{GAtt} to multi-head scenarios, we first consider the case where we average attention weights across attention heads before applying \textsc{GAtt} in the first experiment, which still reveals the superiority of \textsc{GAtt} over \textsc{AvgAtt} and random attribution. In the second experiment, we consider the effect of individual attention heads by selecting a single attention head per layer and observing how much such selection affects the performance of \textsc{GAtt}, which we found that swapping individual attention heads does not essentially play a significant role.

\paragraph{Performance of \textsc{GAtt} according to the number of attention heads.} We carry out an experiment while increasing the number of attention heads from 1 to 8. 

\begin{table}[h]
\centering
\caption{Experimental results with respect to the faithfulness by increasing the number of attention heads from 1 to 8 for 2 and 3-layer GAT models. The table shows the case for the Cora, Citeseer, and Pubmed datasets, measuring $\Delta_{\text{PC}}$ with $\rho_\text{Spearman}$.}\label{table:FaithfulCitationmultihead}
\vskip 0.1in
    \resizebox{\linewidth}{!}{
    \begin{tabular}{lllccccllcccc}
    \toprule
     \multirow{2}{*}{Dataset} & \multirow{2}{*}{Model} & \multirow{2}{*}{Method} & \multicolumn{4}{c}{Number of heads} & \multirow{2}{*}{Model} & \multirow{2}{*}{Method} & \multicolumn{4}{c}{Number of heads} \\ \cmidrule{4-7}\cmidrule{10-13}
     &  &  & 1 & 2 & 4 & 8 &  &  & 1 & 2 & 4 & 8 \\
    \midrule
    \multirow{3}{*}{Cora} & \multirow{3}{*}{2-layer} & \textsc{GAtt} & \textbf{0.8477} & \textbf{0.8625} & \textbf{0.8468} & \textbf{0.8496} & \multirow{3}{*}{3-layer} & \textsc{GAtt} & \textbf{0.8624} & \textbf{0.8857} & \textbf{0.8674} & \textbf{0.7048}\\
    & & \textsc{AvgAtt} & 0.1768 & 0.1807 & 0.1697 & 0.1728 & & \textsc{AvgAtt} & 0.0966 & 0.0965  & 0.0994 & 0.0857 \\
    & & Random & -0.0079 & 0.0031 & 0.0021 & 0.0073 & & Random & 0.0092 & -0.0011 & 0.0001 & -0.0033 \\
    \midrule
    \multirow{3}{*}{Citeseer} & \multirow{3}{*}{2-layer} & \textsc{GAtt} & \textbf{0.8516} & \textbf{0.8441} & \textbf{0.8316} & \textbf{0.8537} & \multirow{3}{*}{3-layer} & \textsc{GAtt} & \textbf{0.8711} & \textbf{0.8545} & \textbf{0.8082} & \textbf{0.6636}\\
    & & \textsc{AvgAtt} & 0.3096 & 0.3049 & 0.2946 & 0.3100 & & \textsc{AvgAtt} & 0.2110 & 0.2069 & 0.1988 & 0.1627 \\
    & & Random & 0.0012 & -0.0087 & 0.0011 & -0.0044 & & Random & -0.0073 & -0.0070 & 0.0041 & 0.0019 \\
    \midrule
    \multirow{3}{*}{Pubmed} & \multirow{3}{*}{2-layer} & \textsc{GAtt} & \textbf{0.8812} & \textbf{0.8661} & \textbf{0.8615} & \textbf{0.8199} & \multirow{3}{*}{3-layer} & \textsc{GAtt} & \textbf{0.8489} & \textbf{0.8237} & \textbf{0.7072} & \textbf{0.5982} \\
    & & \textsc{AvgAtt} & 0.1648 & 0.1643 & 0.1644 & 0.1550 & & \textsc{AvgAtt} & 0.0592 & 0.0580 & 0.0500 & 0.0415 \\
    & & Random & -0.0064 & 0.0003 & -0.0034 & 0.0009 & & Random  & 0.0009 & -0.0019 & -0.0027 & 0.0004 \\
    \bottomrule
   \end{tabular}
   }
\end{table} 

Table~\ref{table:FaithfulCitationmultihead} summarizes the performance of edge attribution calculation methods according to different attention head configurations. For the Cora dataset, we find that the performance trend is largely consistent with our previous discussions in Section~\ref{subsection:faithfulnessexperiment} even with an increased number of attention heads. In the case of other datasets, the results show a declining trend in terms of faithfulness when we increase the number of heads for all cases except the 2-layer GAT on Citeseer. This can be attributed to the phenomenon in which, although increasing the number of heads leads to an increase in the complexity (and eventually the capacity) of the model, not all attention heads are necessarily useful, as was also observed in transformer architectures~\cite{Michel2019attheads}. Similarly, we expect that GATs are capable of learning whether to ignore certain attention heads if necessary, which may explain the discrepancy between the increase in model performance (\textit{i.e.}, test accuracy) and the decrease in faithfulness when using more attention heads.

\paragraph{Analysis of the difference \textit{among} attention heads.} We further conduct additional analysis regarding the effect of multiple attention heads in each layer on \textsc{GAtt}. To this end, we use the 3-layer GAT model with 4 attention heads and observe whether only using a single head in each layer out of 4 heads results in a rather different model behavior, essentially making multiple surrogate GAT models with 1 attention head.

\begin{table}[h]
\small
\centering
    \caption{Experimental results on the standard deviations of $\Delta_{\text{PC}}$ (measured with Pearson's rho $\rho_\text{Pearson}$) by selecting only one attention for each layer out of 4 heads.} 
    \begin{tabular}{lccc}
    \toprule
    Dataset & Cora & Citeseer & Pubmed \\ \midrule
    Standard deviation & 0.0003 & 0.0002 & 0.0002\\
    \bottomrule
   \end{tabular}
    \label{table:AttentionHeadAanlysis}
\end{table}

Table~\ref{table:AttentionHeadAanlysis} above summarizes standard deviations of faithfulness scores ($\Delta_{\text{PC}}$ with the Pearson's rho $\rho_\text{Pearson}$) for the different surrogate GAT models on Cora, Citeseer, and Pubmed datasets. Surprisingly, we observe that the use of different combinations of attention heads does not show a significant difference from each other, as shown in the extremely low standard deviations. Therefore, we conclude that the attention scores themselves are highly likely to be similar for different heads.

\begin{figure}[h]
  \centering
  \begin{subfigure}[b]{0.3\linewidth}
    \includegraphics[width=\linewidth]{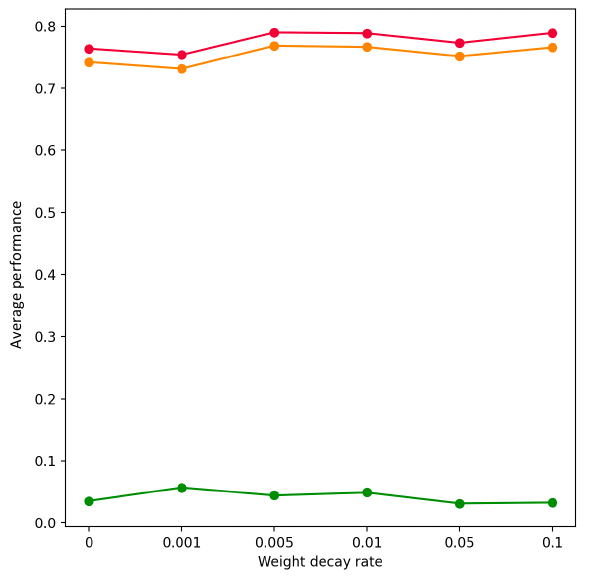}
    \caption{Weight decay.}
    \label{subfig:SupplWeightdecay}
  \end{subfigure} 
  \begin{subfigure}[b]{0.3\linewidth}
    \includegraphics[width=\linewidth]{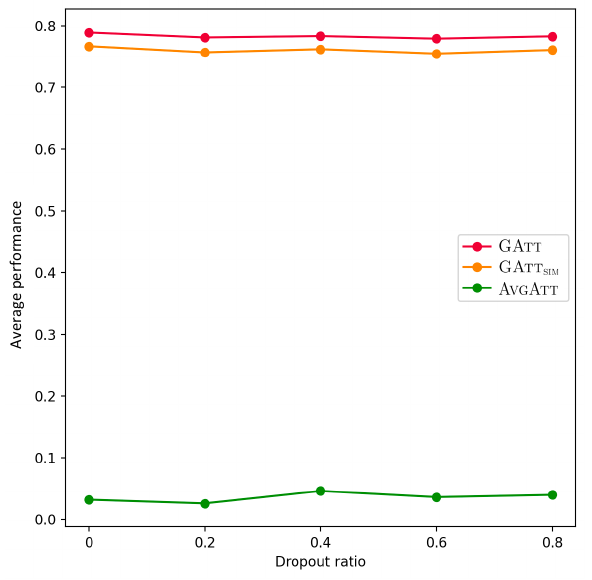}
    \caption{Dropout.}
    \label{subfig:SupplDropout}
  \end{subfigure}
  \caption{Experiment results for the effect of regularization on the faithfulness of GATs. Figures~\ref{subfig:SupplWeightdecay} and~\ref{subfig:SupplDropout} show the effect of weight decay and dropout, respectively, for the Cora dataset. For each regularization case, we report the average faithfulness over all 7 measurements ($\rho_{\text{Pearson}}, \tau_{\text{Kendall}}, \text{and } \rho_{\text{Spearman}}$ for $\Delta_{\text{PC}}$ and $\Delta_{\text{NE}}$ each, and AUROC for $\Delta_\text{P}$).}
  \label{fig:SupplRegularization}
\end{figure}

\subsection{Effect of Regularization During Training (Figure~\ref{fig:SupplRegularization})}\label{suppl:regularization}

We empirically analyze the effects of regularization during training of Att-GNNs on the faithfulness measure, using GAT as a representative model. We choose two popular regularization techniques, \textit{i.e.,} weight decay and dropout, and observe how the faithfulness of attention weights vary according to different experimental settings. We choose the following range of regularization strengths in our experiments:
\begin{itemize}
    \item Weight decay: 0, 0.0005, 0.001, 0.005, 0.01;
    \item Dropout: 0, 0.2, 0.4, 0.6, 0.8.
\end{itemize}

Figure~\ref{fig:SupplRegularization} plots the results showing the effect of different regularization strengths on the faithfulness for the Cora dataset. For both Figures~\ref{subfig:SupplWeightdecay} and \ref{subfig:SupplDropout}, the result clearly reveals that the level of regularization, {\it i.e.}, weight decay or dropout, does not fundamentally alter the performance on faithfulness, which in turn indicates that the attribution scores are robust to different degrees of regularization.

\subsection{Experiments on Graph-Level Tasks (Table~\ref{table:SupplGraphLevelExperiment})}
\label{appendix:graphleveltask}

Although we have focused mainly on node-level tasks in our main experiments, we extend \textsc{GAtt} on \textit{graph-level} tasks and analyze its performance. As a representative model and task, we aim at explaining a graph classification model built upon the GAT architecture. We train a 3-layer GAT with a summation readout function, a typical choice architecture for graph classification. Since the main idea of \textsc{GAtt} is to take the computation tree's point of view during edge attribution, an extension of \textsc{GAtt} to the graph classification problem is quite straightforward. Since the readout function takes the representation vectors for each node and then adds them, we simply follow the same mechanism by adding all the edge attribution scores resulting from setting each node as the target node. In other words, we extend the computation tree to include the readout function. We emphasize that the idea of extending the attention attribution in the existence of an additional module \textit{by following the module's computation process} is a \textit{natural} extension of our methodology~\cite{brocki2023cdam}. However, compared to~\cite{brocki2023cdam}, we directly include the additional module (in this case, the readout function) as part of the computation tree, rather than using its gradients. In other words, the edge attribution function of interest now becomes:
\begin{equation}
    \Phi(\mathcal{A}, e_{i,j}) \triangleq \textsc{Readout}(\phi^v_{i,j}),
\end{equation}
where $\textsc{Readout}(\phi^v_{i,j}) = \sum_{v \in \mathcal{V}} \phi^v_{i,j}$, following the original readout function.

\begin{table*}[h]
\small
\centering
    \caption{Experimental results with respect to the faithfulness on two graph-level datasets by measuring $\Delta_\text{P}$ with AUROC.}
    \begin{tabular}{lccc}
    \toprule
    Dataset & \textsc{GAtt} & \textsc{AvgAtt} & Random \\
    \midrule
    MUTAG & \textbf{0.8692} & 0.4725 & 0.4939 \\
    HIN & \textbf{0.6013} & 0.5740  & 0.4907 \\
    \bottomrule
   \end{tabular}
   \label{table:SupplGraphLevelExperiment}
\end{table*}

In our experiments, we adopt the following two benchmark datasets for graph classification:
\begin{itemize}
    \item \textbf{MUTAG}~\cite{debnath1991mutag} is a real-world chemical dataset where each graph represents a molecule and the atom types are one-hot encoded as node features. Each graph is labeled according to its mutagenic effect, \textit{i.e.,} it is either mutagenic or non-mutagenic.
    \item \textbf{HIN}~\cite{Azzolin2023HINdataset} is a real-world social dataset where each graph is a third-order ego graph for a larger interaction network in a hospital. The node features represents the type of each person, which can be either doctors, nurses, patients, and administrators. The task is to classify whether the ego node is a doctor or a nurse by only looking at its connections with other nodes in the graph. 
\end{itemize}

Table~\ref{table:SupplGraphLevelExperiment} summarizes the performance in faithfulness using two datasets for graph classification. The results demonstrate that \textsc{GAtt} consistently outperforms other two methods even in the graph-level setting. We also observe that \textsc{AvgAtt} is often worse than the case of random attribution, indicating that the na\"ive approach is even more detrimental in graph-level tasks.

\subsection{Empirical Observations on the Proximity Effect (Table~\ref{table:ProximityEffectExperiment})}\label{appendix:proxitmityeffectexperiment}

We further conduct empirical observations on the validity of the proximity effect ({\it i.e.}, our design principle ({\bf P1})) in Section~\ref{subsubsection:Observations}. Specifically, we have run an additional experiment in which, given a graph, we measure the correlation between the edge proximity and the number of appearances in the computation tree.

\begin{table}[h!]
\small
\centering
    \caption{Correlations between edge proximity and the number of appearances in the computation tree.} 
    \resizebox{\textwidth}{!}{
    \begin{tabular}{lccccccccccc}
    \toprule
    Dataset & BA-Shapes & Infection & Cora & Citeseer & Pubmed & Arxiv & MUTAG & HIN & Cornell & Texas & Wisconsin \\ \midrule
    Spearman's rho & -0.7122 & -0.6224 & -0.7872 & -0.7507 & -0.5822 & -0.3905 & -0.9883 & -0.9883 & -0.2732 & -0.2555 & -0.3063\\
    \bottomrule
   \end{tabular}
   }
    \label{table:ProximityEffectExperiment}
\vskip -0.1in
\end{table}

Table~\ref{table:ProximityEffectExperiment} summarizes the correlation between edge proximity and the number of appearances in the computation tree, averaged over all nodes in a total of 11 synthetic and real-world datasets (note that, for convenience, we have randomly sampled 100 nodes for the Arxiv dataset as such sampling does not fundamentally change the overall tendency). Positive correlations imply that edges with closer proximity tend to appear less in the computation tree, and negative correlations vice versa. The results clearly demonstrate that the proximity effect is prevalent in all datasets and is a common feature, thereby justifying our usage as a design principle ({\bf P1}) in \textsc{GAtt}.

\subsection{Runtime Comparison Among Different Explanation Methods (Table~\ref{table:SupplRuntimeExperiment})} \label{subsection:SuppRuntimeResults}

We further conduct experiments in order to measure the runtime among different explanation methods. Specifically, we compare our method \textsc{GAtt} with other baselines, including saliency (SA)~\cite{Simonyan2013saliency}, guided backpropagation (GB)~\cite{Springenberg2014guidedbackprop}, integrated gradient (IG)~\cite{Sundararajan2017integratedgradient}, GNNExplainer (GNNEx)~\cite{Ying2019GNNExplainer}, PGExplainer (PGEx)~\cite{Luo2020PGExplainer}, GraphMask (GM)~\cite{Schlichtkrull2021graphmask}, and FastDnX (FDnX)~\cite{Pereira2023FastDnX}. We calculate the edge attribution scores using the following three datasets:
\begin{itemize}
    \item BA-Shapes: 400 target nodes (\textit{i.e.}, all nodes included in the house-shaped subgraph), using a 3-layer GAT model.
    \item Infection: 974 target nodes (\textit{i.e.}, all nodes with a unique ground truth infection path), using a 3-layer GAT model.
    \item Arxiv: 10,000 target nodes, using a 2-layer GAT model.
\end{itemize}

For \textsc{GAtt}, we consider both the matrix-based computation introduced in Proposition~\ref{proposition:gattcomputation}, as well as the batch computation in Appendix~\ref{appendix:batchcalculation}, which are denoted as \textsc{GAtt} (mat) and \textsc{GAtt} (batch), respectively.

\begin{table*}[h]
\small
\centering
    \caption{Runtime comparison of \textsc{GAtt} against other baselines. The table summarizes the total wall-clock time required in order to calculate the local edge attributions for 400, 973, and 10,000 nodes on the BA-Shapes, Infection, and Arxiv dataset, respectively.}
    \resizebox{\linewidth}{!}{
    \begin{tabular}{lccccccccc}
    \toprule
    Dataset & \textsc{GAtt} (mat)& \textsc{GAtt} (batch) &  SA & GB & IG & GNNEx & PGEx & GM & FDnX\\ 
    \midrule
    \multirow{2}{*}{BA-Shapes} & 23.8s & 0.48s & 1.58s & 1.60s & 69.8s & 250s & 318s & 36.8s & 2.24s \\
    & ($\times$1) & \cellcolor{Turquoise!60} ($\times$0.020) & \cellcolor{Turquoise!25} ($\times$0.066) & \cellcolor{Turquoise!25} ($\times$0.067) & \cellcolor{red!32} ($\times$2.937) & \cellcolor{red!57} ($\times$10.535) & \cellcolor{red!62} ($\times$13.373) & \cellcolor{red!19} ($\times$1.549) & \cellcolor{Turquoise!20} ($\times$0.094) \\ \midrule
    \multirow{2}{*}{Infection} & 83.8s & 7.14s & 4.44s & 4.53s & 235s & 598s & 904s & 136s & 3.90s \\ 
    & ($\times$1) & \cellcolor{Turquoise!21} ($\times$0.085) & \cellcolor{Turquoise!28} ($\times$0.053) & \cellcolor{Turquoise!28} ($\times$0.054) & \cellcolor{red!30} ($\times$2.804) & \cellcolor{red!49} ($\times$7.136) & \cellcolor{red!57} ($\times$10.788) & \cellcolor{red!19} ($\times$1.623) & \cellcolor{Turquoise!31} ($\times$0.047) \\ \midrule
    \multirow{2}{*}{Arxiv} & 924s & OOM & 232s & 232s & 3h 11m & 8h 18m & 14h 54m & OOM & OOM \\
    & ($\times$1) & - & \cellcolor{Turquoise!18} ($\times$0.25) & \cellcolor{Turquoise!18} ($\times$0.25) & \cellcolor{red!60} ($\times$12.40) & \cellcolor{red!80} ($\times$32.33) & \cellcolor{red!91} ($\times$58.05) & - & - \\
    \bottomrule
   \end{tabular}
   }
   \label{table:SupplRuntimeExperiment}
\end{table*}

Table~\ref{table:SupplRuntimeExperiment} summarizes the runtime comparison of \textsc{GAtt} and all the baselines by measuring the wall-clock time required to calculate edge attributions for all three datasets. From the table, we observe the following:
\begin{itemize}
    \item \textbf{\textsc{GAtt} (mat) is reasonably fast compared to other attribution methods}: For all BA-Shapes, Infection, and Arxiv datasets, edge attribution calculation by \textsc{GAtt} shows moderately fast runtimes.
    \begin{itemize}
        \item \textsc{GAtt} (mat) shows much faster runtime compared to GNNEx, PGEx, and GM, which either solve an optimization problem or require a separate learning module. Such baseline approaches not only lead to slow explanations but also expose the quality of explanations to be susceptible to various hyperparameters and random seeds (see relevant discussions in Section~\ref{subsection:accuracyexperiment}). Since \textsc{GAtt} (mat) performs several additional computations based on the existing attention values, the benefits of \textsc{GAtt} (mat) with respect to the runtime over such methods are quite expected.
        \item IG also exhibits a much slower runtime compared to \textsc{GAtt} (mat). Although IG does not involve any optimization or learning process for explanations, it requires calculating an integral of the model gradients along the path from a baseline to the input, resulting in multiple gradient passes since the integral is calculated by a discretized approximation. In contrast, \textsc{GAtt} (mat) does not require any multiple gradient passes or approximate summations, resulting in a noticeably faster runtime.
        \item SA, GB, and FDnX shows faster explanation runtimes compared to \textsc{GAtt} (mat). This is acceptable due to the following reasons. SA and GB basically compute the gradient with respect to the input, along with an operation that is efficiently supported by many machine learning packages (\textit{e.g.}, PyTorch). For FDnX, knowledge distillation onto a surrogate SGC model~\cite{Wu2019SGC} can be performed in a highly effective manner (knowledge distillation takes less than $\sim$1 second for BA-Shapes on a modern GPU . Details on implementation and hardware specifications are found in Appendix~\ref{subsection:implementation}); then, FDnX performs basic matrix multiplications to acquire explanations, thus contributing to a faster runtime compared to \textsc{GAtt} (mat). However, such matrix multiplications result in memory limitations for larger datasets, which is discussed below.
    \end{itemize}
    \item \textbf{\textsc{GAtt} (batch) is \textit{extremely} fast}: For the BA-Shapes dataset, \textsc{GAtt} (batch) shows an extremely fast runtime of 0.48 seconds, which is nearly 50 times faster than \textsc{GAtt} (mat) and almost \textbf{663 times faster} than PGExplainer.
    \begin{itemize}
        \item \textsc{GAtt} (batch) is designed in the sense of inherently exploiting the computational benefits of \textsc{GAtt} (\textit{e.g.}, working with existing attention weights in a matrix form). The goal of \textsc{GAtt} (batch) is to acquire \textit{all} edge attributions associated with the target node \textit{at once} as a matrix form. 
        \item As a consequence, \textsc{GAtt} (batch) exhibits fast runtime, with the cost of full-size matrix multiplications, which may result in out-of-memory (OOM) issues ({\it e.g.}, the case on Arxiv). However, such memory limitation caused by full-size matrix multiplications is not unique for \textsc{GAtt} (batch), but also occurs for FastDnX. As mentioned previously, although the knowledge distillation onto the surrogate model can be performed efficiently (less than 3 seconds), the matrix multiplication during producing explanations causes the OOM problem. This can be mitigated by other techniques such as batch matrix multiplications or other advanced techniques~\cite{Mouhah2023LMM}, which is out of scope of our study. 
        \item Lastly, GM also exhibits OOM issues during learning on Arxiv, being moved into the GPU memory at the beginning of training. 
    \end{itemize}
\end{itemize}

In summary, \textsc{GAtt} makes full use of the benefit of its optimization-and-learning-free approach for edge attribution calculation, making it a sufficiently fast method compared to other attribution calculation methods. Furthermore, \textsc{GAtt} (batch) can be an extremely fast alternative to \textsc{GAtt} (mat), as long as the memory is sufficient for full-size matrix multiplications.

\begin{figure}[h]
  \centering
  \begin{subfigure}[b]{0.3\linewidth}
    \includegraphics[width=\linewidth]{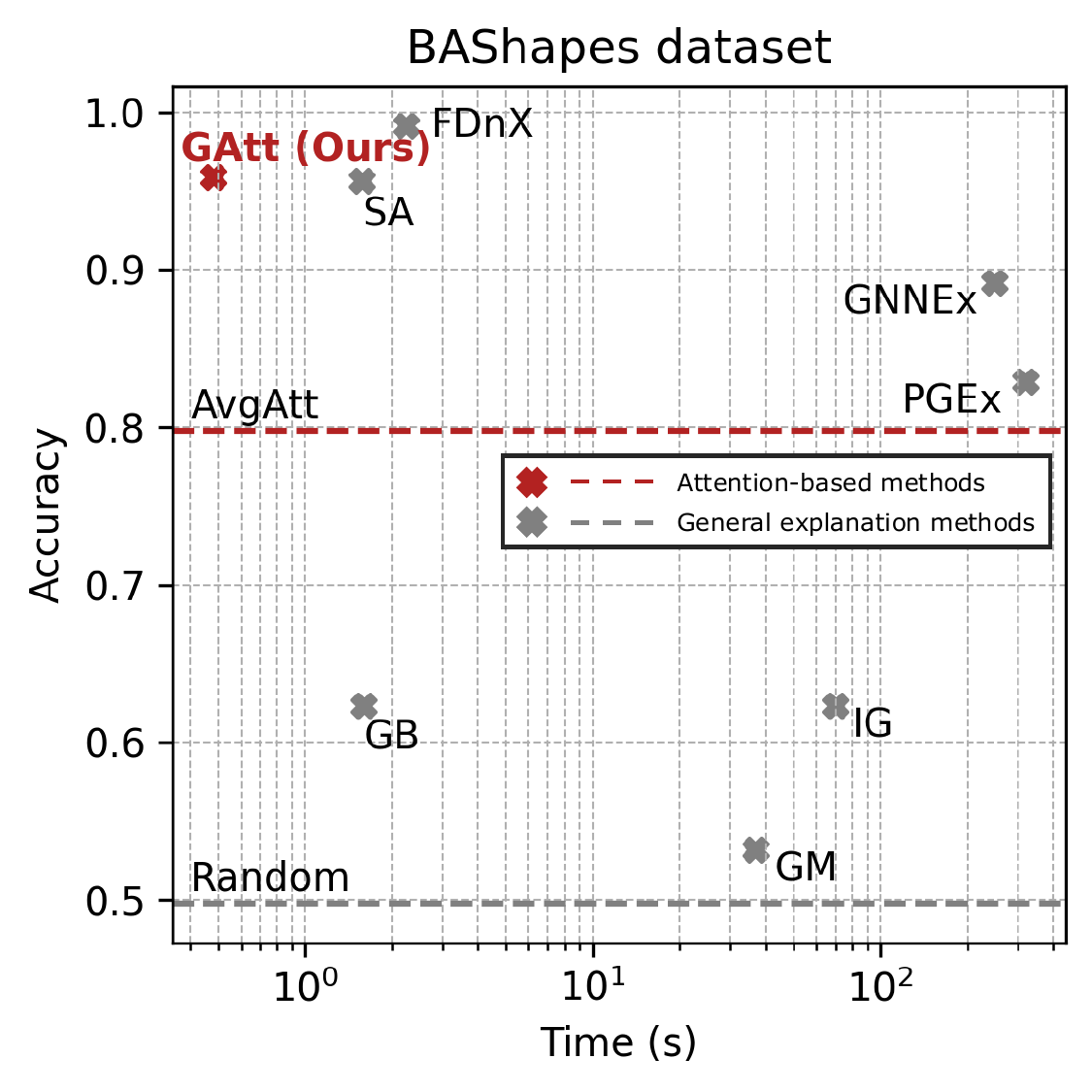}
    \caption{BA-Shapes dataset.}
    \label{subfig:2DplotBA}
  \end{subfigure} 
  \begin{subfigure}[b]{0.3\linewidth}
    \includegraphics[width=\linewidth]{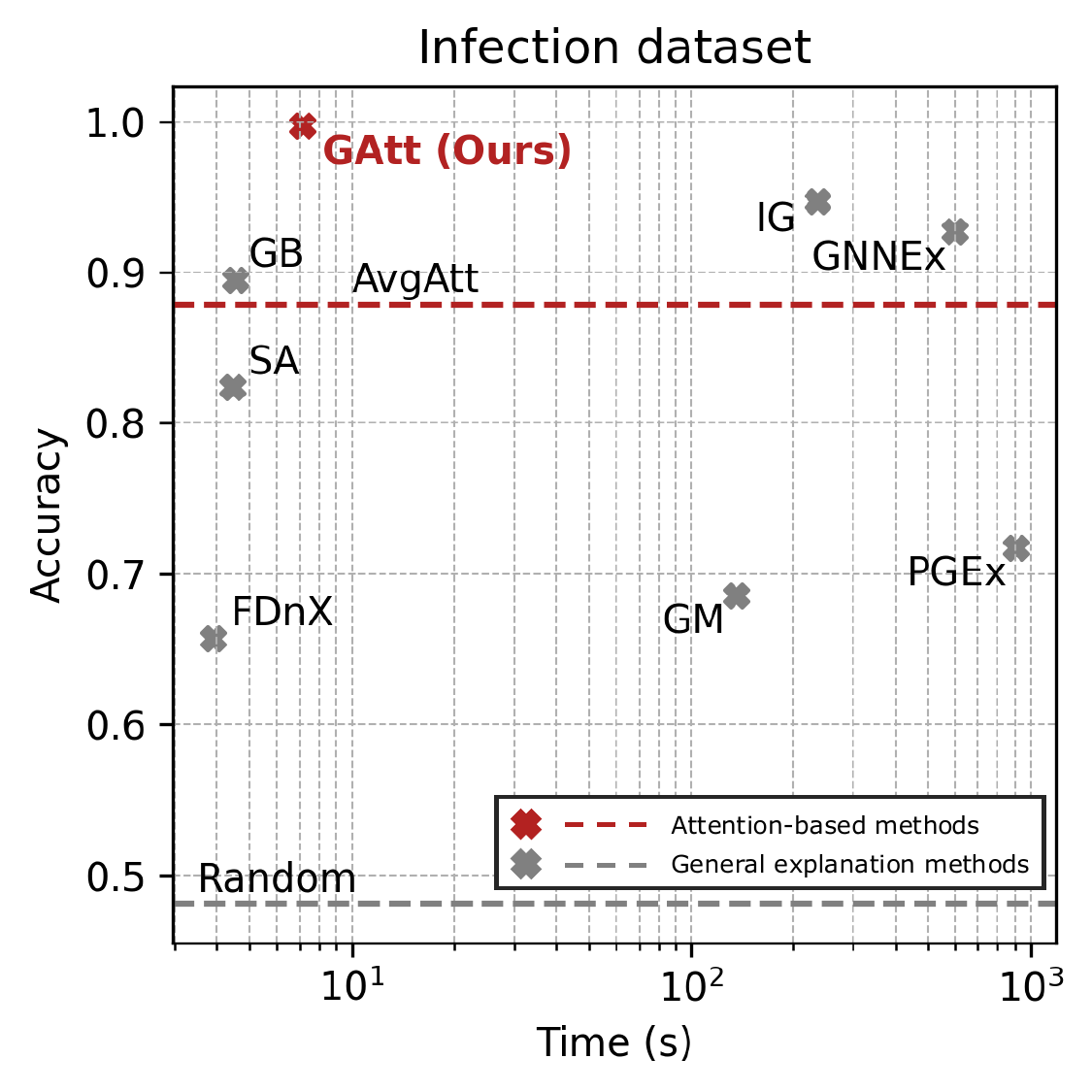}
    \caption{Infection dataset.}
    \label{subfig:2DplotInfection}
  \end{subfigure}
  \caption{An accuracy--runtime plot of various edge attribution methods.}
  \label{fig:2Daccuracyruntimeplot}
\end{figure}

Finally, we provide a comprehensive comparison of different edge attribution methods by illustrating an accuracy--runtime plot. Figure~\ref{fig:2Daccuracyruntimeplot} shows that \textsc{GAtt} offers a promising balance between explanation accuracy and runtime, compared to other edge attribution methods.

\clearpage

\begin{figure}[h!]
  \centering
  \begin{subfigure}[b]{0.44\linewidth}
    \includegraphics[width=\linewidth]{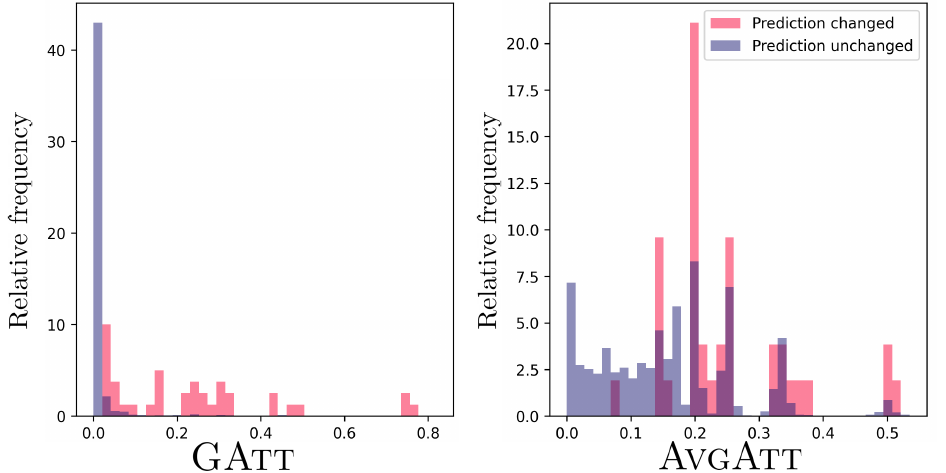}
    \caption{Cora (2-layer GAT).}
    \label{subfig:AUROCGATCora2L}
  \end{subfigure} 
  \begin{subfigure}[b]{0.44\linewidth}
    \includegraphics[width=\linewidth]{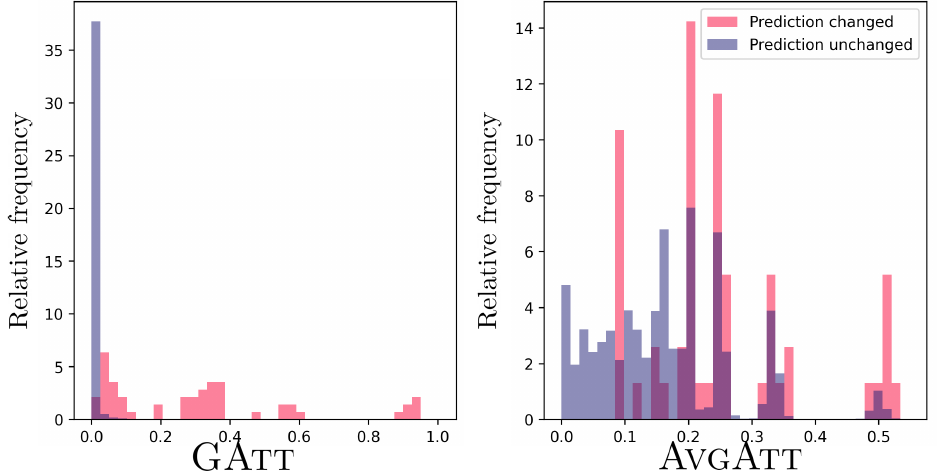}
    \caption{Cora (3-layer GAT).}
    \label{subfig:AUROCGATCora3L}
  \end{subfigure} 
  \begin{subfigure}[b]{0.44\linewidth}
    \includegraphics[width=\linewidth]{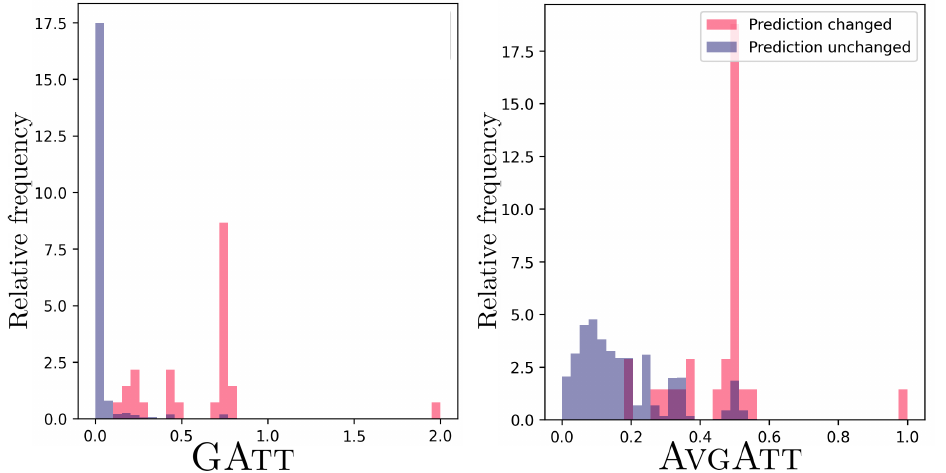}
    \caption{Citeseer (2-layer GAT).}
    \label{subfig:AUROCGATCiteseer2L}
  \end{subfigure} 
  \begin{subfigure}[b]{0.44\linewidth}
    \includegraphics[width=\linewidth]{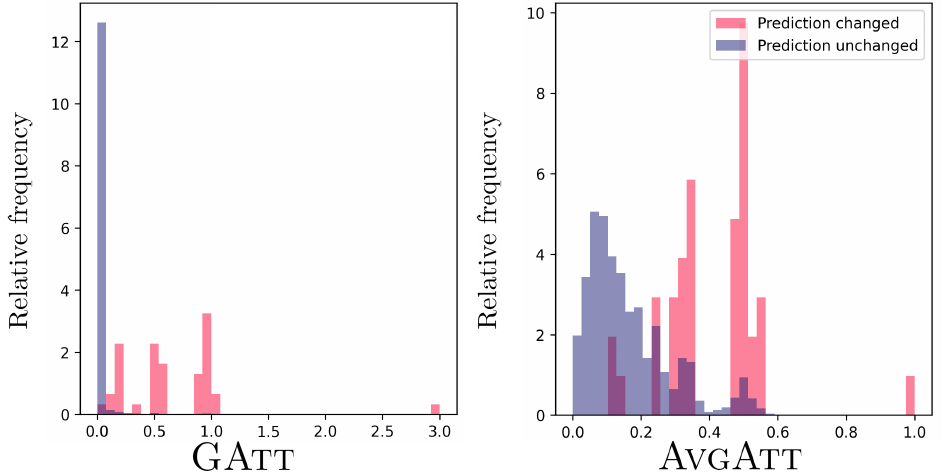}
    \caption{Citeseer (3-layer GAT).}
    \label{subfig:AUROCGATCiteseer3L}
  \end{subfigure} 
  \begin{subfigure}[b]{0.44\linewidth}
    \includegraphics[width=\linewidth]{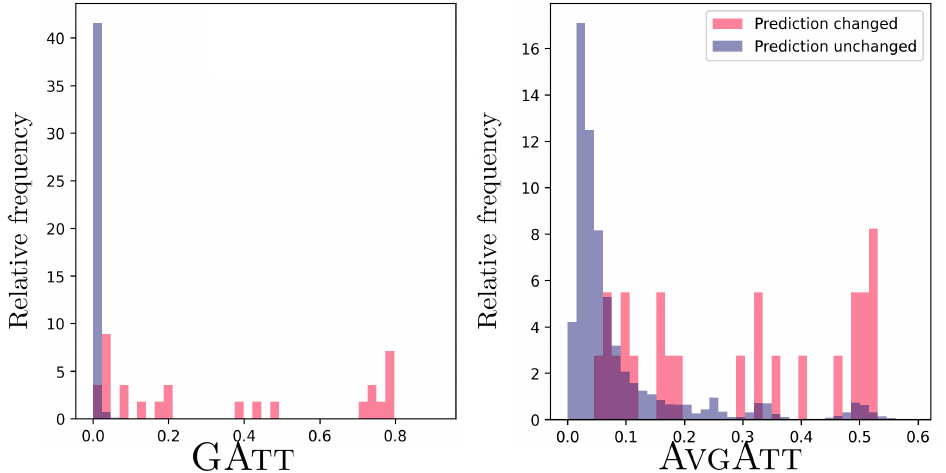}
    \caption{Pubmed (2-layer GAT).}
    \label{subfig:AUROCGATPubmed2L}
  \end{subfigure} 
  \begin{subfigure}[b]{0.44\linewidth}
    \includegraphics[width=\linewidth]{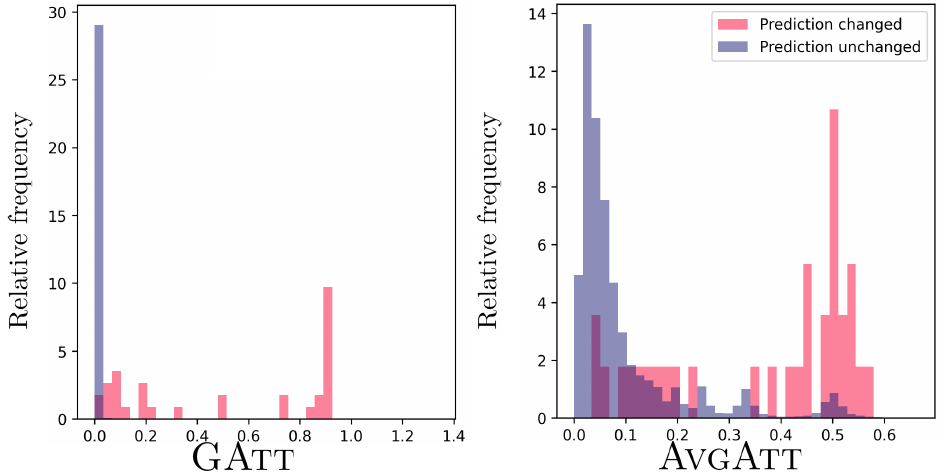}
    \caption{Pubmed (3-layer GAT).}
    \label{subfig:AUROCGATPubmed3L}
  \end{subfigure} 
  \begin{subfigure}[b]{0.44\linewidth}
    \includegraphics[width=\linewidth]{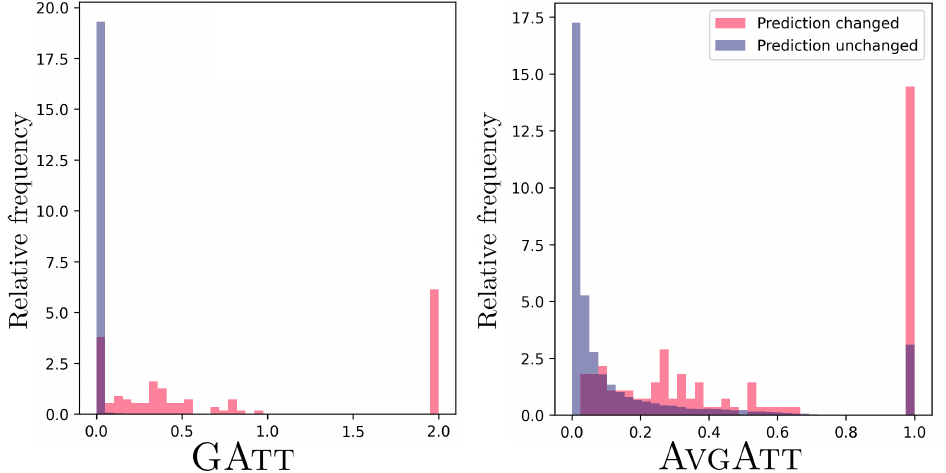}
    \caption{Arxiv (2-layer GAT).}
    \label{subfig:AUROCGATArxiv2L}
  \end{subfigure} 
  \begin{subfigure}[b]{0.44\linewidth}
    \includegraphics[width=\linewidth]{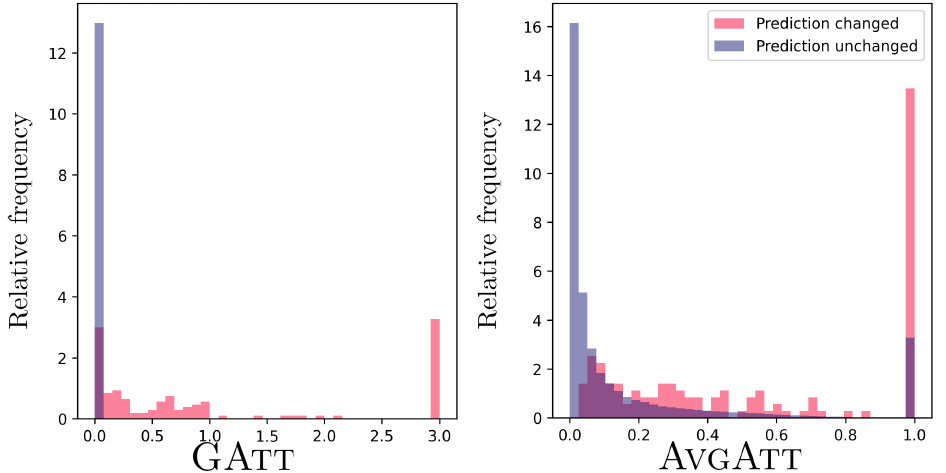}
    \caption{Arxiv (3-layer GAT).}
    \label{subfig:AUROCGATArxiv3L}
  \end{subfigure} 
  \caption{Visualization of experimental results for Cora, Citeseer, Pubmed, and Arxiv datasets, showing normalized histograms between two cases indicating whether the prediction has changed by attention reduction for the GAT model.}
  \label{figure:SupplFaithfulnessExpVisGAT}
\end{figure}

\begin{figure}[h]
  \centering
  \begin{subfigure}[b]{0.44\linewidth}
    \includegraphics[width=\linewidth]{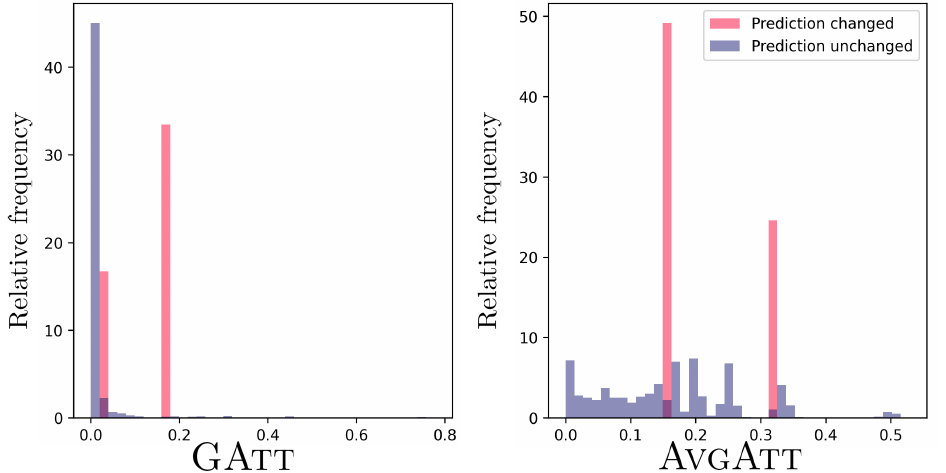}
    \caption{Cora (2-layer GATv2).}
    \label{subfig:AUROCGATv2Cora2L}
  \end{subfigure} 
  \begin{subfigure}[b]{0.44\linewidth}
    \includegraphics[width=\linewidth]{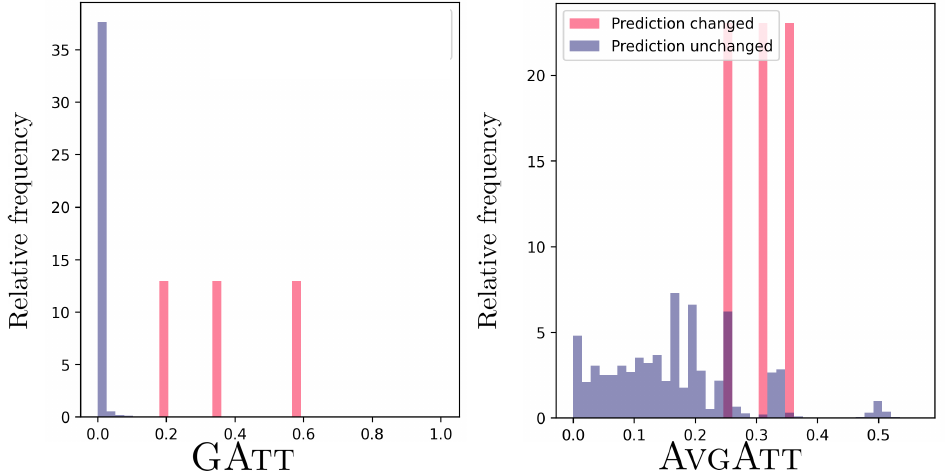}
    \caption{Cora (3-layer GATv2).}
    \label{subfig:AUROCGATv2Cora3L}
  \end{subfigure} 
  \begin{subfigure}[b]{0.44\linewidth}
    \includegraphics[width=\linewidth]{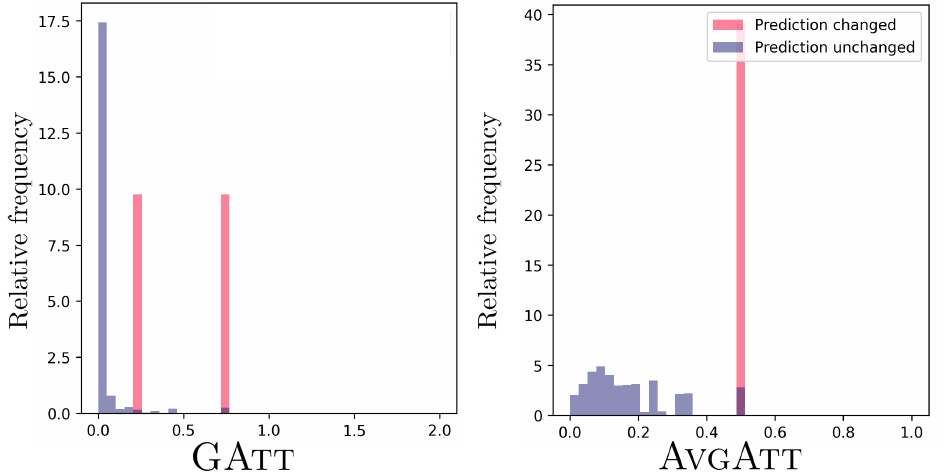}
    \caption{Citeseer (2-layer GATv2).}
    \label{subfig:AUROCGATv2Citeseer2L}
  \end{subfigure} 
  \begin{subfigure}[b]{0.44\linewidth}
    \includegraphics[width=\linewidth]{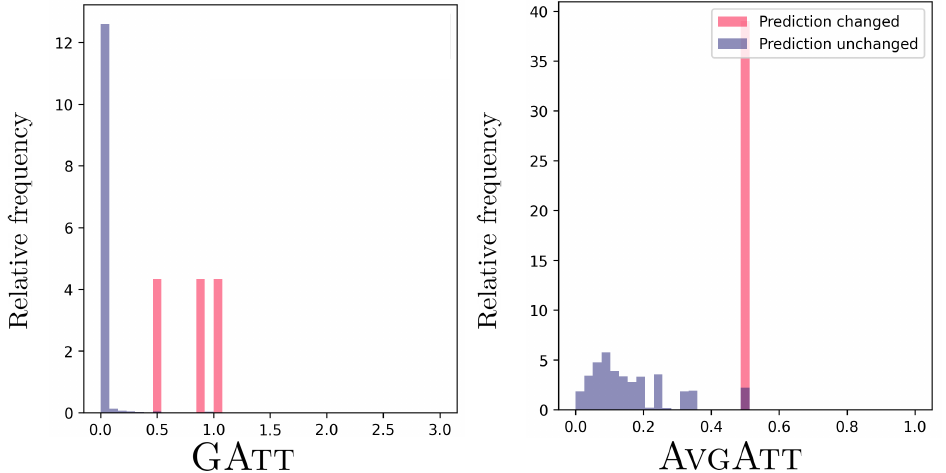}
    \caption{Citeseer (3-layer GATv2).}
    \label{subfig:AUROCGATv2Citeseer3L}
  \end{subfigure} 
  \begin{subfigure}[b]{0.44\linewidth}
    \includegraphics[width=\linewidth]{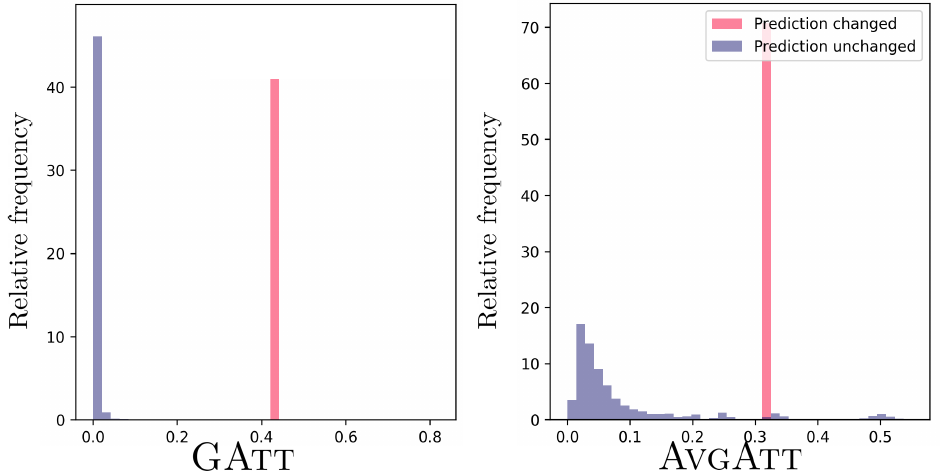}
    \caption{Pubmed (2-layer GATv2).}
    \label{subfig:AUROCGATv2Pubmed2L}
  \end{subfigure} 
  \begin{subfigure}[b]{0.44\linewidth}
    \includegraphics[width=\linewidth]{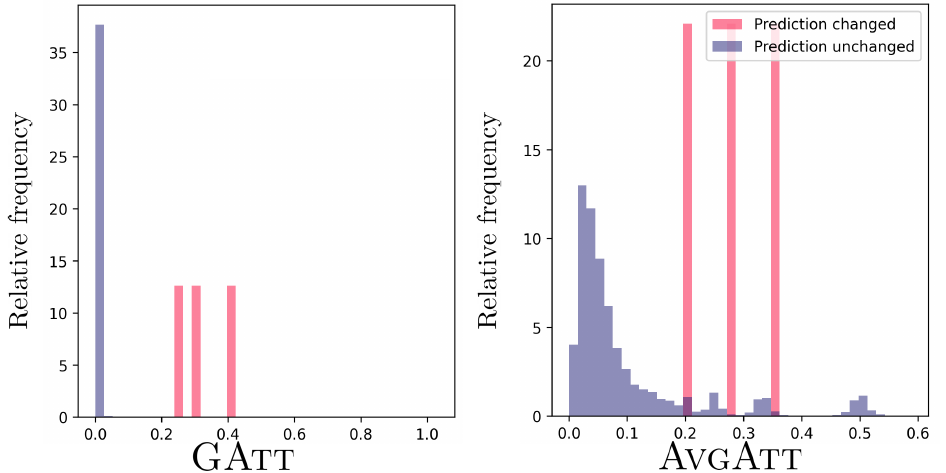}
      \caption{Pubmed (3-layer GATv2).}
    \label{subfig:AUROCGATv2Pubmed3L}
  \end{subfigure} 
  \begin{subfigure}[b]{0.44\linewidth}
    \includegraphics[width=\linewidth]{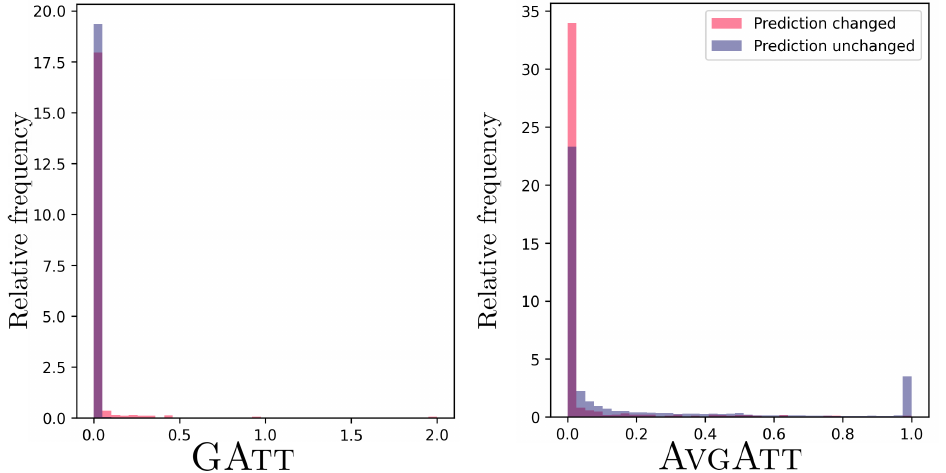}
    \caption{Arxiv (2-layer GATv2).}
    \label{subfig:AUROCGATv2Arxiv2L}
  \end{subfigure} 
  \begin{subfigure}[b]{0.44\linewidth}
    \includegraphics[width=\linewidth]{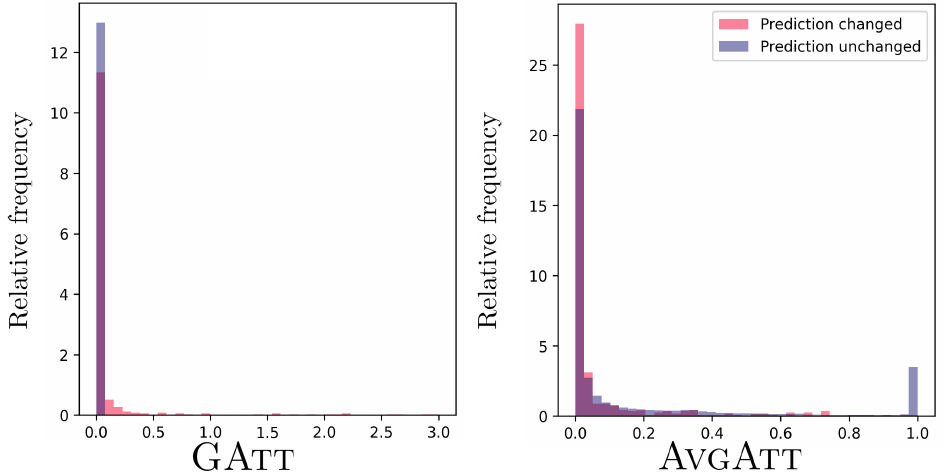}
    \caption{Arxiv (3-layer GATv2).}
    \label{subfig:AUROCGAv2ArxivT2L}
  \end{subfigure} 
  \caption{Visualization of experimental results for Cora, Citeseer, Pubmed, and Arxiv datasets, showing normalized histograms between two cases indicating whether the prediction has changed by attention reduction for the GATv2 model.}
  \label{figure:SupplFaithfulnessExpVisGATv2}
\end{figure}

\subsection{Further Visualizations of Faithfulness for \textsc{GAtt}, \textsc{GAtt}$_\textsc{sim}$, and \textsc{AvgAtt} (Figures~\ref{figure:SupplFaithfulnessExpVisGAT}--\ref{figure:SupplFaithfulnessExpVisGATv2})}\label{subsection:SupplFullResults} 
In Table~\ref{table:FaithfulCitationmain}, we have compared the effectiveness of \textsc{GAtt} against \textsc{AvgAtt} and random attribution. For completeness, we include further experimental results in Figures~\ref{figure:SupplFaithfulnessExpVisGAT} and~\ref{figure:SupplFaithfulnessExpVisGATv2} by plotting histograms for $\Delta_{\text{P}}$. Histograms reveal that, compared to \textsc{GAtt}, \textsc{AvgAtt} poorly assesses whether the prediction might change after attention reduction for both GAT and GATv2 models, while showing that the red bars (the case where the prediction has changed) and the blue bars (the case where prediction has not changed) widely overlap.

\clearpage



\section{Proof of Theorem~\ref{proposition:gattcomputation}}
\label{appendix:theoremproof}

From the fact that the computation tree of node $w$ for an $L$-layer GAT essentially represents the collection of all paths of length $L$ in the graph that ends with node $w$. With this in mind, we recall the original definition of \textsc{GAtt} in Eq.~(\ref{eq:edgeattributiondefinition}):

\begin{equation}
    \phi^w_{i,j} = \sum_{m'=1}^l \sum_{\lambda_{i,j,v}^{m'} \in \Lambda_v^{m'}(e_{i,j})} C(\alpha[\lambda_{i,j,w}^{m'}]) \alpha[\lambda_{i,j,w}^{m'}](1). \notag
\end{equation}

Using the double sum along with different path lengths, \textit{i.e.}, different values of $m$, Eq.~(\ref{eq:edgeattributiondefinition}) can be rewritten as:

\begin{equation}
    \phi^w_{i,j} = \alpha[\lambda_{i,j,w}^1](1) + \sum_{m'=2}^{L} \sum_{\lambda_{i,j,w}^{m'} \in \Lambda_w^{m'}(e_{i,j})} C(\alpha[\lambda_{i,j,w}^{m'}]) \alpha[\lambda_{i,j,w}^{m'}](1). \label{eq:intermediateproof}
\end{equation}

From the fact that $C(\alpha[\lambda_{i,j}^{m'}]) = \prod_{2\leq k \leq m'}\alpha[\lambda_{i,j}^{m'}](k)$, we have $C(\alpha[\lambda_{i,j}^{m'}]) \alpha[\lambda_{i,j}^{m'}](1) = \left(\prod_{2\leq k \leq m'}\alpha[\lambda_{i,j}^{m'}](k)\right) \alpha[\lambda_{i,j}^{m'}](1)$. Then, it follows that:
\begin{itemize}
    \item $\alpha[\lambda_{i,j,w}^{m'}](1) = {\bf I}_{w,j}[\mathbf{A}(L - m' + 1)]_{j,i}$ by definition.
    \item For the argument in the second term in Eq.~(\ref{eq:intermediateproof}),
    \begin{align}
        \sum_{\lambda_{i,j,w}^{m'} \in \Lambda_w^{m'}(e_{i,j})} C(\alpha[\lambda_{i,j,w}^{m'}]) \alpha[\lambda_{i,j,w}^{m'}](1) &= \left(\sum_{\lambda_{i,j,w}^{m'} \in \Lambda_w^{m'}(e_{i,j})} \prod_{2\leq k \leq m'}\alpha[\lambda_{i,j,w}^{m'}](k)\right) \notag \\
        &\times [\mathbf{A}(L - m' + 1)]_{j,i} \label{eq:Supplintermequation} \\
        &= [\mathbf{A}(L)\mathbf{A}(L-1)\cdots\mathbf{A}(L-m'+2)]_{w,j}\notag \\
        &\times [\mathbf{A}(L - m' + 1)]_{j,i}. \label{eq:Supplfinalequation}
    \end{align}
\end{itemize}
Since the term $\left(\sum_{\lambda_{i,j,w}^{m'} \in \Lambda_w(e_{i,j})} \prod_{2\leq k \leq m'}\alpha[\lambda_{i,j,w}^{m'}](k)\right)$ in the right-hand side of Eq. (\ref{eq:Supplintermequation}) represents the sum of the products of all attention weights along the paths from node $j$
 to node $w$ of length $(m'-1)$, by replacing $m=L-m'+1$, we finally have
\begin{equation}
    \phi^v_{i,j} = \sum_{m=1}^{L} [\mathbf{C}_L(L-m)]_{w, j} [{\bf A}(m)]_{j,i}, \notag
\end{equation}
where
\begin{equation}
    \mathbf{C}_L(k) = 
\begin{cases}
    {\bf I},              & \text{if } k=0,\\
    {\bf A}(L){\bf A}(L-1)\cdots {\bf A}(L-k+1),& \text{otherwise,} \notag
\end{cases}
\end{equation}
which completes the proof of this theorem.
\hfill\qedsymbol{}

\section{On the Potential Application of GAtt}
Since Att-GNN has been widely applied as a backbone, \textsc{GAtt} can be seamlessly used as an effective tool of explaining the underlying model to unlock valuable insights that have been previously difficult to discover. Therefore, \textsc{GAtt} has the potential to be used to explain practical applications, which we briefly outline:
\begin{itemize}
    \item Travel time estimation in transportation: Att-GNN has seen empirical success in travel time estimation (e.g.,~\citep{Fang2020constgat} from Baidu), and we expect the application of GAtt to such a domain so as to reveal which road pattern the model looks towards its estimation.
    \item Drug-target interaction in bioinformatics: Another exciting application where Att-GNN is often used as a backbone model lies in drug-target interaction prediction~\citep{Nguyen2021graphdta}, where GAtt aims to confirm that the underlying model has learned the correct substructures for molecule representations, or even has discovered novel ones that dominantly affect the drug-target interaction.
\end{itemize}

\begin{figure}[h!]
  \centering
  \begin{subfigure}[b]{0.35\linewidth}
    \includegraphics[width=\linewidth]{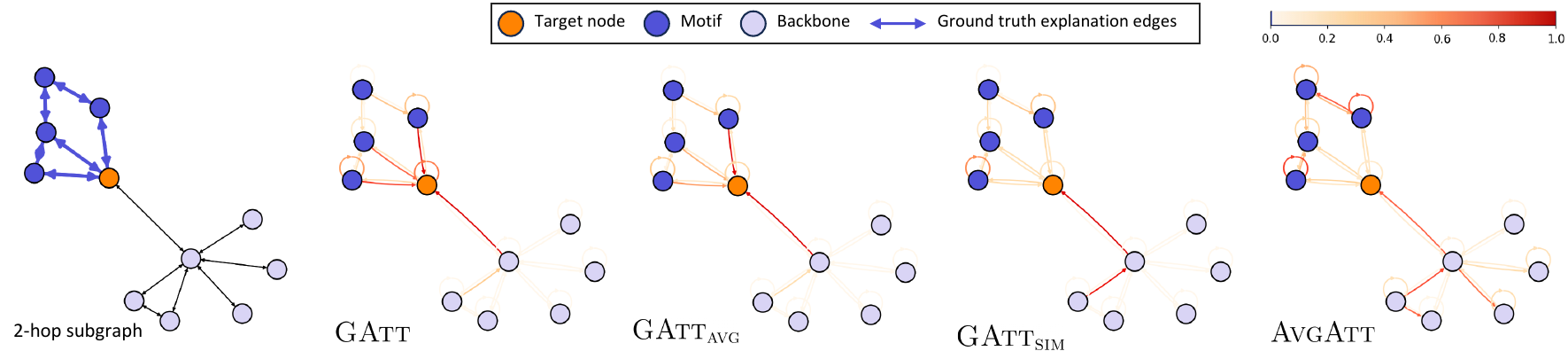}
    \caption{Node 600 (2-layer GAT).}
  \end{subfigure}
  \begin{subfigure}[b]{0.35\linewidth}
    \includegraphics[width=\linewidth]{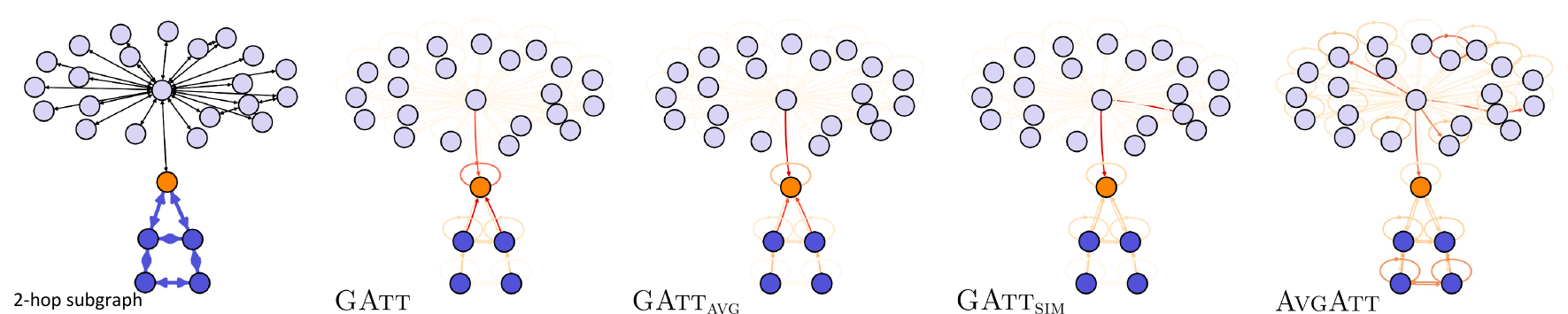}
    \caption{Node 629 (2-layer GAT).}
  \end{subfigure}
  \begin{subfigure}[b]{0.35\linewidth}
    \includegraphics[width=\linewidth]{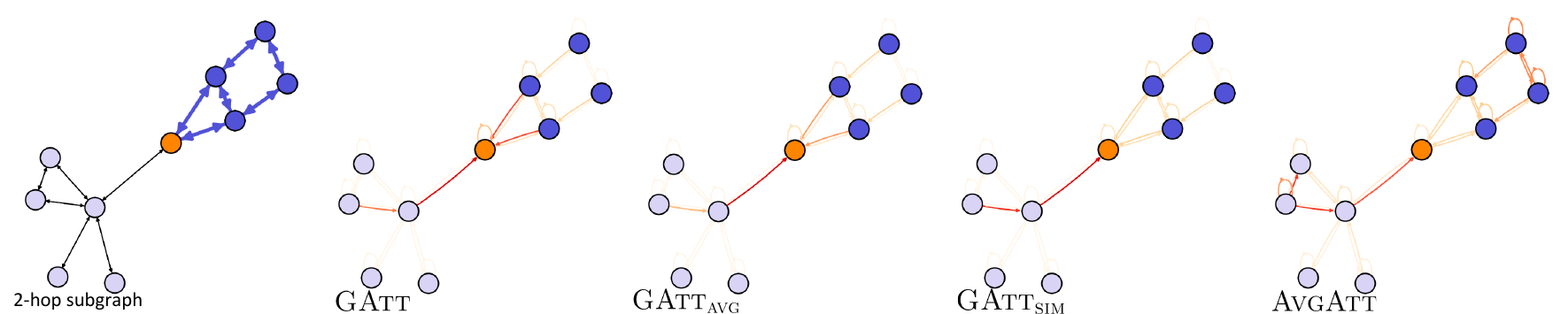}
    \caption{Node 634 (2-layer GAT).}
  \end{subfigure}
  \begin{subfigure}[b]{0.35\linewidth}
    \includegraphics[width=\linewidth]{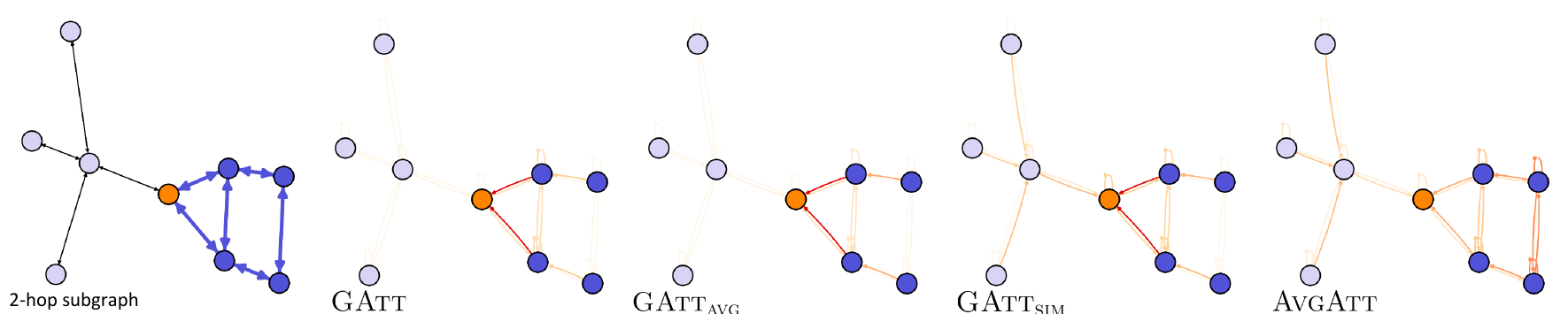}
    \caption{Node 679 (2-layer GAT).}
  \end{subfigure}
  \begin{subfigure}[b]{0.35\linewidth}
    \includegraphics[width=\linewidth]{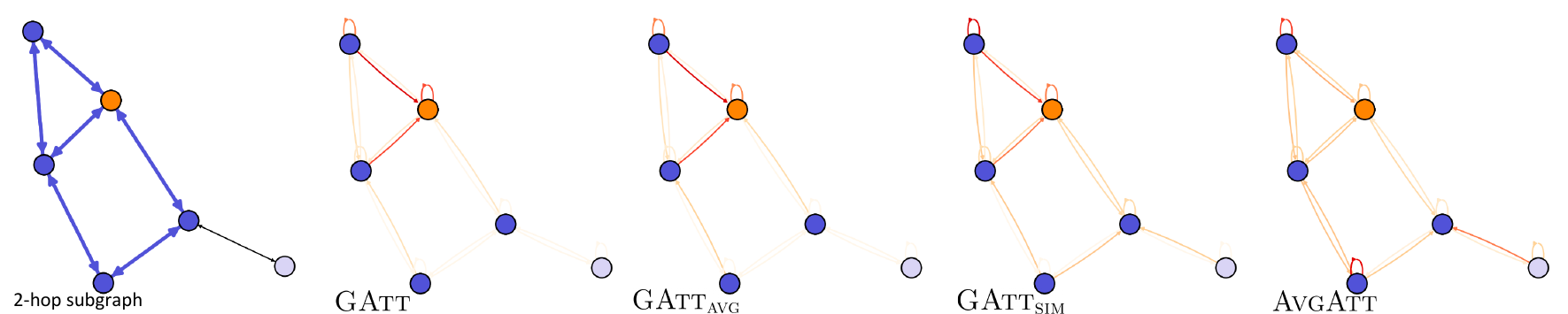}
    \caption{Node 690 (2-layer GAT).}
  \end{subfigure}
  \begin{subfigure}[b]{0.35\linewidth}
    \includegraphics[width=\linewidth]{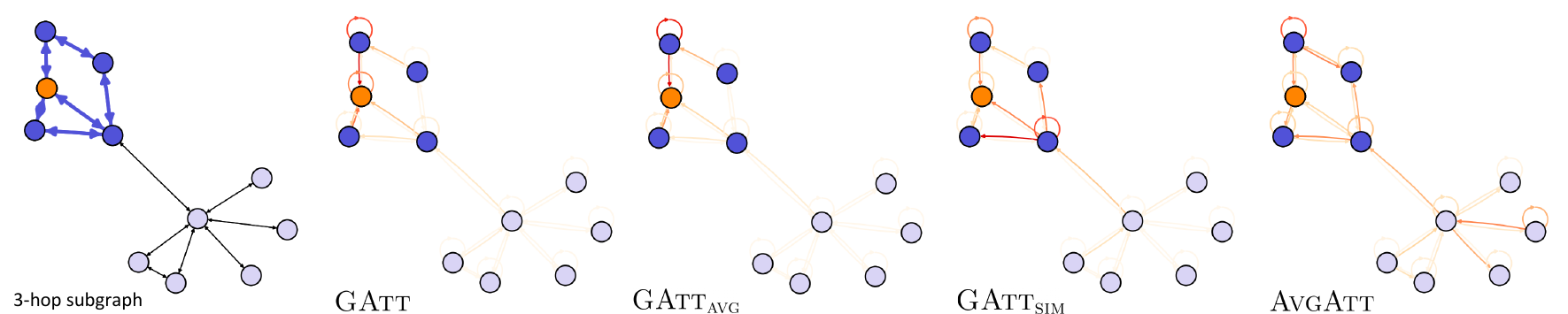}
    \caption{Node 601 (3-layer GAT).}
  \end{subfigure}
  \begin{subfigure}[b]{0.35\linewidth}
    \includegraphics[width=\linewidth]{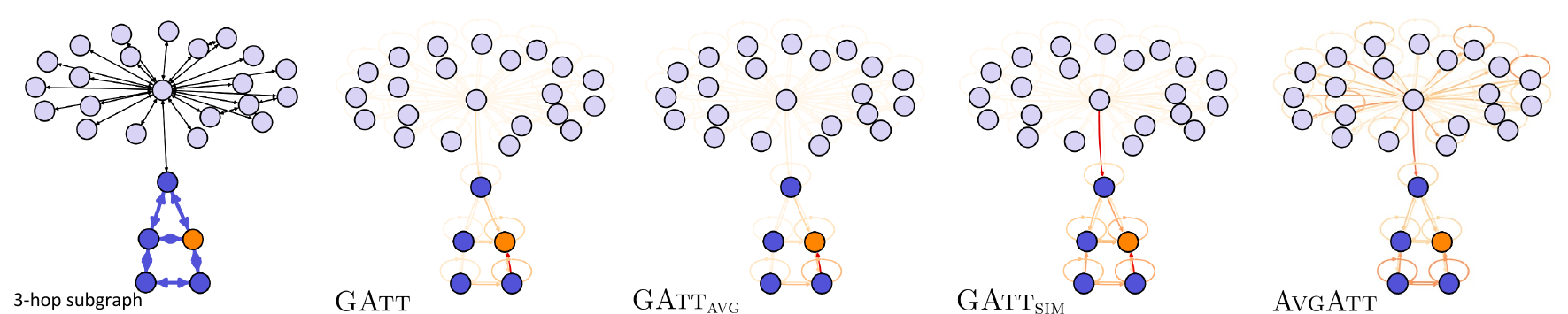}
    \caption{Node 626 (3-layer GAT).}
  \end{subfigure}
  \begin{subfigure}[b]{0.35\linewidth}
    \includegraphics[width=\linewidth]{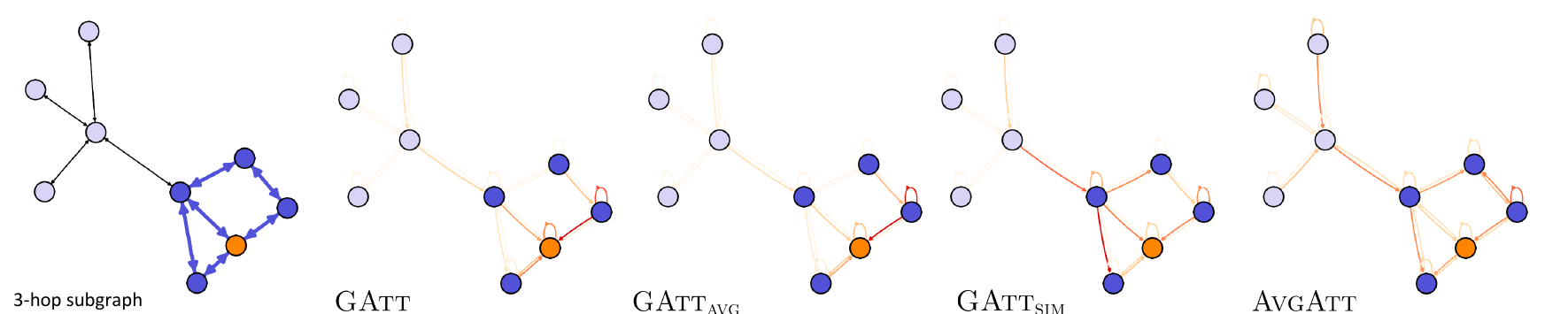}
    \caption{Node 635 (3-layer GAT).}
  \end{subfigure}
  \begin{subfigure}[b]{0.35\linewidth}
    \includegraphics[width=\linewidth]{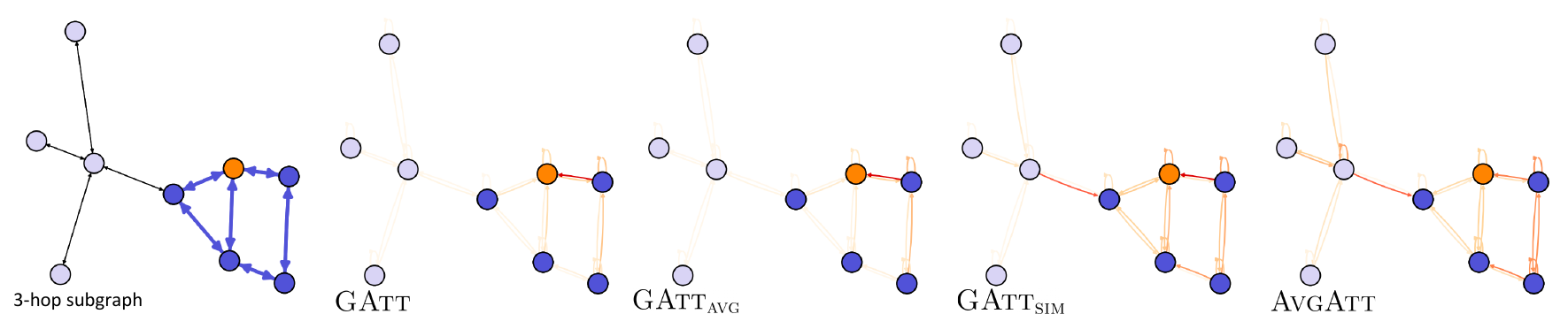}
    \caption{Node 676 (3-layer GAT).}
  \end{subfigure}
  \begin{subfigure}[b]{0.35\linewidth}
    \includegraphics[width=\linewidth]{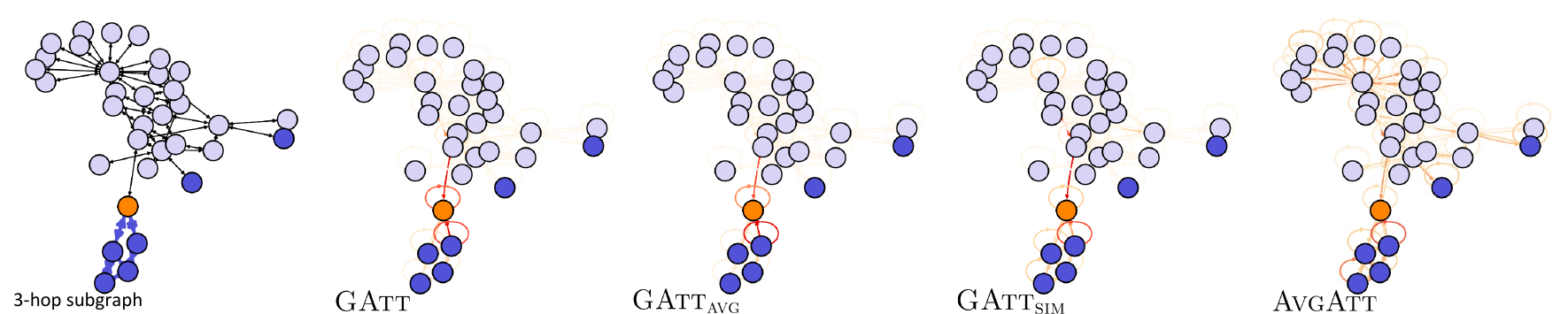}
    \caption{Node 687 (3-layer GAT).}
  \end{subfigure}
  \caption{Case studies on the BA-Shapes dataset for GAT.}
  \vskip -0.2in
  \label{fig:SupplCaseStudiesBAShapes}
\end{figure}

\begin{figure}[h!]
  \centering
  \begin{subfigure}[b]{0.35\linewidth}
    \includegraphics[width=\linewidth]{Case_study_Infection_2layer_node2_full.pdf}
    \caption{Node 2 (2-layer GAT).}
  \end{subfigure}
  \begin{subfigure}[b]{0.35\linewidth}
    \includegraphics[width=\linewidth]{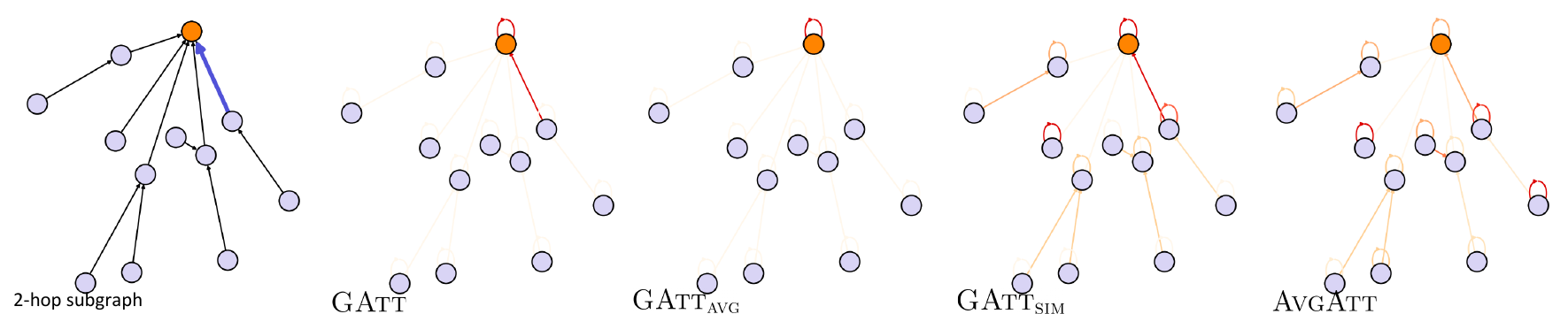}
    \caption{Node 44 (2-layer GAT).}
  \end{subfigure}
  \begin{subfigure}[b]{0.35\linewidth}
    \includegraphics[width=\linewidth]{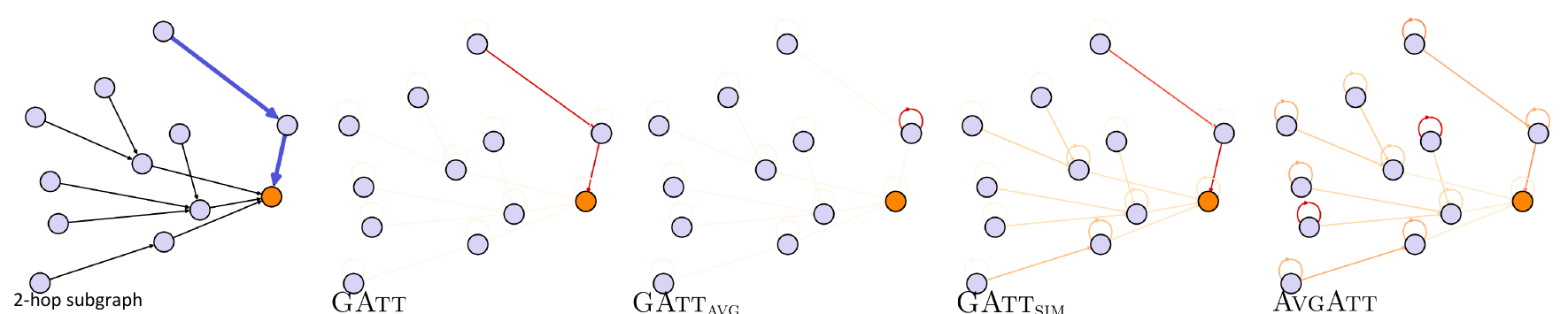}
    \caption{Node 54 (2-layer GAT).}
  \end{subfigure}
  \begin{subfigure}[b]{0.35\linewidth}
    \includegraphics[width=\linewidth]{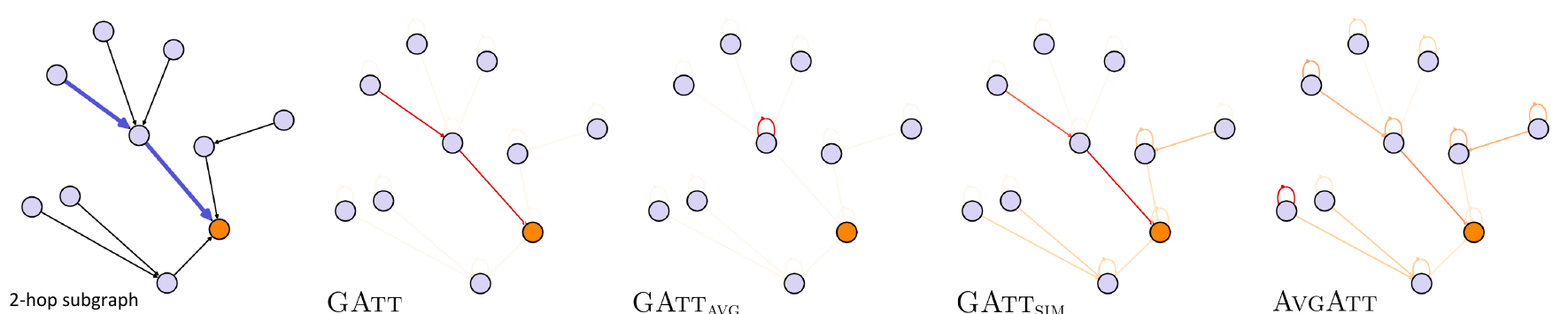}
    \caption{Node 355 (2-layer GAT).}
  \end{subfigure}
  \begin{subfigure}[b]{0.35\linewidth}
    \includegraphics[width=\linewidth]{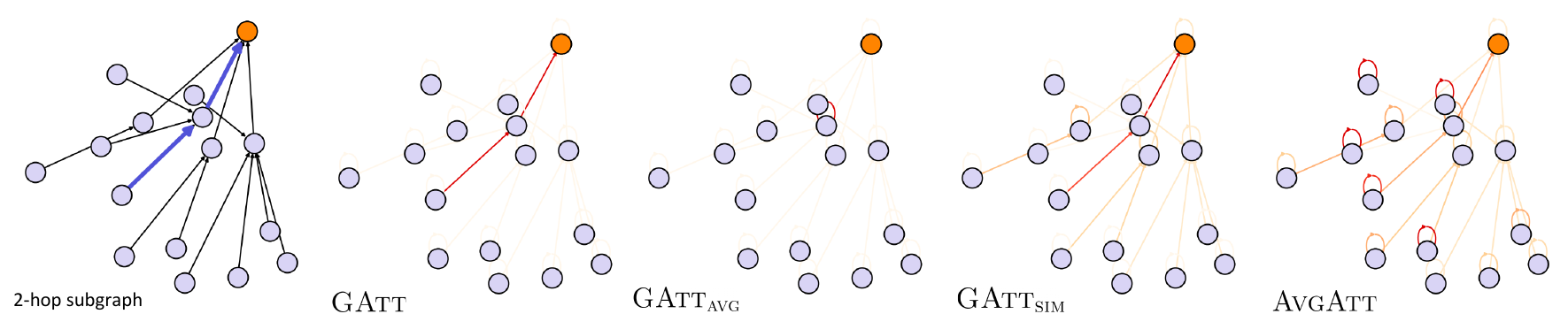}
    \caption{Node 326 (2-layer GAT).}
  \end{subfigure}
  \begin{subfigure}[b]{0.35\linewidth}
    \includegraphics[width=\linewidth]{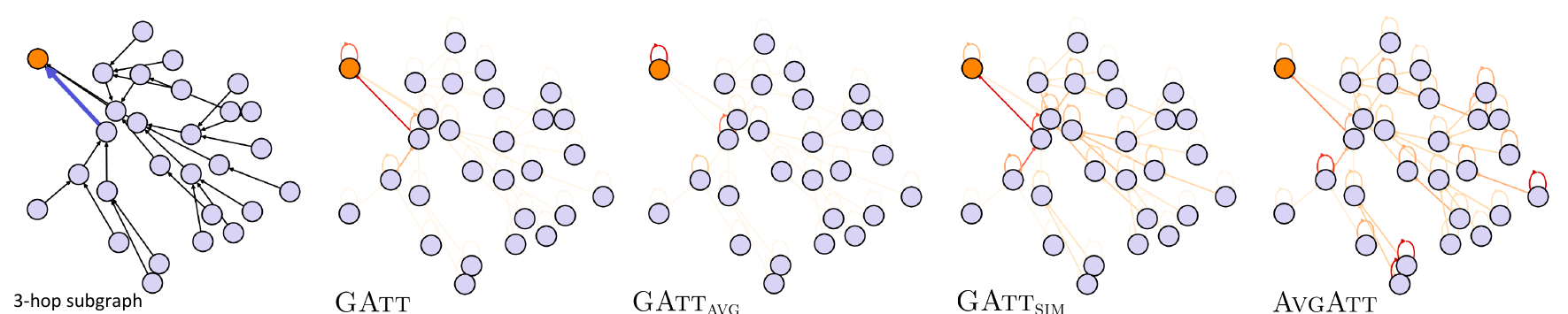}
    \caption{Node 2 (3-layer GAT).}
  \end{subfigure}
  \begin{subfigure}[b]{0.35\linewidth}
    \includegraphics[width=\linewidth]{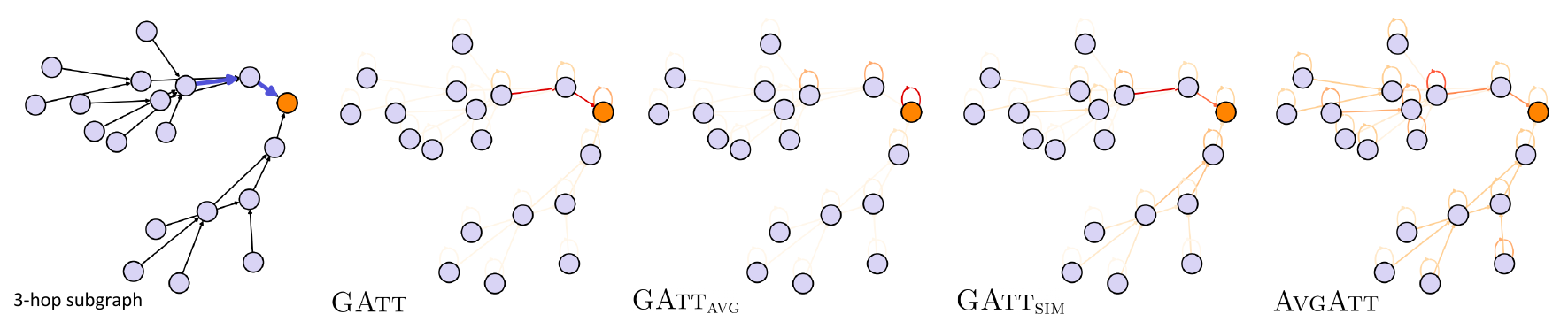}
    \caption{Node 44 (3-layer GAT).}
  \end{subfigure}
  \begin{subfigure}[b]{0.35\linewidth}
    \includegraphics[width=\linewidth]{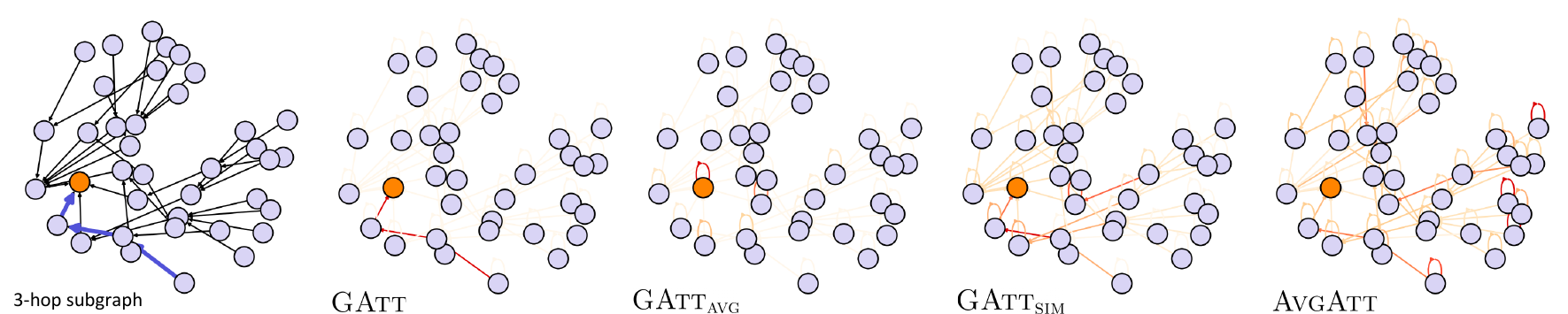}
    \caption{Node 54 (3-layer GAT).}
  \end{subfigure}
  \begin{subfigure}[b]{0.35\linewidth}
    \includegraphics[width=\linewidth]{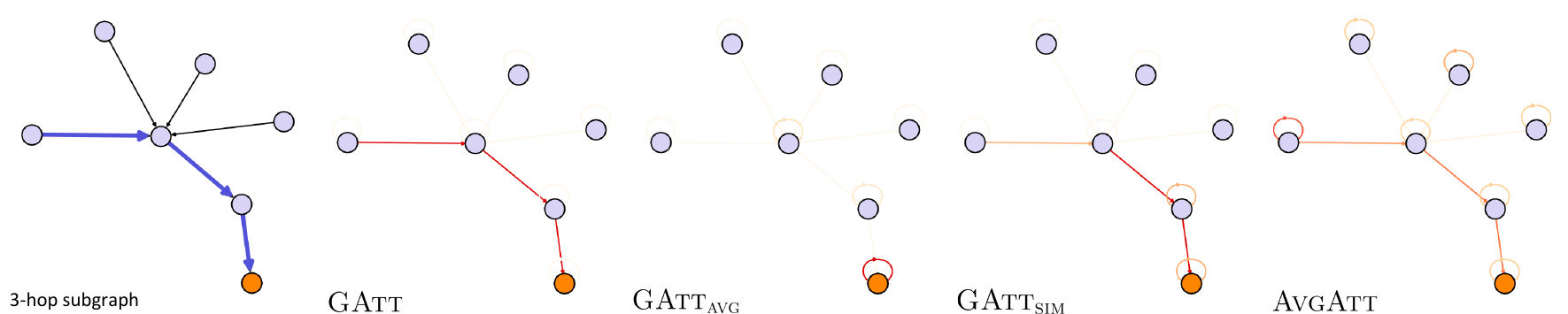}
    \caption{Node 355 (3-layer GAT).}
  \end{subfigure}
  \begin{subfigure}[b]{0.35\linewidth}
    \includegraphics[width=\linewidth]{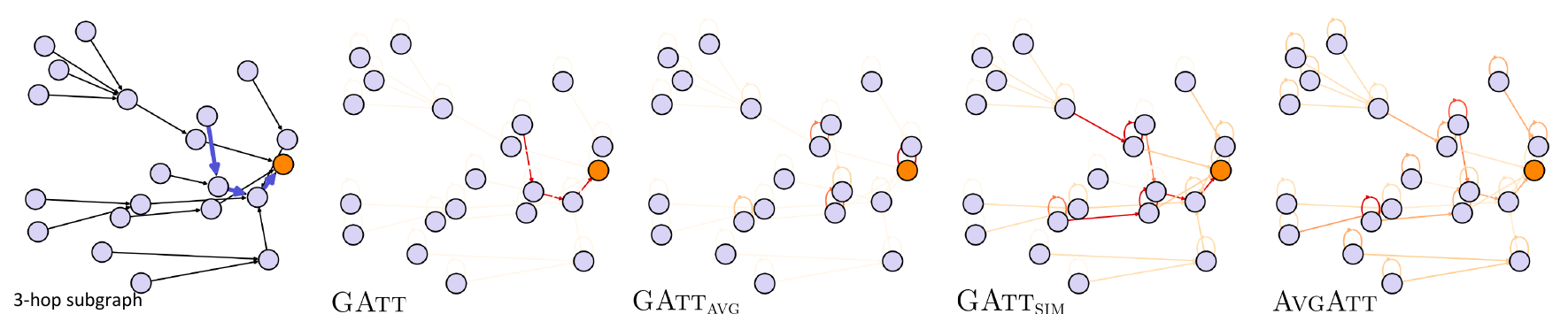}
    \caption{Node 326 (3-layer GAT).}
  \end{subfigure}
  \caption{Case studies on the Infection dataset for GAT.}
  \vskip -0.4in
  \label{fig:SupplCaseStudiesInfection}
\end{figure}

\begin{figure}[h!]
  \centering
  \begin{subfigure}[b]{0.35\linewidth}
    \includegraphics[width=\linewidth]{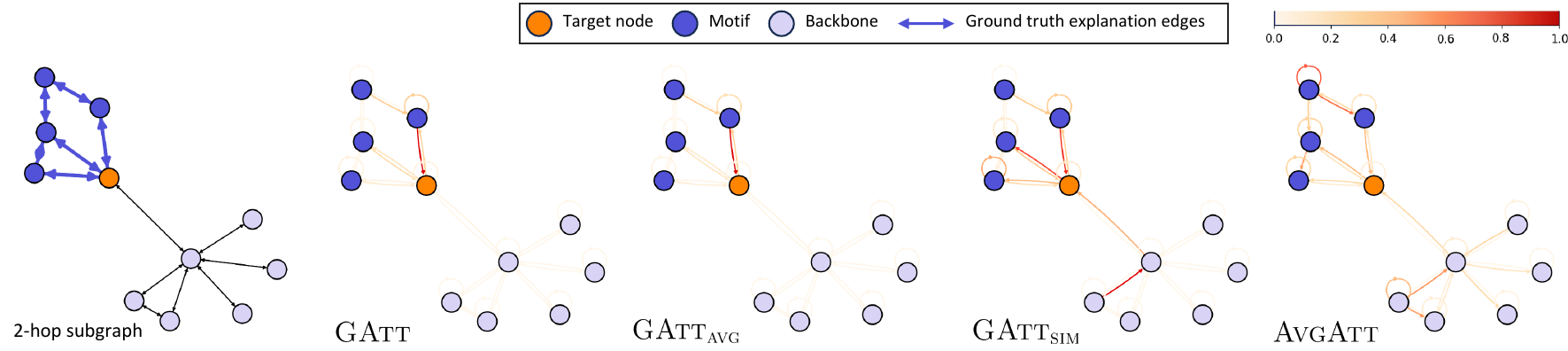}
    \caption{Node 600 (2-layer GATv2).}
  \end{subfigure}
  \begin{subfigure}[b]{0.35\linewidth}
    \includegraphics[width=\linewidth]{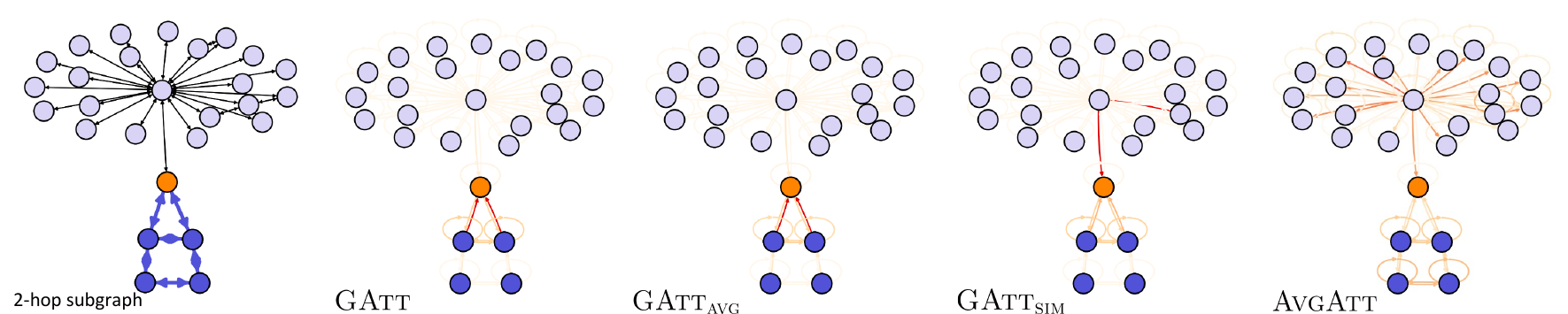}
    \caption{Node 629 (2-layer GATv2).}
  \end{subfigure}
  \begin{subfigure}[b]{0.35\linewidth}
    \includegraphics[width=\linewidth]{Case_study_BAShapes_2layer_node629_full_GATv2.pdf}
    \caption{Node 634 (2-layer GATv2).}
  \end{subfigure}
  \begin{subfigure}[b]{0.35\linewidth}
    \includegraphics[width=\linewidth]{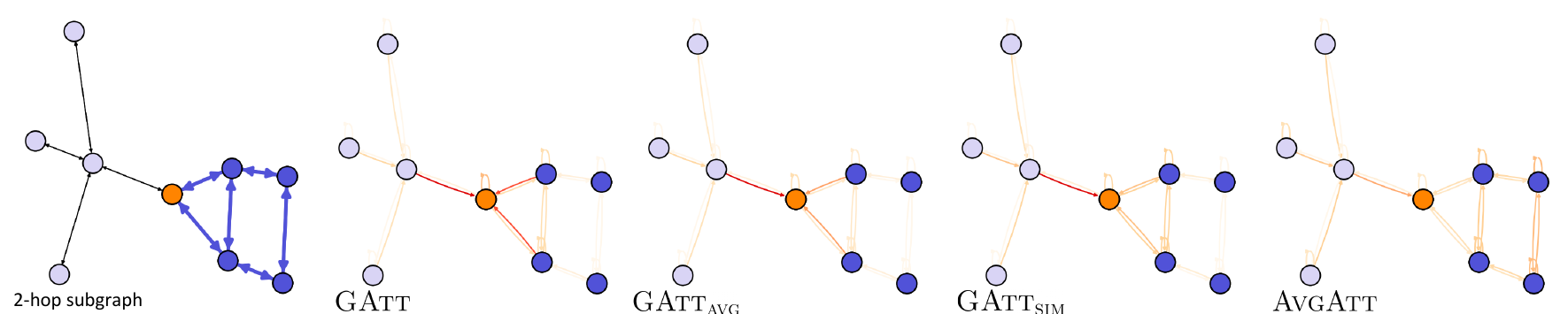}
    \caption{Node 679 (2-layer GATv2).}
  \end{subfigure}
  \begin{subfigure}[b]{0.35\linewidth}
    \includegraphics[width=\linewidth]{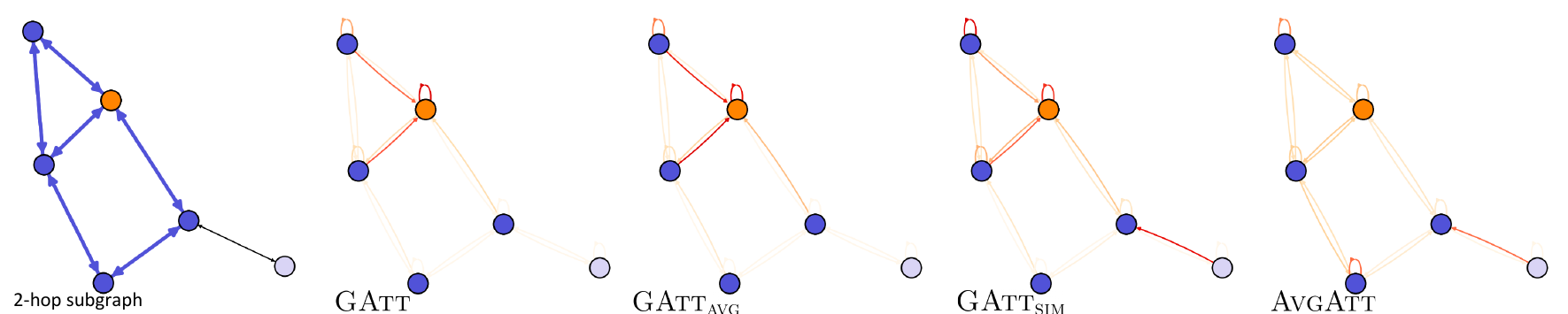}
    \caption{Node 690 (2-layer GATv2).}
  \end{subfigure}
  \begin{subfigure}[b]{0.35\linewidth}
    \includegraphics[width=\linewidth]{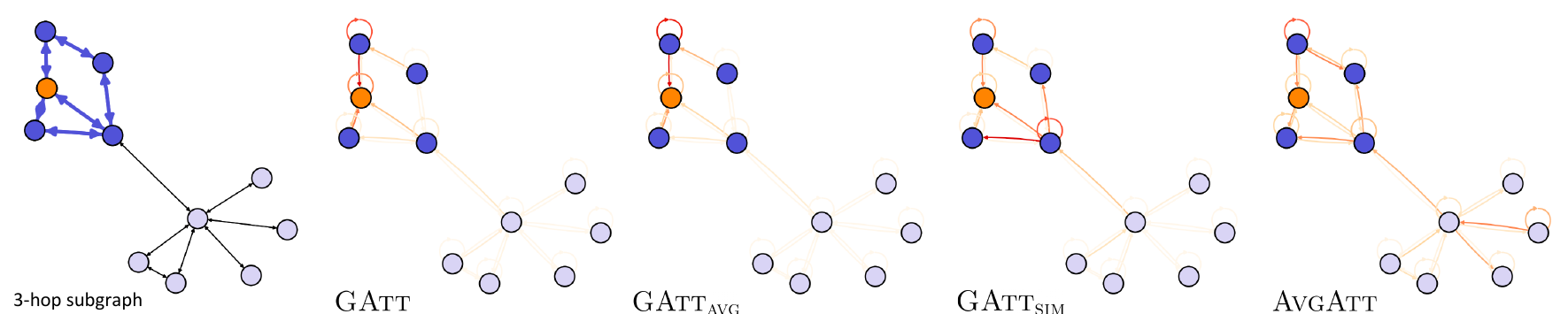}
    \caption{Node 601 (3-layer GATv2).}
  \end{subfigure}
  \begin{subfigure}[b]{0.35\linewidth}
    \includegraphics[width=\linewidth]{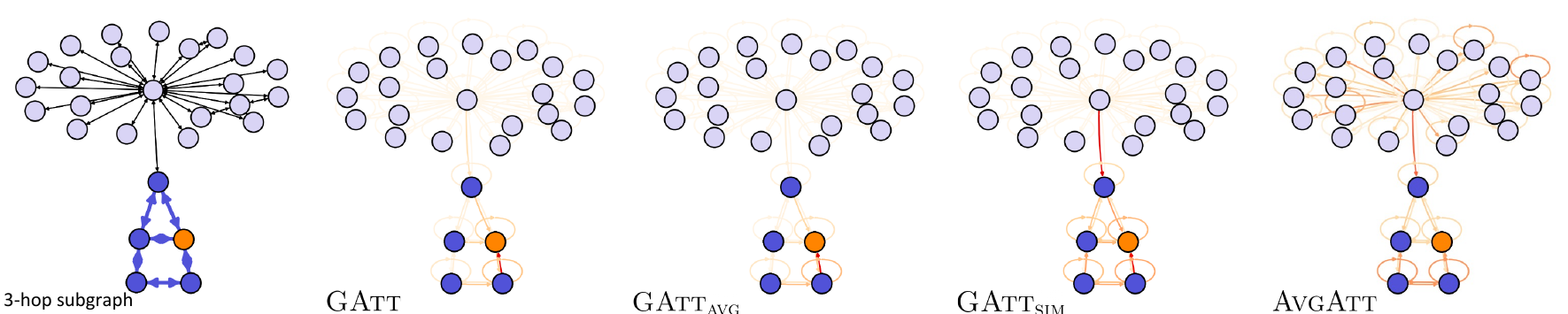}
    \caption{Node 626 (3-layer GATv2).}
  \end{subfigure}
  \begin{subfigure}[b]{0.35\linewidth}
    \includegraphics[width=\linewidth]{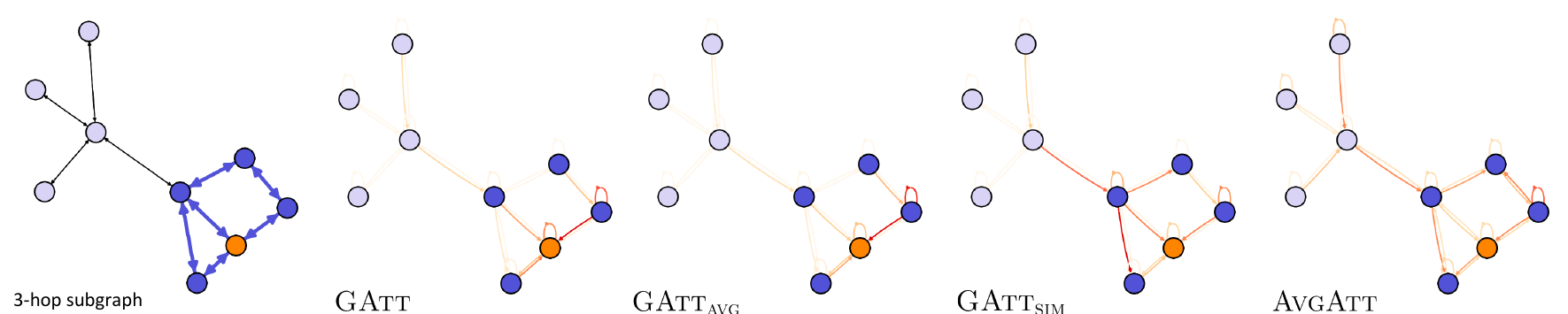}
    \caption{Node 635 (3-layer GATv2).}
  \end{subfigure}
  \begin{subfigure}[b]{0.35\linewidth}
    \includegraphics[width=\linewidth]{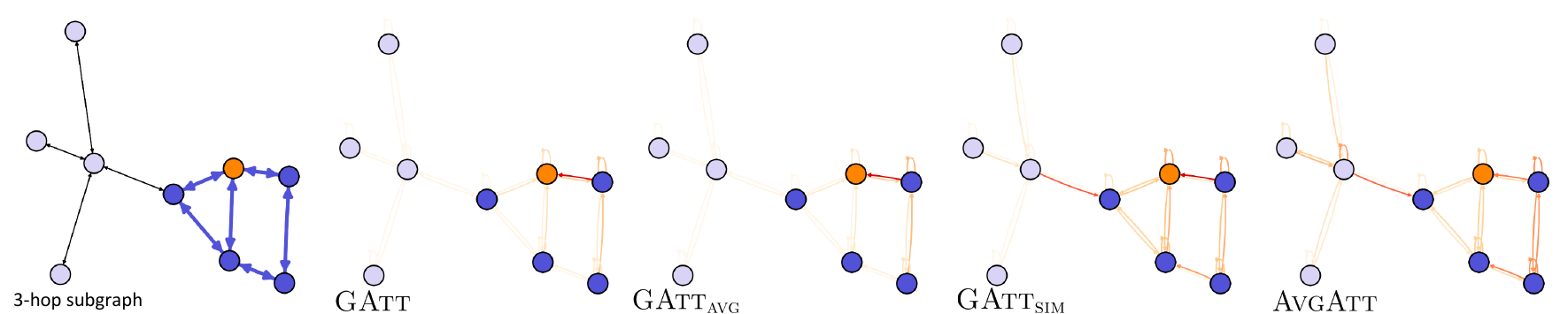}
    \caption{Node 676 (3-layer GATv2).}
  \end{subfigure}
  \begin{subfigure}[b]{0.35\linewidth}
    \includegraphics[width=\linewidth]{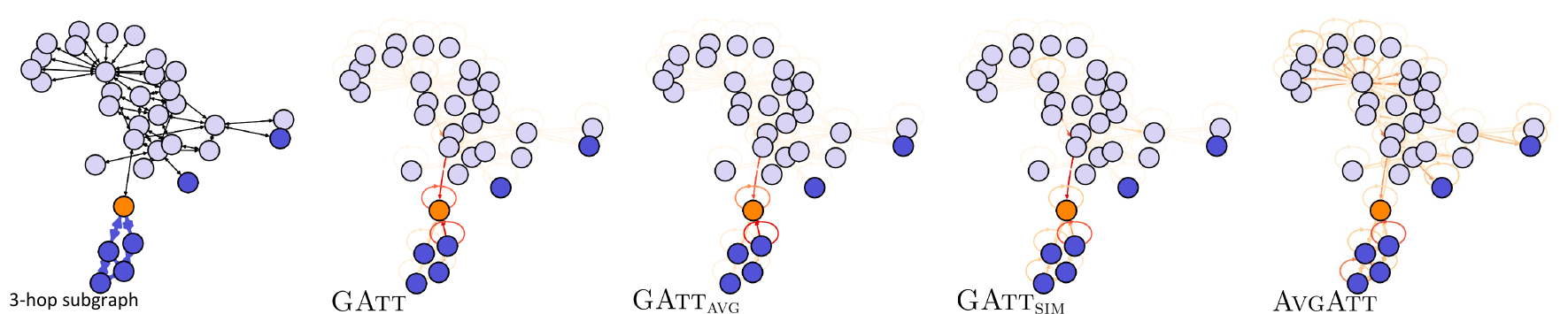}
    \caption{Node 687 (3-layer GATv2).}
  \end{subfigure}
  \caption{Case studies on the BA-Shapes dataset for GATv2.}
  \label{fig:SupplCaseStudiesBAShapesGATv2}
\end{figure}

\begin{figure}[h!]
  \centering
  \begin{subfigure}[b]{0.35\linewidth}
    \includegraphics[width=\linewidth]{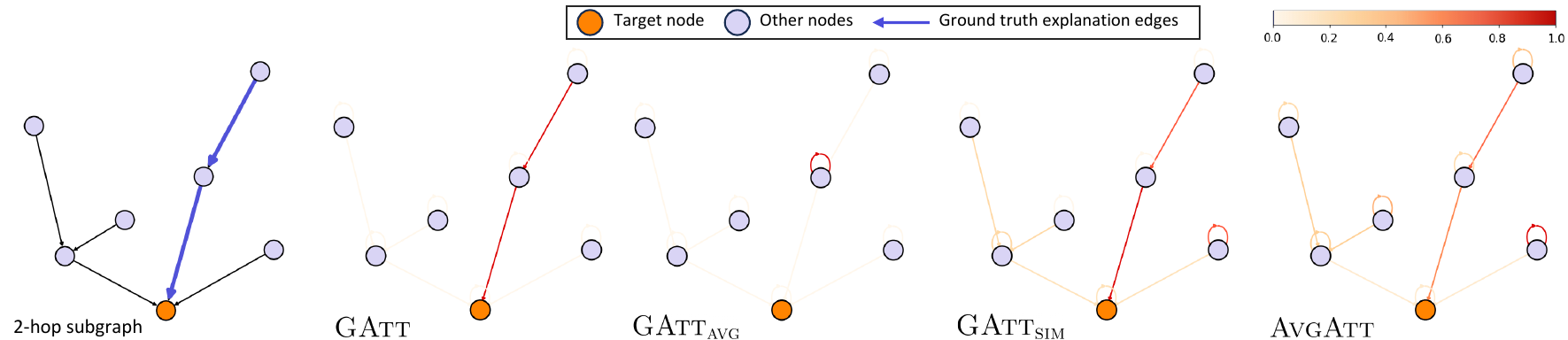}
    \caption{Node 2 (2-layer GATv2).}
  \end{subfigure}
  \begin{subfigure}[b]{0.35\linewidth}
    \includegraphics[width=\linewidth]{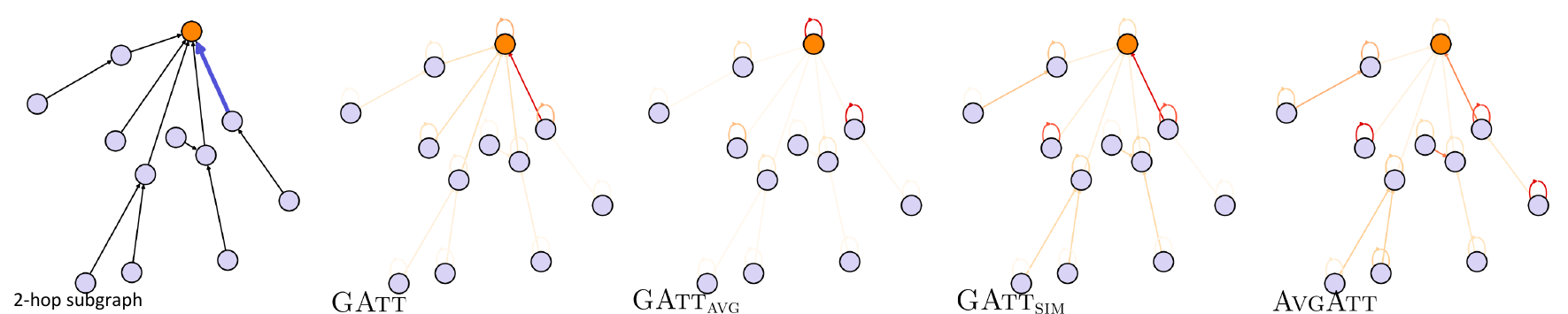}
    \caption{Node 44 (2-layer GATv2).}
  \end{subfigure}
  \begin{subfigure}[b]{0.35\linewidth}
    \includegraphics[width=\linewidth]{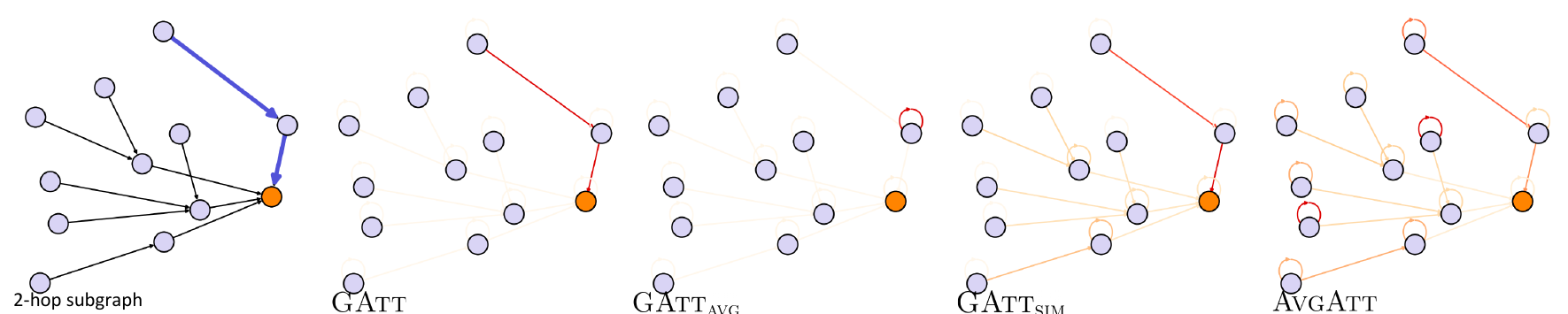}
    \caption{Node 54 (2-layer GATv2).}
  \end{subfigure}
  \begin{subfigure}[b]{0.35\linewidth}
    \includegraphics[width=\linewidth]{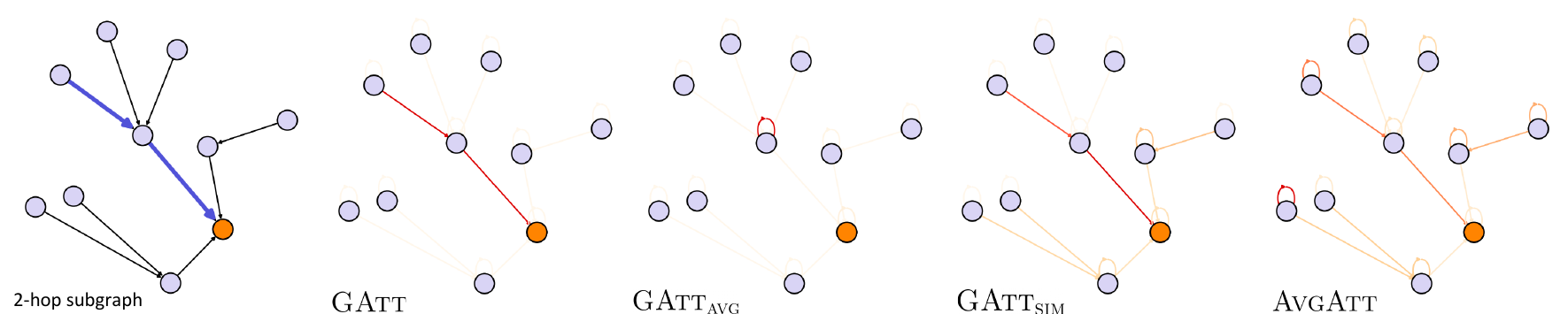}
    \caption{Node 355 (2-layer GATv2).}
  \end{subfigure}
  \begin{subfigure}[b]{0.35\linewidth}
    \includegraphics[width=\linewidth]{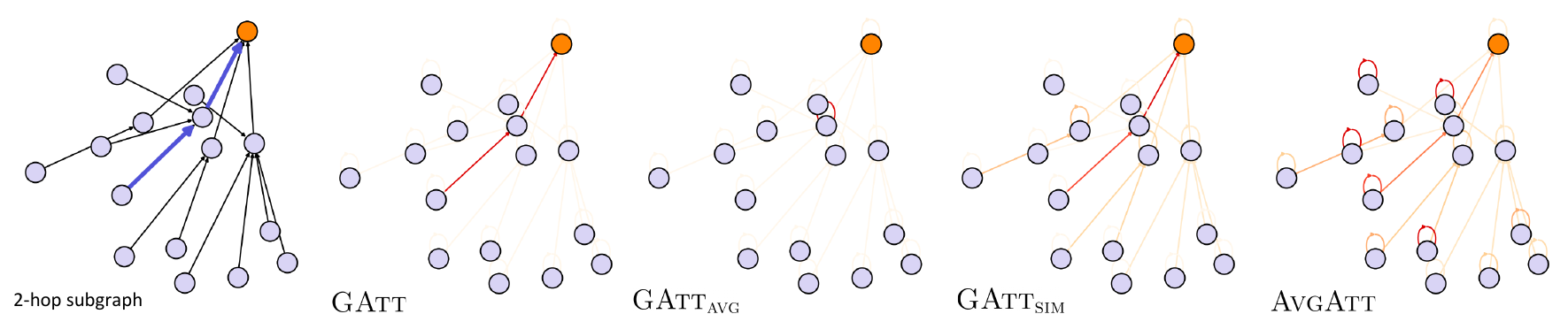}
    \caption{Node 326 (2-layer GATv2).}
  \end{subfigure}
  \begin{subfigure}[b]{0.35\linewidth}
    \includegraphics[width=\linewidth]{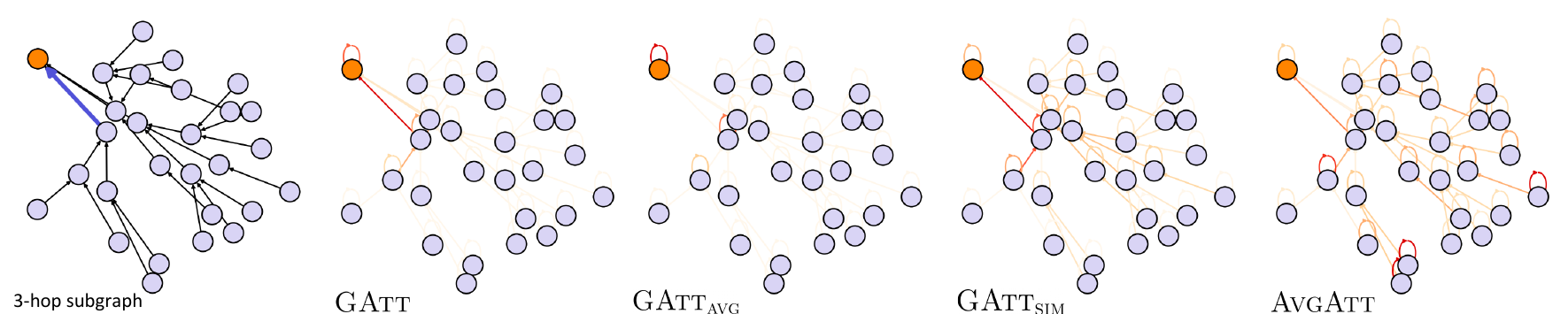}
    \caption{Node 2 (3-layer GATv2).}
  \end{subfigure}
  \begin{subfigure}[b]{0.35\linewidth}
    \includegraphics[width=\linewidth]{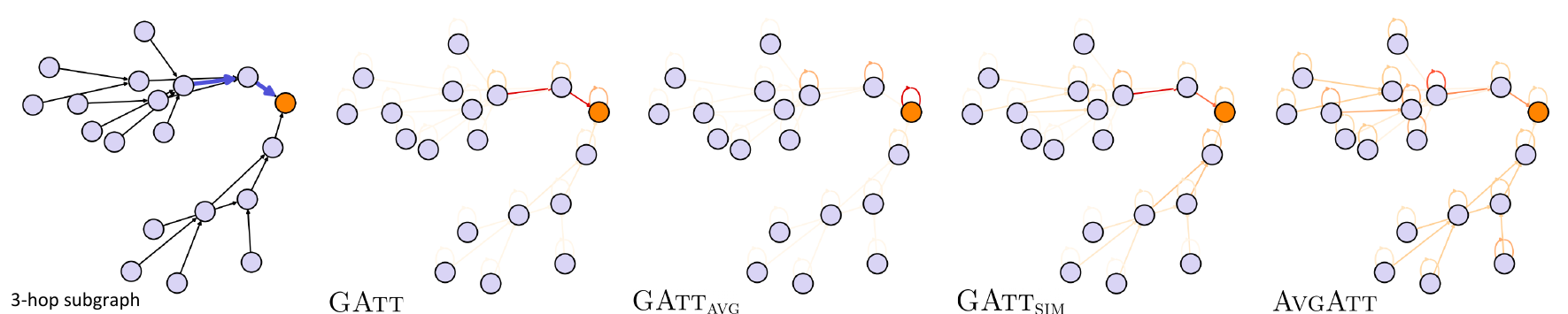}
    \caption{Node 44 (3-layer GATv2).}
  \end{subfigure}
  \begin{subfigure}[b]{0.35\linewidth}
    \includegraphics[width=\linewidth]{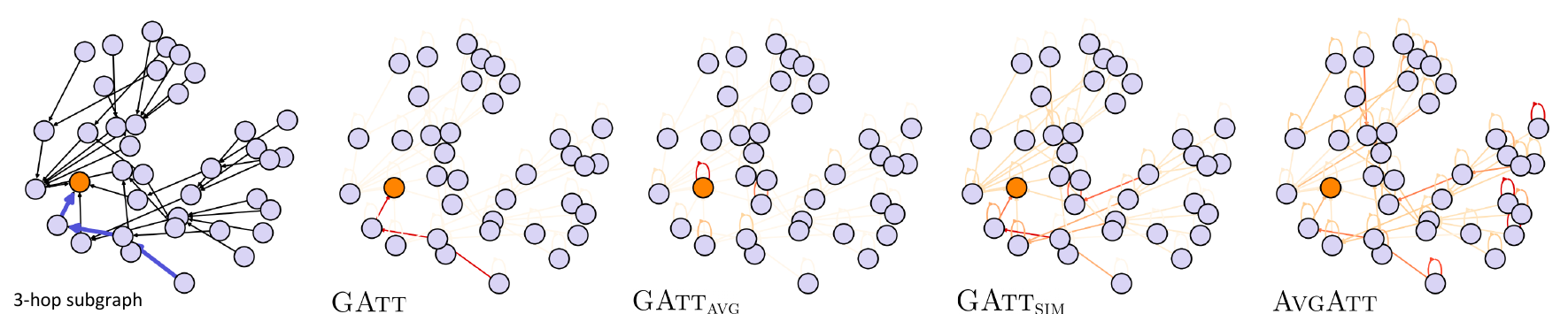}
    \caption{Node 54 (3-layer GATv2).}
  \end{subfigure}
  \begin{subfigure}[b]{0.35\linewidth}
    \includegraphics[width=\linewidth]{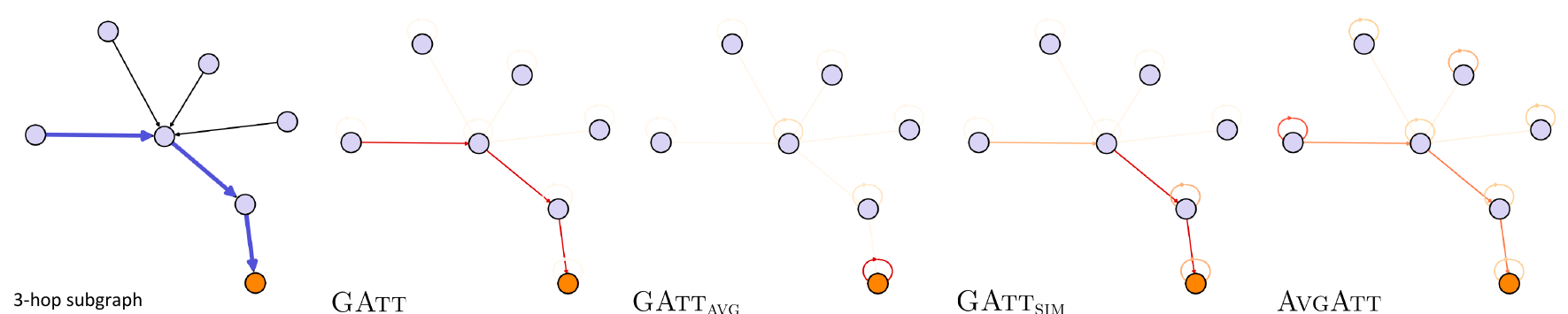}
    \caption{Node 355 (3-layer GATv2).}
  \end{subfigure}
  \begin{subfigure}[b]{0.35\linewidth}
    \includegraphics[width=\linewidth]{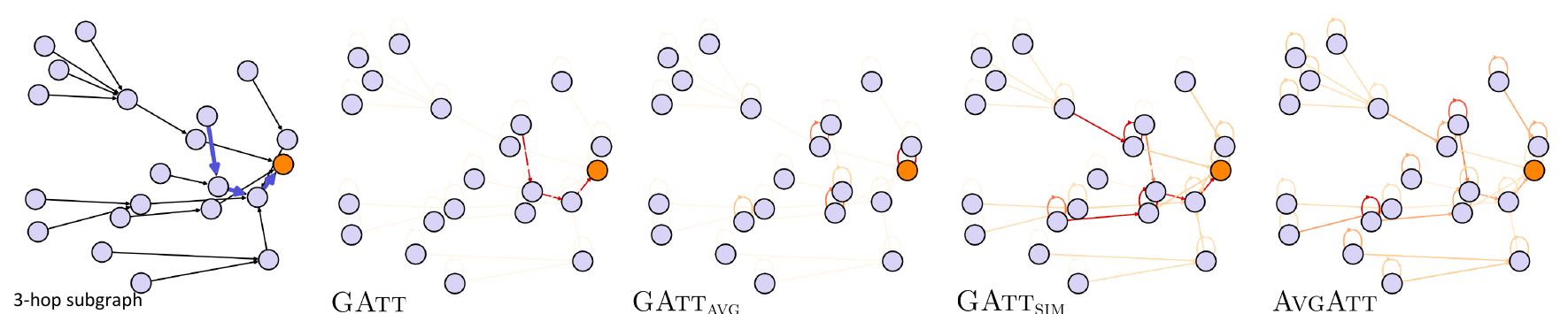}
    \caption{Node 326 (3-layer GATv2).}
  \end{subfigure}
  \caption{Case studies on the Infection dataset for GATv2.}
  \label{fig:SupplCaseStudiesInfectionGATv2}
\end{figure}

\section{Further Case Studies (Figures~\ref{fig:SupplCaseStudiesBAShapes}--\ref{fig:SupplCaseStudiesInfectionGATv2})}
\label{appendix:casestudies}
We further conduct case studies for the BA-Shapes and Infection datasets by both visualizing the ground truth explanations and the edge attributions for randomly selected target nodes. Figures~\ref{fig:SupplCaseStudiesBAShapes}, \ref{fig:SupplCaseStudiesInfection}, \ref{fig:SupplCaseStudiesBAShapesGATv2}, and~\ref{fig:SupplCaseStudiesInfectionGATv2} show multiple examples of applying different edge attribution calculation methods on BA-Shapes and Infection for GAT and GATv2 models. In all cases, we observe a consistent trend: 1) \textsc{GAtt} of GAT/GATv2s attends more on edges in the ground truth explanations, 2) \textsc{AvgAtt} fails to identify such behaviors and its edge attributions tend to be spread over the graph, and 3) \textsc{GAtt}$_\textsc{sim}$ shows edge attribution patterns that lie between those of \textsc{GAtt} and \textsc{AvgAtt}. We refer to Appendix~\ref{subsection:SuperGAT} for case studies on \textbf{SuperGAT}.

\end{document}